\documentclass[10pt,twocolumn,letterpaper]{article}

\usepackage[pagenumbers]{iccv} %

\usepackage[accsupp]{axessibility} 

\usepackage[utf8]{inputenc} %
\usepackage[T1]{fontenc}    %
\usepackage{url}            %
\usepackage{booktabs}       %
\usepackage{amsfonts}       %
\usepackage{nicefrac}       %
\usepackage{microtype}      %
\usepackage{multirow}   %
\usepackage{algorithmic}
\usepackage{algorithm}
\usepackage{wrapfig}
\usepackage{graphicx}
\usepackage{comment}
\usepackage{amsmath,amssymb} %
\usepackage{subcaption}
\usepackage{bbding}
\usepackage{pifont}
\usepackage{booktabs} %
\usepackage{amsthm}

\usepackage{amsmath,amsfonts,bm}

\def\1{\bm{1}}

\def\eps{{\epsilon}}

\def\vx{{\bm{x}}}
\def\vy{{\bm{y}}}

\def\mE{{\bm{E}}}

\def\mG{{\bm{G}}}
\def\mH{{\bm{H}}}

\def\mM{{\bm{M}}}
\def\mN{{\bm{N}}}

\def\mX{{\bm{X}}}

\DeclareMathAlphabet{\mathsfit}{\encodingdefault}{\sfdefault}{m}{sl}
\SetMathAlphabet{\mathsfit}{bold}{\encodingdefault}{\sfdefault}{bx}{n}

\def\gD{{\mathcal{D}}}

\def\gL{{\mathcal{L}}}

\def\gN{{\mathcal{N}}}

\def\gY{{\mathcal{Y}}}

\def\sR{{\mathbb{R}}}

\DeclareMathOperator*{\argmax}{arg\,max}

\usepackage{arydshln}
\usepackage[export]{adjustbox}
\usepackage{color}
\usepackage{soul}
\usepackage{multirow}
\usepackage{xcolor}
\usepackage{wrapfig}

\definecolor{darkred}{rgb}{0.7,0.1,0.1}
\definecolor{darkgreen}{rgb}{0.1,0.7,0.1}
\definecolor{cyan}{rgb}{0.7,0.0,0.7}
\definecolor{dblue}{rgb}{0.2,0.2,0.8}
\definecolor{maroon}{rgb}{0.76,.13,.28}
\definecolor{burntorange}{rgb}{0.81,.33,0}
\definecolor{tealblue}{rgb}{0.212,0.459, 0.533}
\definecolor{myyellow}{rgb}{0.8627451 , 0.75294118, 0.20784314]}

\definecolor{mypink}{rgb}{0.93359375, 0.62109375, 0.83984375}

\definecolor{pp}{rgb}{0.43921569, 0.18823529, 0.62745098}
\definecolor{rr}{rgb}{0.5254902 , 0.00784314, 0.12941176}
\definecolor{bb}{rgb}{0.09019608, 0.23529412, 0.37647059}
\definecolor{yy}{rgb}{0.49803922, 0.3372549 , 0.0}
\definecolor{gg}{rgb}{0.02352941, 0.3372549 , 0.17647059}

\ifdefined\ShowNotes
  \newcommand{\colornote}[3]{{\color{#1}\bf{#2: #3}\normalfont}}
\else
  \newcommand{\colornote}[3]{}
\fi

\definecolor{lightred}{rgb}{0.9,0.4,0.4}

\newcommand{\darkgreen}[1]{{\color{darkgreen}{#1}}}

\definecolor{almostblack}{HTML}{111111}

\definecolor{darkpink}{rgb}{0.98,0.81,0.89}

\definecolor{OursColor}{rgb}{0.83,0.89,0.94.}

\newcommand{\best}[1]{\textbf{#1}}
\newcommand{\second}[1]{\underline{#1}}

\definecolor{wolto}{rgb}{.75,0.75,0.75}

\usepackage{xcolor}
\usepackage{soul}

\newlength\savewidth

\definecolor{turquoise}{cmyk}{0.65,0,0.1,0.1}
\definecolor{purple}{rgb}{0.65,0,0.65}
\definecolor{darkgreen}{rgb}{0.0, 0.5, 0.0}
\definecolor{darkred}{rgb}{0.5, 0.0, 0.0}
\definecolor{darkblue}{rgb}{0.0, 0.0, 0.5}
\definecolor{blue}{rgb}{0.0, 0.0, 1.0}
\definecolor{orange}{rgb}{1.0,0.5,0.0}

\definecolor{code_bg}{rgb}{0.95, 0.95, 0.95}

\makeatletter
\renewcommand{\paragraph}{%
  \@startsection{paragraph}{4}%
  {\z@}{0.3ex \@plus 1ex \@minus .1ex}{-1em}%
  {\normalfont\normalsize\bfseries}%
}
\makeatother

\newif\ifproofread

\usepackage{symbols}
\usepackage{multirow}
\usepackage{pmboxdraw}
\usepackage{mathtools}

\usepackage{thm-restate}
\usepackage{enumitem}

\usepackage{listings}
\usepackage{colortbl}
\newcommand{\myparagraph}[1]{\vspace*{0pt}{\noindent \bf #1}}

\definecolor{iccvblue}{rgb}{0.21,0.49,0.74}
\usepackage[pagebackref, breaklinks, colorlinks, allcolors=iccvblue]{hyperref}

\def\paperID{6519} %
\def\confName{ICCV}
\def\confYear{2025}

\title{Heatmap Regression without Soft-Argmax 
\\ for Facial Landmark Detection}

\author{
    Chiao-An Yang \quad
    Raymond A. Yeh
    \\
    Department of Computer Science, Purdue University
}

\begin{document}
\maketitle
\begin{abstract}
Facial landmark detection is an important task in computer vision with numerous applications, such as head pose estimation, expression analysis, face swapping, etc. Heatmap regression-based methods have been widely used to achieve state-of-the-art results in this task. These methods involve computing the argmax over the heatmaps to predict a landmark. Since argmax is not differentiable, these methods use a differentiable approximation, Soft-argmax, to enable end-to-end training on deep-nets. In this work, we revisit this long-standing choice of using Soft-argmax and demonstrate that it is not the only way to achieve strong performance. Instead, we propose an alternative training objective based on the classic structured prediction framework. Empirically, our method achieves state-of-the-art performance on three facial landmark benchmarks (WFLW, COFW, and 300W), converging $2.2\times$ faster during training while maintaining better/competitive accuracy. Our code is available here\footnote{\url{https://github.com/ca-joe-yang/regression-without-softarg}}.
\end{abstract}

\section{Introduction}

Facial landmark detection,~\ie, finding a set of pre-defined landmarks on a given facial image, is an important and classic problem in computer vision. The detected landmarks can aid numerous downstream applications~\cite{gao2024self, Masietal:2018:deepfacesurvey, gu2019mask, zheng2021farl, deng2019arcface, bao2018towards, deng2019accurate, feng2021learning, Danecek2022EMOCA, wood20223d, jiang2005efficient, cao2014face, bulat2017far, kumar2018disentangling, qian2024gaussianavatars, liang2024generalizable},~\eg, face recognition, face alignment, face synthesis, head pose estimation, expression analysis, face swapping,~\etc. Research in facial landmark detection has a long history, from earlier works~\cite{cootes2001active, matthews2004active, cristinacce2008automatic, blanz2003face, saragih2007nonlinear, sauer2011accurate, kahraman2007active, cootes1992active,cootes1995active,milborrow2008locating, feng2015cascaded,xiong2013supervised} to deep learning based methods~\cite{sun2013deep, lin2021structure, xia2022splt}. Notably, methods based on heatmap regression~\cite{li2024towards, wu2018lab, dapogny2019decafa, hrnet, qian2019avs, dong2018style, kumar2020luvli, wang2019awing, JLS21pipnet, li20DAG, huang2021adnet, lan2021hih, zhou2023star, zhang2024open, micaelli2023recurrence, dang2025cascaded, gao2024self} have found success and gained popularity. 

As the name suggests, heatmap regression methods leverage an intermediate representation of heatmaps, one for each landmark. To predict a landmark, these methods compute the $\tt argmax$ over the coordinates of a given heatmap, \ie, it returns the coordinate that contains the maximum value in the heatmap. However, $\tt argmax$ is not differentiable, hence,~\citet{nibali2018numerical} propose \text{\tt Soft-argmax} to relax the $\tt argmax$ for end-to-end training. Regression losses,~\eg, $\ell_2$-loss, on the predicted coordinate is used to train the model. Next, another challenge in landmark detection is the semantic ambiguity in the annotations~\cite{wu2018lab, liu2019semantic, dong2018sbr, dong2020supervision, kumar2020luvli, huang2021adnet}, meaning that there is some degree of uncertainty in the annotations during the labeling process.
To address this issue, several works have studied how to design more robust loss functions~\cite{feng2018wing, wang2019awing, huang2021adnet},~\eg,~\citet{zhou2023star} propose STAR loss, a \textit{self-adaptive} loss to address the semantic uncertainty. 

In this work, we revisit the idea of \text{\tt Soft-argmax} and find that it may be unnecessary. Instead, we propose an alternative training objective based on the framework of structured prediction that does not require a differentiable coordinate prediction. To address the semantic ambiguity, we propose image-aware label smoothing, which blurs the annotation along the edges of the facial image to simulate label uncertainty.
With these two techniques, we achieve a heatmap regression-based method \textit{without} \text{\tt Soft-argmax} or a complex loss design.

We evaluate our approach on three established benchmarks, namely, WFLW~\cite{wu2018lab}, COFW~\cite{burgos-artizzu_perona_dollar_2022}, and 300W~\cite{sagonas2013300}. Our approach achieves state-of-the-art performance with a faster training convergence speed while being intuitive and principled.  
{\noindent\bf Our contributions are as follows:}
\begin{itemize}
\item We propose to train heatmap regression-based models using a training objective derived from the structured prediction framework and show that the common choice of {\tt Soft-argmax} may not be necessary.
\item We introduce an image-aware label smoothing technique to better capture annotation uncertainty.
\item Extensive experiments and ablation studies demonstrate our method's competitive performance to state-of-the-art while achieving a $2.2\times$ faster training convergence.
\end{itemize}

\section{Related Work}
We provide a high-level description of related works on facial landmark detection, structured prediction, and label smoothing. Technical details are reviewed in~\secref{sec:background}.

{\noindent\bf Facial landmark detection.}
The methods for landmark detection can be largely categorized into two branches: coordinates regression and heatmap regression.
Coordinate regression methods~\cite{zhou2013extensive, zhang2015learning, trigeorgis2016mnemonic, feng2018wing, browatzki20203fabrec, li2022repformer, li2022DTLD} consider the problem as a regression and directly predict the 2D coordinates.
~\citet{zhang2015learning} are the first to leverage additional facial features as auxiliary information. 
Based on empirical observation, Wing loss~\cite{feng2018wing} is proposed to increase the contribution of predictions with smaller errors.
MDM~\cite{trigeorgis2016mnemonic} introduces
the concept of memory in a coarse-to-fine pipeline. \citet{li2022repformer} employ landmark-to-landmark and landmark-to-memory attention modules to aggregate information to transformer queries. 
\citet{li2022DTLD} experiment with cascaded transformers to explore the structured relationship among facial landmarks.

On the other hand, heatmap regression methods predict landmark heatmaps as intermediate outputs.
The coordinates are then acquired by applying the soft approximation~\cite{tai2019fhr} of {\tt argmax} or other functions with similar effects. The main challenge, as opposed to coordinate regression which directly regresses to two numbers, is to be able to predict high-quality landmark heatmaps.
Stack hourglass network~\cite{newell2016stacked} is initially used in human pose estimation and is later introduced to increase the quality of facial landmark detection~\cite{yang2017hgfacial}. 
HRNet~\cite{hrnet} maintains high-resolution representations by exchanging the features from different stages. 
\citet{robinson2019laplace} consider facial landmarks as random variables under Laplace distributions and propose LaplaceKL loss to minimize its KL divergence between prediction and ground truth. 
\citet{yu2021heatmap} focus on solving the quantization error in heatmap regression. 
\citet{dapogny2019decafa} consider each stage as an individual alignment task. Auxiliary information, such as facial contours, is also discussed.
LAB~\cite{wu2018lab} is the first boundary-aware face alignment algorithm on facial landmark detection.

Besides the landmark heatmap, ADNet~\cite{huang2021adnet} generates an edge heatmap and a point heatmap as localization guidance in each hourglass network. It is later refined by STAR~\cite{zhou2023star} to address the annotation ambiguity, applying PCA to the predicted heatmaps to formulate the direction and intensity of the ambiguity. Unlike these heatmap regression works, our method avoids the {\tt Soft-argmax} approximation by taking inspiration from structured prediction. We also propose an image-aware label smoothing on the \textit{heatmaps annotations} to address semantic ambiguity, whereas~\citet {zhou2023star} smooths the heatmap predictions. Recently, DISPAL~\cite{lan2024does} has proposed using the similarity among values of neighboring pixels in addition to the maximum of the heatmap for landmark prediction. 

Note, as these works use {\tt Soft-argmax}, our proposed method can complement their approach. We demonstrate this using STAR~\cite{zhou2023star}, the current SOTA with public code.

{\noindent\bf Structured prediction} aims to model relations between the output variables. Seminal works~\cite{lafferty2001conditional,taskar2003max,tsochantaridis2005large} (\eg, Structured Support Vector Machine (SSVM)/ Max-Margin Markov Network) expanded the framework of linear classifiers (\eg, SVM, logistic regression) to consider more complex output structures. 
Beyond linear models, structure prediction has also been incorporated into deep-nets~\cite{bridle1989training,domke2013structured,li2014high,lin2015deeply,tompson2014joint,chen2014semantic,chen2015learning,schwing2015fully,zheng2015conditional,yeh2017interpretable,graber2018deep,
graber2019graph}. 
A concise tutorial has also been tailored for the computer vision audience~\cite{iccv2015schwing}, with applications in classification and semantic segmentation. In this work, we derive the training objective from the deep structure prediction framework and achieve state-of-the-art performance in the highly competitive task of facial landmark detection. 

{\noindent\bf Label smoothing.} Deep-nets are known to overfit and be overconfident in their prediction. One common solution is to train the model using ``soft'' targets. Instead of a one-hot target with a probability of one in a single class, the target is smoothed with a uniform distribution over all classes~\cite{szegedy2016rethinking,he2019bag} or directly penalizing the confidence of a prediction~\cite{ pereyra2017regularizing,muller2019does}. More advanced label smoothing techniques have also been proposed,~\eg, smoothing in a data-dependent way for a classification problem~\cite{li2020regularization}. While our label smoothing is also data-dependent, our smoothing method is
inspired by the uncertainty of the annotator; hence, we smooth around the facial boundaries.

\section{Background}\label{sec:background}
We briefly review the necessary background and notation to understand this paper.

{\noindent\bf Landmark detection.}
Let $\mX \in \sR^{3 \times H \times W}$ to be an input facial image labeled with $N$ landmarks $\vy = [\vy_1, \vy_2, \cdots, \vy_N]$, where each $\vy_n$ represents the pixel location $(u,v) \in \{0, \dots, H-1\}\times \{0, \dots, W-1\}$ of the landmark, \ie, $\vy \in \sR^{N \times 2}$. 
The task of landmark detection is to learn a model, parameterized by $\theta$, that predicts landmark $\hat\vy$ given the input face image $\mX$.  

{\noindent\bf Heatmap regression.}
Seminal work by~\citet{newell2016stacked} proposes a stacked hourglass architecture and formulates the learning of the model as a heatmap regression task. In heatmap regression methods, given an input, the model outputs $N$ heatmap $\hat\mH = [\hat\mH_1, \dots, \hat\mH_N] \in \sR^{N\times H \times W}$ where each heatmap $\hat\mH_n$ represents a score of each pixel being a landmark. To predict a landmark, one will compute the argmax for each of the heatmaps, \ie,
\bea\label{eq:inference}
\hat{\vy}_n = \argmax_{\vy_n' \in \gY} \hat\mH_n[\vy_n'],
\eea
where $\gY \triangleq \{0, \dots, H-1\}\times \{0, \dots, W-1\}$ denotes all the possible pixel locations.

To train the model,~\citet{newell2016stacked} create ``ground-truth'' heatmaps $\mH_n$ by constructing a 2D Gaussian centered at each ground-truth landmark $\vy_n$. Training is formulated as minimizing the mean-squared error between the predicted heatmap $\hat\mH$ and the ground-truth $\mH$:
\bea\label{eq:l2_map}
\min_\theta \sum_{(\mX,\vy)\in\gD}\|\hat\mH(\mX;\theta)-\mH(\vy)\|_2^2,
\eea
where $\gD$ denotes the training dataset. 

While this formulation successfully learns a model, it is unclear whether using mean-square error on heatmaps is a suitable loss. Importantly, the evaluation metric of landmark detection compares between the predicted and ground-truth \textit{landmark} and \textit{not} heatmaps. One idea is to use a differentiable argmax such that the models can be trained using a loss function on the landmarks, which we discuss next.

{\noindent\bf Soft-argmax.}
To enable training using the gradient method on losses between the predicted and ground-truth landmarks,~\citet{nibali2018numerical} propose a soft-argmax operation. Given a heatmap $\mH_n$, the {\tt Soft-argmax} normalizes it into a probability distribution and computes the expectation over the coordinates. Formally, $\tilde{\vy}_n =$
\bea\label{eq:soft-argmax}
 {\text{\tt Soft-argmax}}(\hat\mH_n) \triangleq \sum_{\vy_n'} \vy_n' \cdot \texttt{Softmax}(\hat\mH_n)[\vy_n'],
\eea
where \texttt{Softmax} normalizes over all the spatial dimensions to form a probability distribution.
With {\tt Soft-argmax} defined, the model can be trained by minimizing a loss, \eg, $\ell_2$-loss as follows:
\bea\label{eq:l2_soft}
\min_\theta \sum_{(\mX,\vy)\in\gD} \sum_n \|\tilde\vy_n - \vy_n\|_2^2.
\eea
While soft-argmax has been widely used~\cite{lan2021hih, zhou2023star}, we re-examine whether such relaxation is a good design choice.

{\noindent \bf Deep structured learning.}
The design of training objective over an argmax in~\equref{eq:inference} is well studied in structured prediction. A typical choice for linear models is to use the structural hinge loss~\cite{taskar2003max}, leading to Structured-SVMs~\cite{tsochantaridis2005large}. 

\citet{iccv2015schwing, chen2015learning} further generalize the idea and present deep structured learning with a training objective: 
\bea\label{eq:ds_learn}\nonumber
&\hspace{-6cm}\min_\theta \frac{C}{2}\|\theta\|_2^2+\\ & \hspace{-0.5cm}\sum_{(\mX,\vy) \in \gD} \underbrace{\max_{\hat\vy} (\Delta(\vy, \hat\vy) + F(\hat\vy,\mX;\theta)) }_{\small \text{Loss-augmented inference}} - F(\vy,\mX;\theta),
\eea
where $\Delta$ denotes the margin term ({\it a.k.a.} task-loss), and $F$ denotes a score function represented by a deep-net with parameters $\theta$, and $C \in \sR$ is a hyper-parameter controlling the regularization term.
Note, in the case where the score is a linear model, \ie, a ``one layer deep-net'' with the choice $F=\theta^\intercal\texttt{vec}(\mX)$, then~\equref{eq:ds_learn} recovers the objective of a Structured-SVM. 

In this work, we demonstrate that this framework provides a principled alternative to the existing training methods for
heatmap regression in~\equref{eq:l2_map} or soft-argmax in~\equref{eq:l2_soft} commonly used by prior works~\cite{huang2021adnet, zhou2023star, lan2021hih}. 

\section{Approach}
We first analyze the shortcomings of {\tt Soft-argmax}. We then describe how to formulate landmark detection with the deep structured learning framework along with a label smoothing technique to handle noisy annotations. 

\subsection{Limitation of Soft-argmax}
\label{sec:limit}
As reviewed in~\secref{eq:soft-argmax}, {\tt Soft-argmax} approximates the $\tt argmax$ with the expected value of the argument over a distribution form by normalizing a heatmap. Below, we show why this approximation may not be desirable. For readability, we illustrate the example using a ``1-D heatmap''. Let the ground-truth landmark $\vy=2$, and consider two heatmaps in~\equref{eq:example} and~\equref{eq:example2}.
\bea\label{eq:example}
\def\arraystretch{1.2}
\begin{array}{|c|c|c|c|c|c|}
\multicolumn{6}{c}{\text{Soft-argmax } \tilde{\vy}^{(1)}=2}\\
\hline
\vy'^{(1)} & 0 & 1 & 2 &3 &4\\
\hline
{\tt Softmax}(\hat\mH^{(1)}) & 0.0 & 0.0 & 1.0 & 0.0 & 0.0\\
\hline
\multicolumn{6}{c}{
\text{\phantom{\Large A} Unimodal heatmap}}\\
\end{array} 
\\
\def\arraystretch{1.2}
\begin{array}{|c|c|c|c|c|c|}
\multicolumn{6}{c}{\text{Soft-argmax } \tilde{\vy}^{(2)}=2}\\
\hline
\vy'^{(2)} & 0 & 1 & 2 &3 &4\\
\hline
{\tt Softmax}(\hat\mH^{(2)}) & 0.4 & 0.1 & 0.0 & 0.1 & 0.4\\
\hline
\multicolumn{6}{c}{
\text{\phantom{\Large A}Bimodal heatmap}}\\
\end{array} 
\label{eq:example2}
\eea
{\it Observation:} Both the unimodal and bimodal heatmap results in the same {\tt Soft-argmax} output $\tilde{\vy}^{(1)}=\tilde{\vy}^{(2)}=2$. In other words, the $\ell_2$-loss $\norm{\tilde{\vy}^{(2)}-\vy}=0$ which means the training has converged. 
However, the {\bf bimodal heatmap makes an incorrect} $\tt argmax$ prediction as $\{0,4\} = \argmax_{\vy'} \hat\mH^{(2)}[\vy'] \neq 2$. 
More cases with a mismatch can be easily found, \eg, when there are even more peaks. We believe this mismatch may lead to optimization difficulties.

We empirically validate the hypothesis and provide empirical study in our experiments (see Sec.~\ref{exp:toy}). With this observation in mind, we rethink whether {\tt Soft-argmax} is necessary to train an effective model. More analysis is provided in Appendix~\ref{supp:analysis-softargmax}. We now discuss our proposed training objective based on structured prediction.

\subsection{Deep Structured Landmark Detection}

{\bf \noindent Problem formulation.}
As reviewed in~\secref{sec:background}, given an image $\mX$, the model aims to predict the ground truth landmarks $\vy$ consist $N$ individual landmarks $\vy_n$, each corresponding to an $(u,v)$ 2D landmark coordinates, \eg, the top of the lips. 

We propose to formulate the training of a landmark detection model as deep structured learning in~\equref{eq:ds_learn}. The key question is how to design the score function $F(\vy, \mX; \theta)$. First, we choose a score function that is decomposed into a sum of score functions $F_n$ for each landmark, \ie,
\bea
F(\vy,\mX;\theta) \triangleq \sum_n F_n(\vy_n,\mX;\theta),
\eea
where $F_n(\vy_n,\mX;\theta)$ corresponds to a channel of the stacked hourglass architecture's output. More precisely, the score $ F_n(\vy_n,\mX;\theta) \triangleq \hat{\mH}_n[\vy_n]$. 

Substituting the function of $F$ into~\equref{eq:ds_learn} and simplify, we arrive at the training objective:
\bea\nonumber
&\hspace{-.1cm}\min_\theta \frac{C}{2}\|\theta\|_2^2  + \sum_{(\mX,\vy) \in \gD} \sum_{n=1}^N \max_{\hat\vy_n} {\color{blue}\Large(}\Delta(\vy_n, \hat\vy_n) +\\ &\hspace{-0.2cm} F_n(\hat\vy_n,\mX;\theta){\color{blue}\Large)} - F_n(\vy_n,\mX;\theta).
\eea
Next,
the objective can be relaxed using the fact that
\bea
\epsilon \ln \sum_i \exp \frac{\vx_i}{\epsilon} 
\xrightarrow[]{\epsilon \rightarrow 0} \max_i \vx_i
\eea
to obtain the final training objective
\bea\label{eq:loss-augmented}
\min_\theta \frac{C}{2}\|\theta\|_2^2 +
\sum_{(\mX, \vy) \in \gD} \sum_n \gL(\mX, \vy_n, \theta),
\eea
where we subsume all terms into a loss $\gL(\mX, \vy_n, \theta)$ as 
\bea\nonumber
\epsilon \ln \left(\sum_{\hat \vy_n}  \exp \frac{\Delta(\vy_n, \hat \vy_n) + F_n(\hat \vy_n, \mX, \theta)}{\epsilon}\right) - F_n(\vy_n, \mX, \theta).
\eea

For an effective margin $\Delta$, it needs to capture some notion of ``similarity'' for the task, \ie, how close is the prediction $\hat\vy_n$ to ground-truth $\vy_n$ and satisfy $\Delta(\vy,\vy)=0$.
For simplicity, we 
choose the margin term to be a standard $\text{smooth-}\ell_1$ distance~\cite{girshick2015fast}, \ie,
\bea
\Delta(\vy_n, \hat\vy_n) = 
\alpha \cdot \begin{cases}
\frac{0.5}{s} \|\hat\vy_n - \vy_n\|^2_2 \text{, if } \|\hat\vy_n - \vy_n\|^1_1 < s,
\\
\|\hat\vy_n - \vy_n\|^1_1 - 0.5s \text{, otherwise}.
\end{cases}
\label{eq:smooth_l1}
\eea
This is a widely used distance function in previous work ~\cite{huang2021adnet, zhou2023star}.
Here, the threshold $s$ is set as $0.01$ and $\alpha$ is a weighting factor controlling the significance of the margin term. 

Finally, any gradient-based optimization methods,~\eg, Adam~\cite{adam}, can be used to optimize the training objective in~\equref{eq:loss-augmented}. We note that as $\hat{\vy}_n$ is not a function of $\theta$, no gradients are computed through the margin term $\Delta$.

\begin{figure}[t]
    \small
    \centering
    \includegraphics[width=\linewidth]{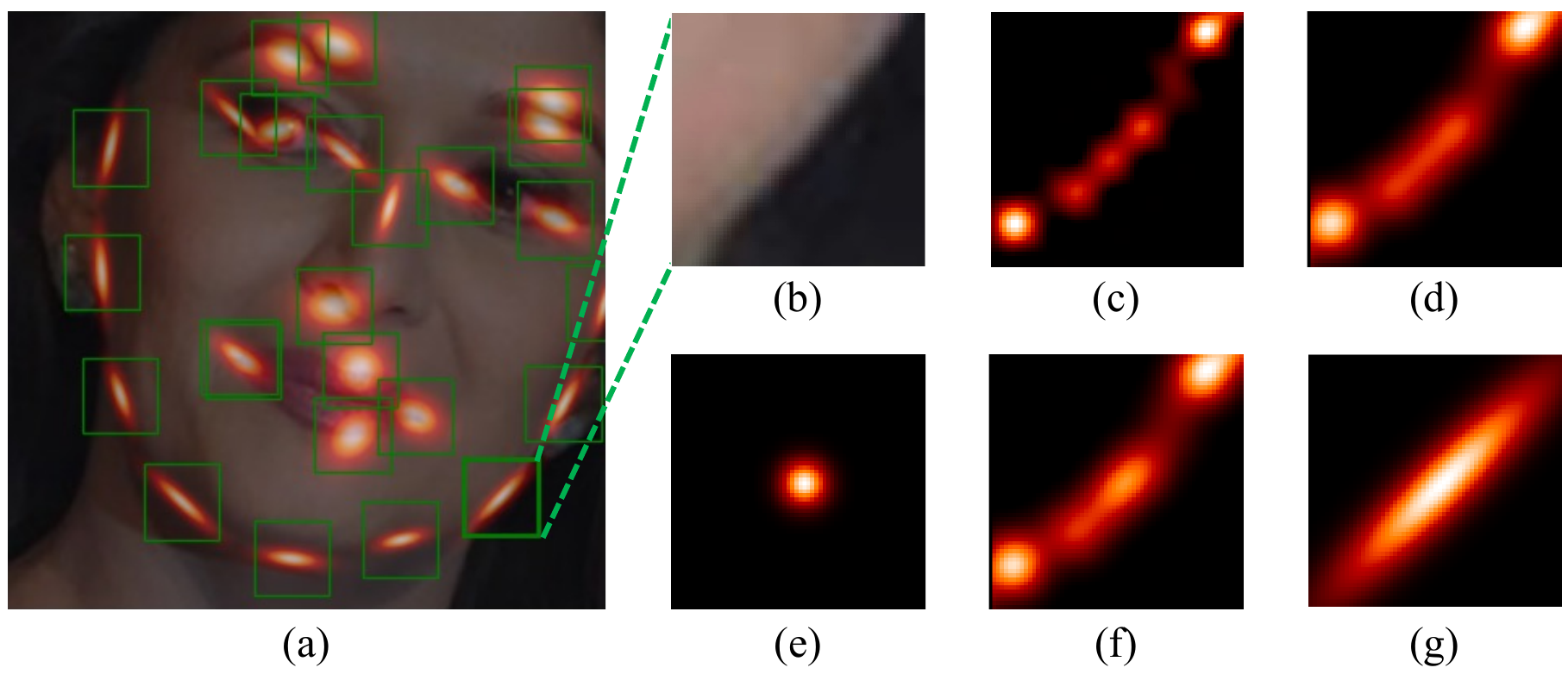}
    \vspace{-.55cm}
    \caption{\textbf{Our label smoothing.} We visualize the pipeline of our implementation on label smoothing. (a) is the overall label smoothing result. (b-g) are intermediate products cropped around the selected ground truth landmark $\vy$. (b) is the input image patch. (c) is the unprocessed pseudo edge heatmap. (d) is the processed edge heatmap. (e) is a Gaussian heatmap centered around $\vy$. (f) is the weighted summation of (d) and (e). (g) is the Gaussian heatmap based on the covariance of (f).
    }
    \vspace{-0.3cm}
    \label{fig:label_smoothing}
\end{figure}

{\noindent\bf Image-aware label smoothing.} The annotation for landmark detection is noisy due to semantic ambiguity~\cite{wu2018lab,
liu2019semantic,
kumar2020luvli}. To handle this uncertainty, we propose to use label-smoothing. Instead of having a single ``one-hot'' annotation, we construct a smooth distribution $\mG$ over it, \ie, the training objective in~\equref{eq:loss-augmented} is smoothed into
\bea\label{eq:loss-smooth}
\min_\theta \frac{C}{2}\|\theta\|_2^2  + \sum_{(\mX, \vy) \in \gD} 
\sum_n \sE_{\vy'_n \sim \mG(\vy_n)} \left[
\gL(\mX, \vy'_n, \theta)
\right],
\eea
where we approximate the expected value with Monte Carlo samples.

As in prior works of semantic ambiguity, the insight into constructing this smooth distribution $\mG$ comes from the observation that the annotation \textit{varies along} the local edges around the landmark (see~\figref{fig:label_smoothing} (a), (b)). 
To construct $\mG$, we leverage the pseudo edge heatmap (shown in~\figref{fig:label_smoothing} (c)) generated from the landmark annotations. To create such a pseudo edge heatmap, we follow a hand-designed procedure in prior works~\cite{wu2018lab, wang2019awing}. This involves interpolating landmark annotations to create boundary lines followed by a distance transform which weights each pixel by its distance to the boundary lines. 

Next, we notice the edge heatmap is often disjointed; hence,
we apply blurring and sharpening transformations to enhance the quality (\figref{fig:label_smoothing} (d)). To ensure that the ground-truth location is the maximum location, we add a normal Gaussian distribution (\figref{fig:label_smoothing} (e)) centered at each landmark, resulting in~\figref{fig:label_smoothing} (f). Finally, we extract a Gaussian distribution centered at $\vy_n$ with a covariance estimated from (f) to be used as our smoothed distribution $\mG$.
Please note that the pseudo edge heatmap is generated from ground truth landmark annotations and was also used by the compared baselines~\cite{zhou2023star, huang2021adnet}, hence, no additional data was used.

\begin{figure*}[t]
\centering
    \begin{subfigure}[h!]{0.3\textwidth}
        \includegraphics[width=\textwidth]{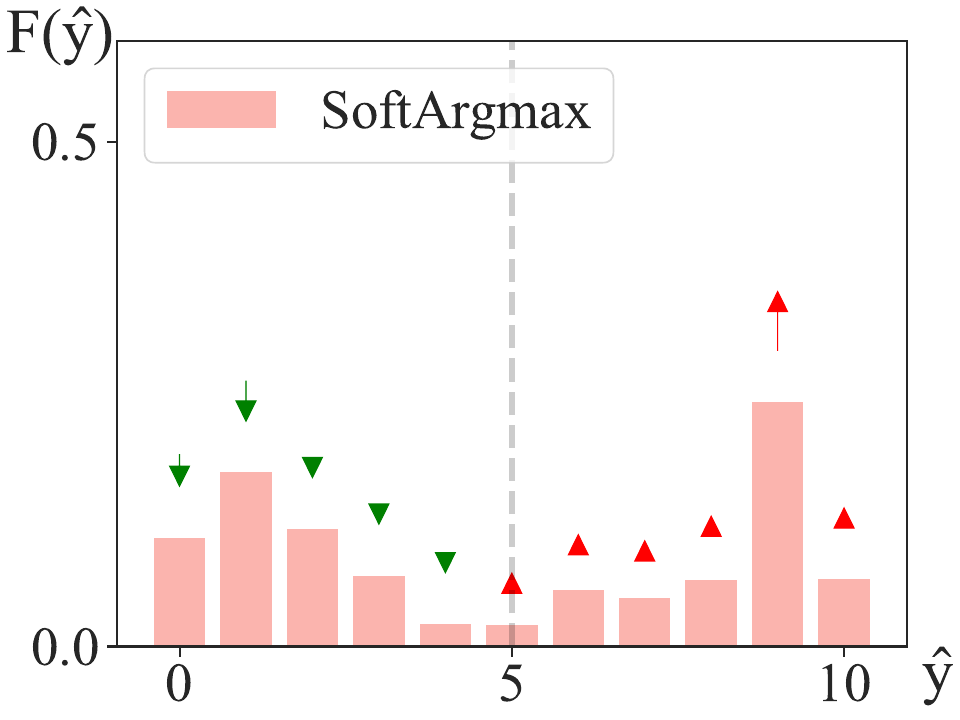}
        \caption{{\tt Soft-argmax} at step 10.}
    \end{subfigure}
    \hfill
    \begin{subfigure}[h!]{0.3\textwidth}
        \includegraphics[width=\textwidth]{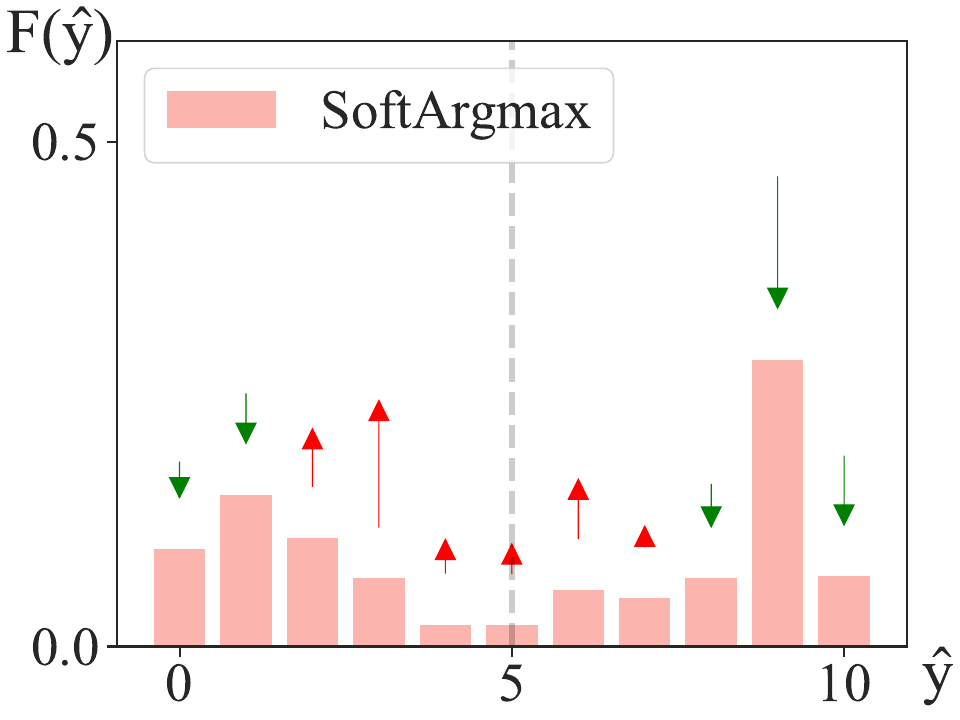}
        \caption{{\tt Soft-argmax} at step 20.}
    \end{subfigure}
        \hfill
    \begin{subfigure}[h!]{0.3\textwidth}
        \includegraphics[width=\textwidth]{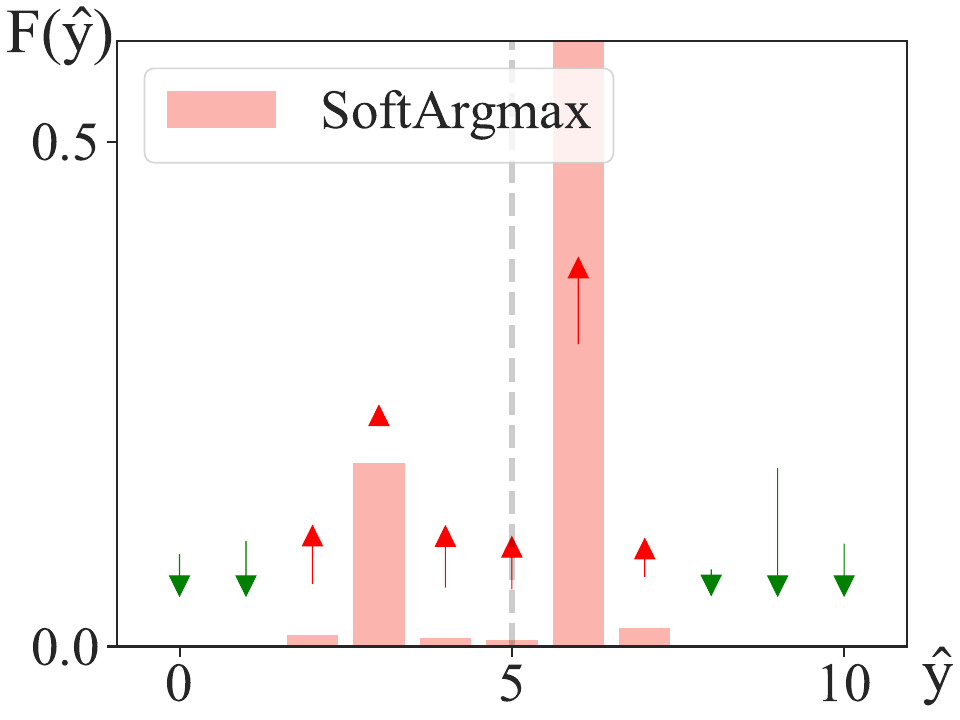}
        \caption{{\tt Soft-argmax} at step 50.}
    \end{subfigure}
    \begin{subfigure}[h!]{0.3\textwidth}
        \includegraphics[width=\textwidth]{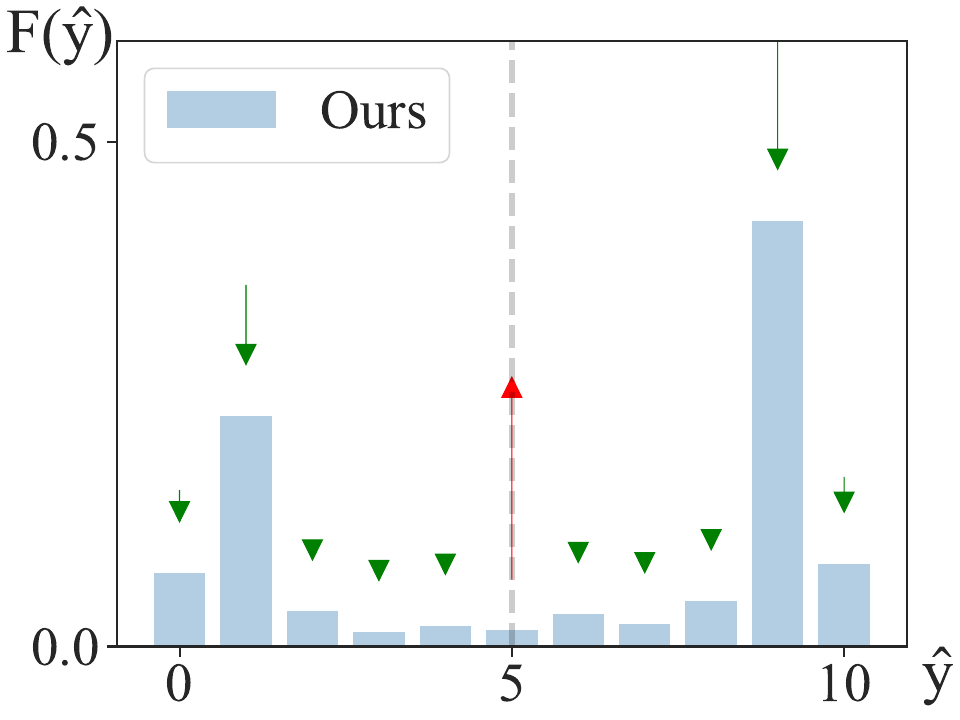}
        \caption{Ours at step 10.}
    \end{subfigure}    
    \hfill
    \begin{subfigure}[h!]{0.3\textwidth}
        \includegraphics[width=\textwidth]{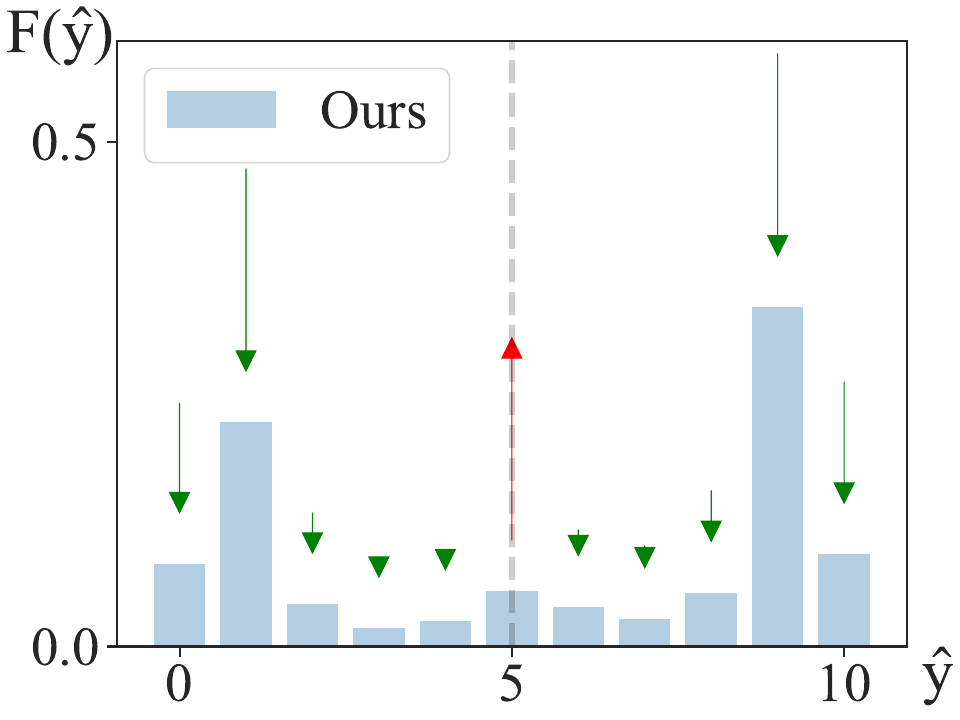}
        \caption{Ours at step 20.}
    \end{subfigure}
    \hfill
    \begin{subfigure}[h!]{0.3\textwidth}
        \includegraphics[width=\textwidth]{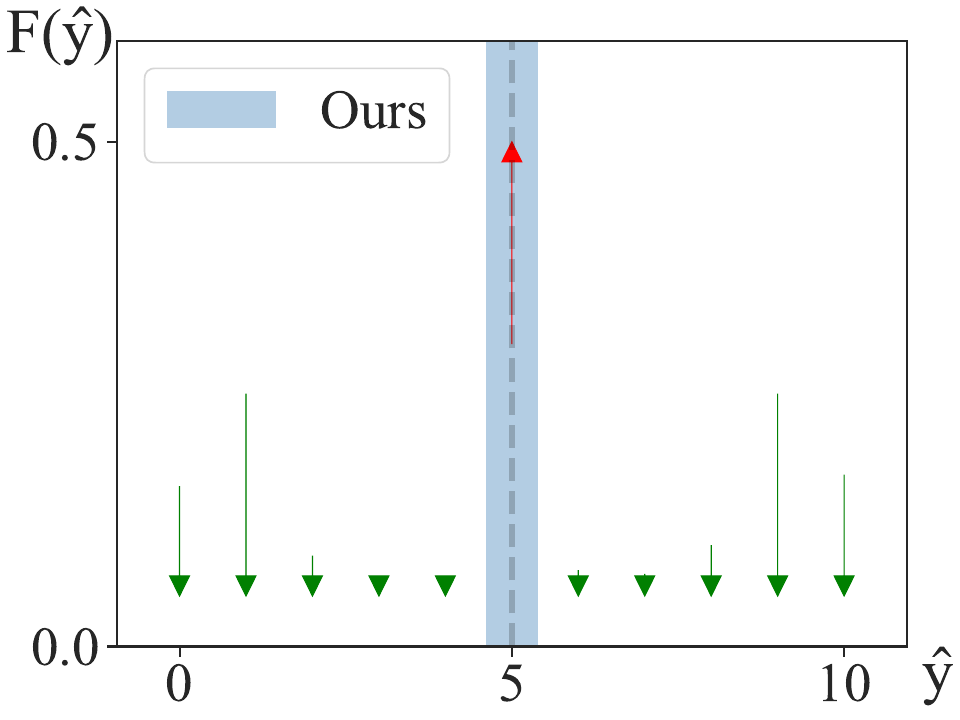}
        \caption{Ours at step 50.}
    \end{subfigure}
  \vspace{-0.2cm}
  \caption{
  \textbf{Comparison between {\tt Soft-argmax} and ours.} (a-c) visualize each $F$ trained with Eq.~\eqref{eq:l2_soft}, i.e. {\tt Soft-argmax} and $\ell_2$-loss at step 10, 20, and 50. (d-f) visualize each $F$ trained with our loss at steps 10, 20, and 50. We use \textcolor{red}{$\uparrow$} to denote where $F$ is increasing,~\ie, positive updates, and \darkgreen{$\downarrow$} to denote where $F$ is decreasing,~\ie, negative updates.
  }
  \vspace{-.3cm}
  \label{fig:toy}
\end{figure*}

\section{Experiments}\label{sec:exp}
We start with a simplified model to analyze the differences between our training objective versus $\ell_2$-loss + {\tt Soft-argmax}. We then conduct experiments on three standard landmark detection benchmarks, followed by ablation studies. Finally, we show that our loss achieves significantly faster convergence speed.

\subsection{Analysis on a simplified model}
\label{exp:toy}
We consider a simplified model with a single landmark on a 1D heatmap. We choose the dataset to have a single training sample at $\vy=5$, in this case, we can remove the dependencies on the input $\mX$ and directly model the heatmap with the model parameters. That is,
the heatmap $\hat\mH[k] \triangleq \theta_k$ is parameterized by $\theta \in \sR^{11}$. 
In~\figref{fig:toy}, we visualize how gradient descent updates the heatmap, \ie, $\theta^{(i+1)} \leftarrow \theta^{(i)} - \eta \nabla_\theta \gL$, when optimizing with the objective $\ell_2$-loss + {\tt Soft-argmax} (\equref{eq:l2_soft}), or our proposed method (\equref{eq:loss-augmented}). We initialize $\theta^{(0)}$ to be a bimodal distribution with  $\theta^{(0)}_9$ and $\theta^{(0)}_1$ each having the highest value and second highest values. For our method, we choose the margin term $\Delta$ to be the $\ell_2$-loss.

In~\figref{fig:toy} (a-c), we show how $\theta$ changes when training with $\ell_2$-loss + {\tt Soft-argmax} (\equref{eq:l2_soft}) with arrows showing the update. As can be observed, $\theta_{5}$ is being updated in the correct direction, however, its update magnitude is fairly small. Furthermore, the updates for the other parameters are not consistent.
Remember, as $\vy=5$, the heatmap should have $\theta_5$ being the largest element.

In~\figref{fig:toy} (d-f), we show how $\theta$ changes when training with our proposed method (\equref{eq:loss-augmented}). We observe that the updates are straightforward. The parameter $\theta_{5}$ consistently receives positive updates, while all other parameters receive negative updates. After a few steps, the argmax is already located $\hat\mH[5]$, with a significant gap between its maximum and second maximum. That is, we observe faster convergence of our approach.
Overall, we observe that $\ell_2$-loss + {\tt Soft-argmax} gradually ``moves'' the peaks towards the target $\vy=5$. In contrast, our proposed approach (\equref{eq:loss-augmented}) directly increases the peak at the target $\vy=5$ while decreasing the others.

\begin{table}[t]
\centering
    \begin{tabular}{lcccccccccc} 
    \specialrule{.15em}{.05em}{.05em}
    Method & 
    Type  & 
    \multicolumn{1}{c}{NME $\downarrow$} & $\text{FR} \downarrow$ & $\text{AUC} \uparrow$ \\ 
    \midrule
    Wing~\cite{feng2018wing} & C & 4.99 & 6.00 & 0.550 \\
    SLPT~\cite{xia2022splt} &  C &  4.14 & 2.76 & 0.595 \\
    RePFormer~\cite{li2022repformer} & C & 4.11 & - & - \\
    DTLD~\cite{li2022DTLD} & C & {4.08} & 2.76 & -  \\ 
    \midrule
    DeCaFa~\cite{dapogny2019decafa} & H & 4.62 & 4.84 & 0.563 \\
    HRNet~\cite{hrnet} & H & 4.60 & 4.64 & -  \\
    LUVLi~\cite{kumar2020luvli} & H & 4.37 & 3.12 & 0.577 \\
    Awing~\cite{wang2019awing} & H & 4.36 & 2.84 & 0.572  \\
    PIPNet~\cite{JLS21pipnet} & H & 4.31 & - & -  \\
    DAG~\cite{li20DAG} & H & 4.21 & 3.04 & 0.589  \\
    ADNet~\cite{huang2021adnet} & H  & 4.14 & 2.72 & {0.602} \\
    HIH~\cite{lan2021hih} & H & 
    {4.08} & {2.60} & \second{0.605}  \\
    KeyPosS~\cite{bao2023keyposs} & H & \second{4.00} & - & - \\
    STAR~\cite{zhou2023star}     & H  & 
    {4.02} & \second{2.32} & \second{0.605} \\
    \rowcolor{OursColor} Ours & H & 
    \best{3.99} & \best{1.84} & \best{0.606}
    \\
    \specialrule{.15em}{.05em}{.05em}
    \end{tabular}
    \vspace{-0.1cm}
    \captionof{table}{
\textbf{Comparison to SOTA facial landmark detection methods on the full set of WFLW.} We report the inter-ocular NME$\downarrow$, $\text{FR}_{10\%} \downarrow$, and $\text{AUC}_{10\%} \uparrow$ on WFLW~\cite{burgos2013robust}. ``C'' and ``H'' correspond to coordinate/heatmap-regression, respectively. 
\label{tab:main-WFLW}
}
\vspace{-0.3cm}
\end{table}
\begin{table*}[t]
\centering
\begin{tabular}{lccccccc} 
\specialrule{.15em}{.05em}{.05em}
Method & 
Type  & 
WFLW-L & WFLW-E & WFLW-I
& WFLW-M
& WFLW-O
& WFLW-B
\\ 
\cmidrule{1-8}
Wing~\cite{feng2018wing} & C & 8.75 & 5.36 & 4.93 & 5.41 & 6.37 & 5.81 \\
RePFormer~\cite{li2022repformer} & C & 7.25 & \second{4.22} & 4.04 & 3.91 & 5.11 & 4.76 \\
\cmidrule{1-8}
DeCaFA~\cite{dapogny2019decafa} & H & 8.11 & 4.65 & 4.41 & 4.63 & 5.74 & 5.38 \\
AWing~\cite{wang2019awing} & H & 7.38 & 4.58 & 4.32 & 4.27 & 5.19 & 4.96 \\
ADNet~\cite{huang2021adnet} & H & 6.96 & 4.38 & 4.09 & 4.05 & 5.06 & 4.79 \\
HIH~\cite{lan2021hih} & H & 6.87 & \best{4.06} & 4.34 & \second{3.85} & 4.85 & 4.66 \\
STAR~\cite{zhou2023star} & H & \second{6.76} & {4.27} & \second{3.97} & \best{3.83} & \second{4.80} & \second{4.58} \\
\rowcolor{OursColor} Ours & H & \best{6.58} & {4.26} & \best{3.90} & {3.89} &  \best{4.74} & \best{4.57} \\
\specialrule{.15em}{.05em}{.05em}
\end{tabular}
\vspace{-0.05cm}
\caption{
\textbf{Comparison to SOTA facial landmark detection methods on the subsets of WFLW.} We report the inter-ocular NME$\downarrow$ on the six subsets of WFLW~\cite{burgos2013robust},~\ie, large pose (WFLW-L), expression (WFLW-E), illumination (WFLWI), make-up (WFLW-M), occlusion (WFLW-O), and blur (WFLW-B). Type ``C'' and ``H'' stand for coordinate-regression and heatmap-regression methods. }
\label{tab:WFLW-subsets}

\end{table*}

\begin{table*}[t]
\centering
\begin{minipage}[t]{0.48\linewidth}
    \vspace{0pt}
    \setlength{\tabcolsep}{5pt}

\centering
\begin{tabular}{lccccccccc} 
\specialrule{.15em}{.05em}{.05em}
\multicolumn{1}{l}{\multirow{1}{*}{Method}} & 
\multicolumn{1}{l}{\multirow{1}{*}{Type}}  & 
\multicolumn{1}{c}{NME $\downarrow$} & $\text{FR} \downarrow$ & $\text{AUC} \uparrow$ \\ 
\midrule
DAC-CSR~\cite{feng2017dynamic} & C & 6.03 & 4.73 & - \\
\citet{wu2015robust} & C & 5.93 & - & - \\
Wing~\cite{feng2018wing} & C & 5.44 &  3.75 & -\\
DCFE~\cite{valle2018deeply} & C & 5.27 & 7.29 & 0.359 \\
SLPT~\cite{xia2022splt} & C & 4.79 & 1.18 & - \\
\midrule
MHHN~\cite{wan2020robust} & H & 4.95 & 1.78 \\
Awing~\cite{wang2019awing} & H & 4.94  & 0.99 & 0.488 \\
ADNet~\cite{huang2021adnet} & H  & 4.68 & \second{0.59} & 0.532 \\
HIH~\cite{lan2021hih} & H & {4.63} & \best{0.39} & - \\
STAR~\cite{zhou2023star}     & H  & \second{4.62} & {0.79} & 0.540 \\
\rowcolor{OursColor} Ours & H & \best{4.58} & {0.79} & \best{0.544}
\\
\specialrule{.15em}{.05em}{.05em}
\end{tabular}
\caption{
\textbf{Comparison to SOTA on COFW.} ``C'' and ``H'' stand for coordinate-regression and heatmap-regression methods.
}
\label{tab:main-COFW}

\end{minipage}
\hfill
\begin{minipage}[t]{0.48\linewidth}
    \vspace{0pt}
    \setlength{\tabcolsep}{5pt}

\centering
\begin{tabular}{lccccccccc} 
\specialrule{.15em}{.05em}{.05em}
{\multirow{1}{*}{Method}} & 
{\multirow{1}{*}{Type}}  & 
Full & Comm. & Chal. \\ 
\midrule
SLPT~\cite{xia2022splt} & C & 3.17 & 2.75 & 4.90 \\
RePFormer~\cite{li2022repformer} & C & 3.01 & - & - \\
DTLD~\cite{li2022DTLD} & C & 2.96 & 2.59 & {4.50} \\ 
\midrule
DeCaFa~\cite{dapogny2019decafa} & H & 3.39 & 2.93 & 5.26 \\
HRNet~\cite{hrnet} & H & 3.32 & 2.87 & 5.15 \\
HIH~\cite{lan2021hih} & H & 3.09 & 2.65 & 4.89 \\
Awing~\cite{wang2019awing} & H & 3.07 & 2.72 & 4.52 \\
ADNet~\cite{huang2021adnet} & H  & 2.93 & \second{2.53} & 4.58 \\
KeyPosS~\cite{bao2023keyposs} & H & {3.34} & - & - \\
STAR~\cite{zhou2023star}     & H  & \best{2.87} & \best{2.52} & \second{4.32} \\
\rowcolor{OursColor} Ours & H & \best{2.87} & \second{2.53} & \best{4.27} 
\\
\specialrule{.15em}{.05em}{.05em}
\end{tabular}
\caption{
\textbf{Inter-ocular NME$\downarrow$ comparisons on 300W and common/challenging subsets.} 
}
\label{tab:main-300W}

\end{minipage}
\vspace{-0.15cm}
\end{table*}

\subsection{Landmark detection}
{\bf\noindent Experiment setup.} We conduct experiments on commonly used landmark detection benchmarks:
\begin{itemize}
    \item WFLW~\cite{wu2018lab} contains 7,500 training and 2,500 test images, each labeled with 98 facial landmarks. WFLW contains six subsets, including ``large pose'', ``expression'', ``illumination'', ``make-up'', ``occlusion'', and ``blur''.
These subsets identify the more challenging conditions for facial landmark detection.
\item COFW~\cite{burgos-artizzu_perona_dollar_2022} contains 1,345 training and 507 test images, each labeled with 68 facial landmarks. 300W contains 3,148 and 689 test images, each labeled with 29 facial landmarks. 
\item 300W~\cite{sagonas2013300} test images are split into two: a common subset (Comm.) with 554 images and a challenging (Chal.) subset with 135 images under more diverse illumination conditions and variations of expressions and poses.
\end{itemize}

\begin{figure*}[t]
    \begin{subfigure}[h!]{0.32\textwidth}
        \includegraphics[width=\textwidth]{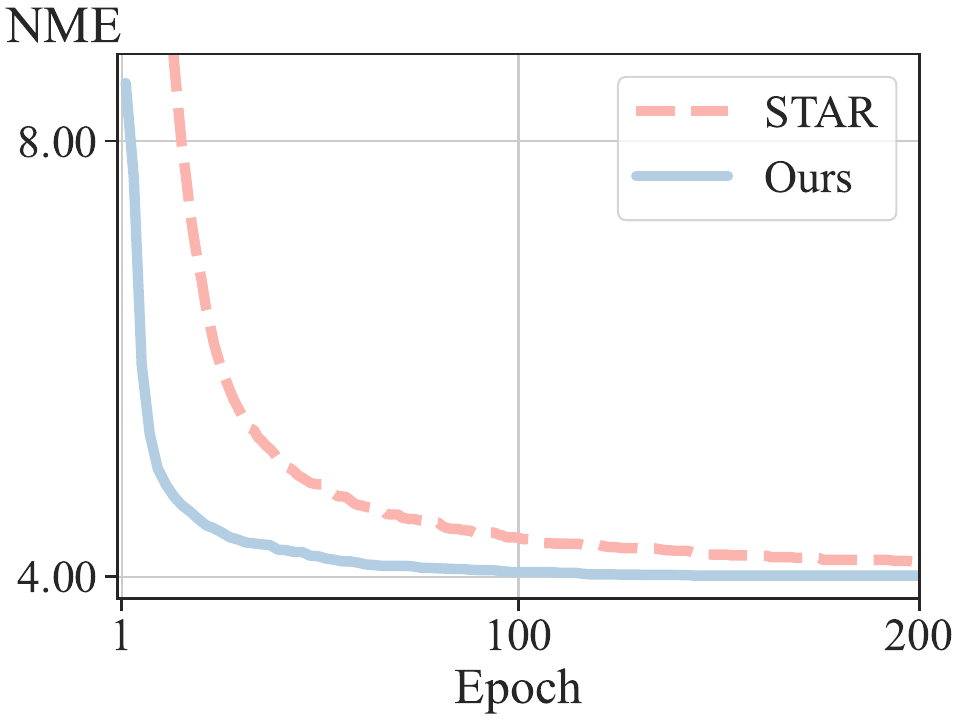}
        \caption{WFLW.}
    \end{subfigure}
    \hfill
    \begin{subfigure}[h!]{0.32\textwidth}
        \includegraphics[width=\textwidth]{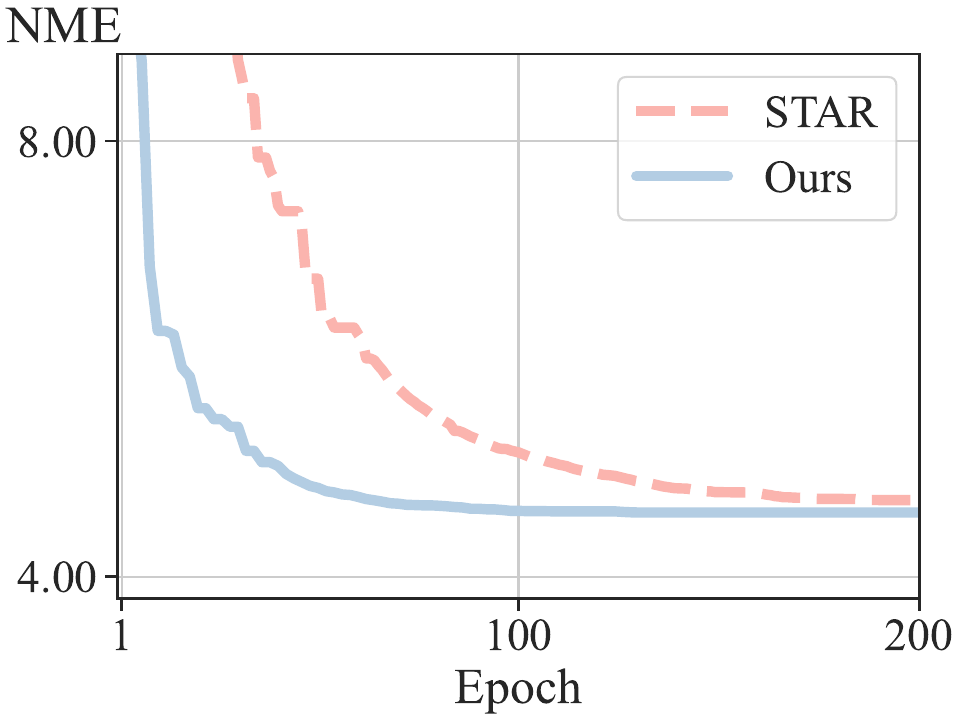}
        \caption{COFW.}
    \end{subfigure}
    \hfill
    \begin{subfigure}[h!]{0.32\textwidth}
        \includegraphics[width=\textwidth]{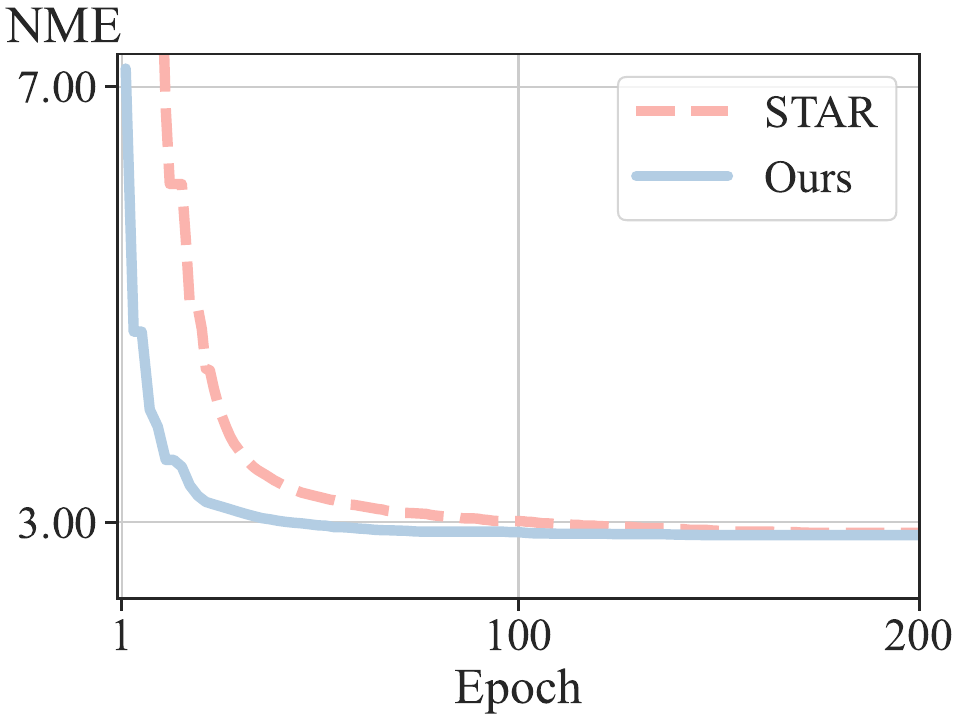}
        \caption{300W.}
    \end{subfigure}
    \vspace{-0.15cm}
  \caption{ \textbf{Comparison of NME curve throughout training between STAR and ours.} We plot the curve of inter-ocular NME$\downarrow$ over training epochs for both methods. The performance improves and converges much faster using our method than using STAR~\cite{zhou2023star}.
  }
  \label{fig:efficiency}
  \vspace{-0.1cm}
\end{figure*}

{\bf\noindent Evaluation metric.} As in prior works,
we report on three metrics: \textit{(a) Normalized Mean Error (NME$\downarrow$), (b) Failure Rate (FR$\downarrow$), and (c) Area Under Curve (AUC$\uparrow$)}. NME is defined as the average $\ell_2$-loss  over
the landmarks:
$
\text{NME}(\vy_n, \hat \vy_n) \triangleq \frac{1}{N\cdot d} \sum_{n=1}^N \| \vy_n - \hat \vy_n \|_2,
$
where $d$ is the distance used for normalizing the error. 

The FR is the ratio of ``failed'' cases, where a case is considered to fail if its NME exceeds a given threshold. 
The AUC measures the area under the Cumulative Error Distribution (CED) curve from zero to a given NME threshold. 
Following prior works~\cite{zhou2023star, huang2021adnet}, we use inter-ocular distance as $d$ for both WFLW and 300W and inter-pupil distance for COFW. The FR and AUC threshold is set to 10\% for WFLW and COFW, and 5\% for 300W.

{\noindent\bf Implementation details.} 
As we are innovating on the loss function, we strictly follow the data preparation and the deep-net architecture from STAR by~\citet{zhou2023star}. 
During training, each input image first
has the face region cropped out and resized to 256px $\times$ 256px. The image is then processed through the following random data augmentations, including rotation, scaling, brightness, blur, cutout, and horizontal flip.

For the deep-net architecture, we use the same four-stacked hourglass architecture~\cite{zhou2023star} to predict the heatmap. We replaced their {\tt Soft-argmax} layer along with {\tt ReLU} which they used to normalize the heatmap. Note that STAR~\cite{zhou2023star} uses the Anisotropic Attention Module (AAM), an attention mask to guide landmark prediction. The AAM is trained with an auxiliary task of predicting edge heatmaps derived from the boundary lines. We also included such auxiliary tasks in our training. 
Additional implementation details and hyperparameters are documented in the Appendix~\secref{supp:details}.

{\bf\noindent Quantitative results.} 
For baselines, we compare to the SOTA facial landmark detection methods with available code or reproducibility.
In~\tabref{tab:main-WFLW}, we report the quantitative comparison on the full test set of WFLW. The best result is {\bf bolded} and the second best is \underline{underlined}. Our method achieves state-of-the-art performance under all evaluation metrics. More notably, we outperform STAR~\cite{zhou2023star} by 0.02 on NME and 0.32 on FR, which are comparable to the gains achieved by prior works. Note, %
this result is significant. 
We trained our model three times and the NME's standard deviation is $0.003$. 

In~\tabref{tab:WFLW-subsets}, we report the results on the six more challenging subsets of WFLW. We achieve state-of-the-art on 4 subsets while being competitive on the remaining 2 subsets. Notably, our method outperforms STAR~\cite{zhou2023star} the most under the ``large-pose'' category (WFLW-L), where the input facial images have larger motion and the poses are far from frontal. We achieve an NME improve of 0.18. 

In~\tabref{tab:main-COFW}, we report the comparison between baselines and ours on COFW. Our method achieves state-of-the-art performance in terms of NME and AUC on COFW. 

In~\tabref{tab:main-300W}, we report the performance under the subsets of 300W. Again, our method remains competitive to SOTA under all categories. Additional results are in ~\secref{supp:details}. 

\begin{table}[t]
    \centering
    \renewcommand{\arraystretch}{0}
    \begin{tabular}{c@{\hskip 1pt}c c@{\hskip 1pt}c}
        \multicolumn{2}{c}{STAR} & \multicolumn{2}{c}{Ours} \\
        {Epoch 1} & {Epoch 5} & {Epoch 1} & {Epoch 2} \\
        \includegraphics[width=0.23\linewidth]{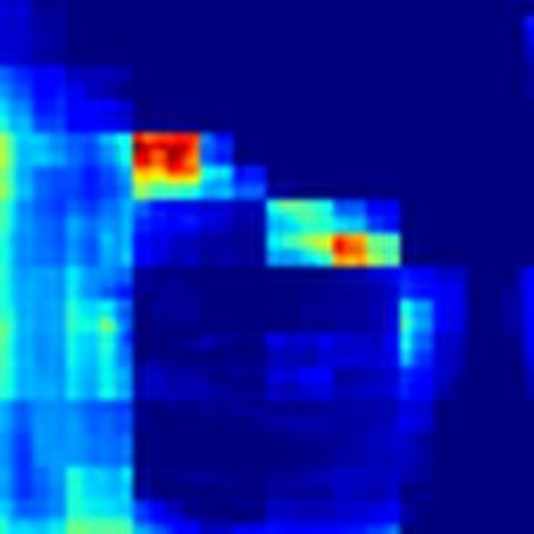} &
        \includegraphics[width=0.23\linewidth]{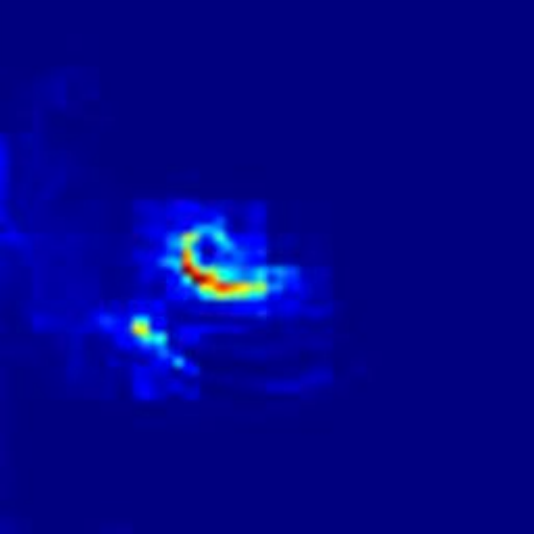} &
        \includegraphics[width=0.23\linewidth]{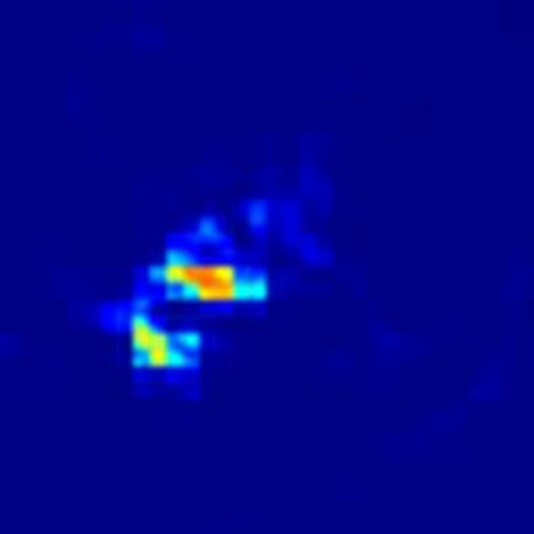} &
        \includegraphics[width=0.23\linewidth]{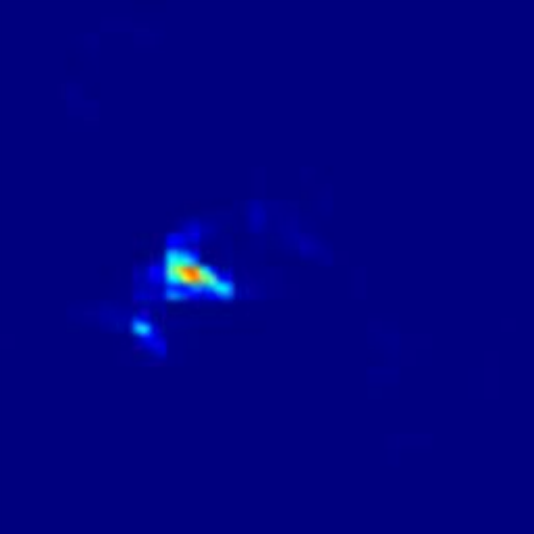} \\
        \vspace{3pt} \\
        \includegraphics[width=0.23\linewidth]{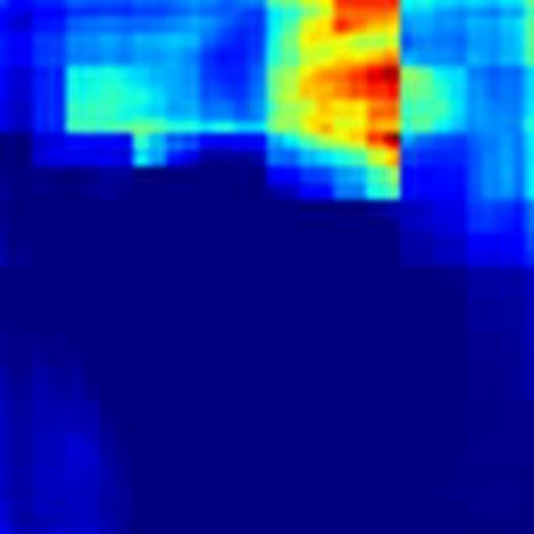} &
        \includegraphics[width=0.23\linewidth]{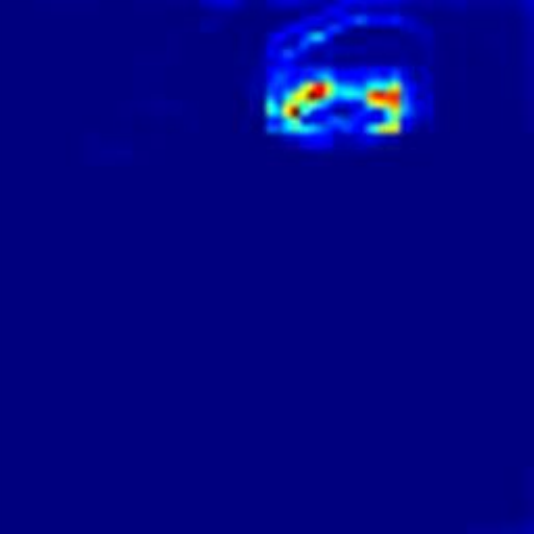} &
        \includegraphics[width=0.23\linewidth]{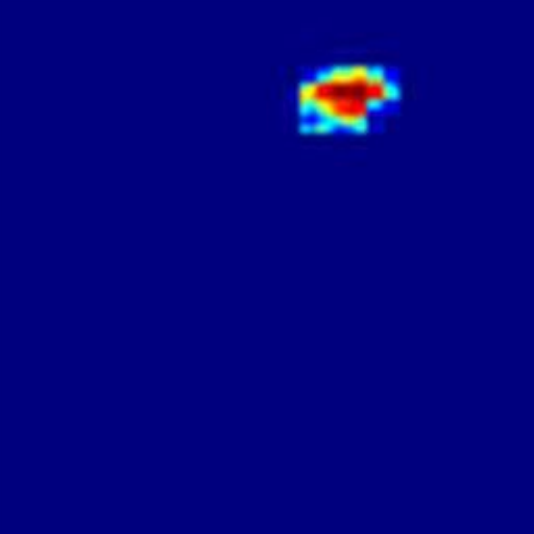} &
        \includegraphics[width=0.23\linewidth]{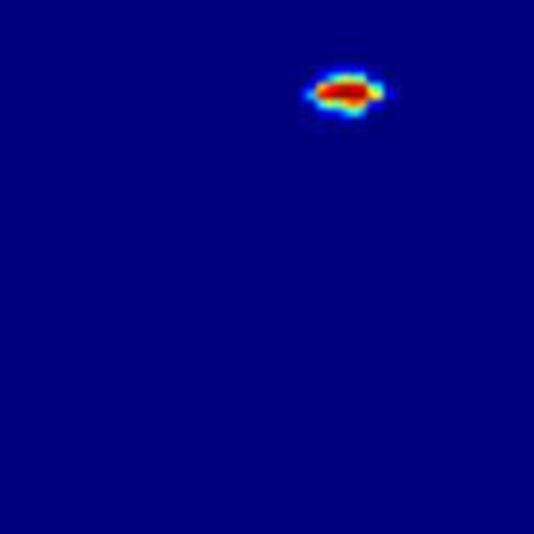} \\
        \vspace{3pt} \\
        \includegraphics[width=0.23\linewidth]{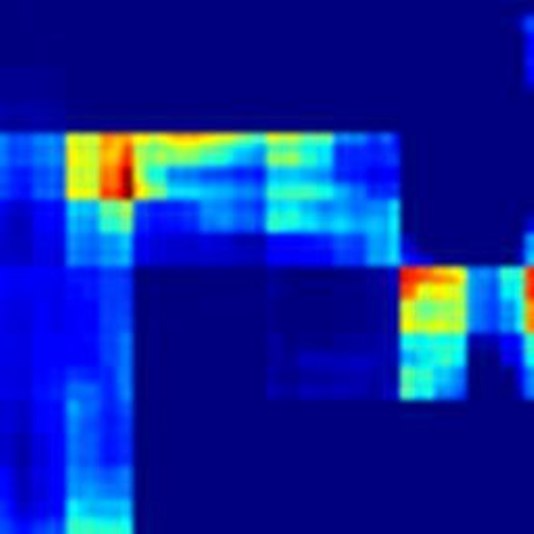} &
        \includegraphics[width=0.23\linewidth]{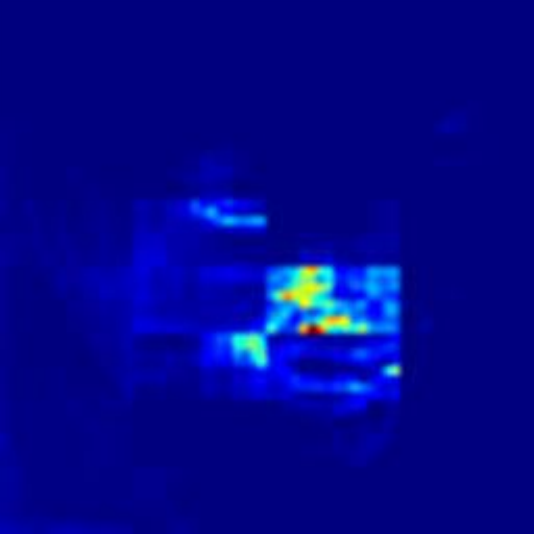} &
        \includegraphics[width=0.23\linewidth]{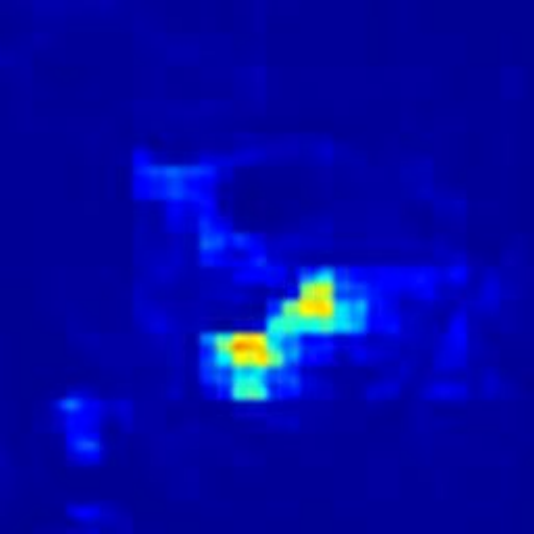} &
        \includegraphics[width=0.23\linewidth]{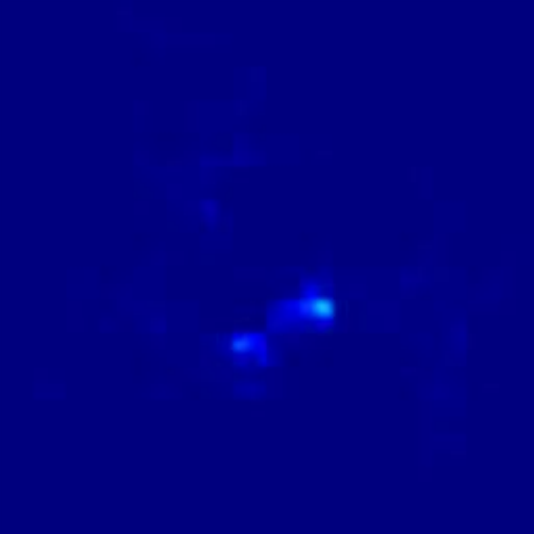} 
    \end{tabular}
    \vspace{-0.05cm}
    \captionof{figure}{
        \textbf{Comparison of convergence efficiency between STAR's {\tt Soft-argmax} and ours.} The first two columns visualize the $\hat \mH$ trained with STAR~\cite{zhou2023star}'s {\tt Soft-argmax} at epochs 1 and 5. The last two columns visualize the $\hat \mH$ trained with our loss at epochs 1 and 2. Each row corresponds to a different sample.
    }
    \label{fig:2d_vis}
    \vspace{-0.1cm}
\end{table}

\begin{figure*}[t]
\centering
    \includegraphics[width=0.91\textwidth]{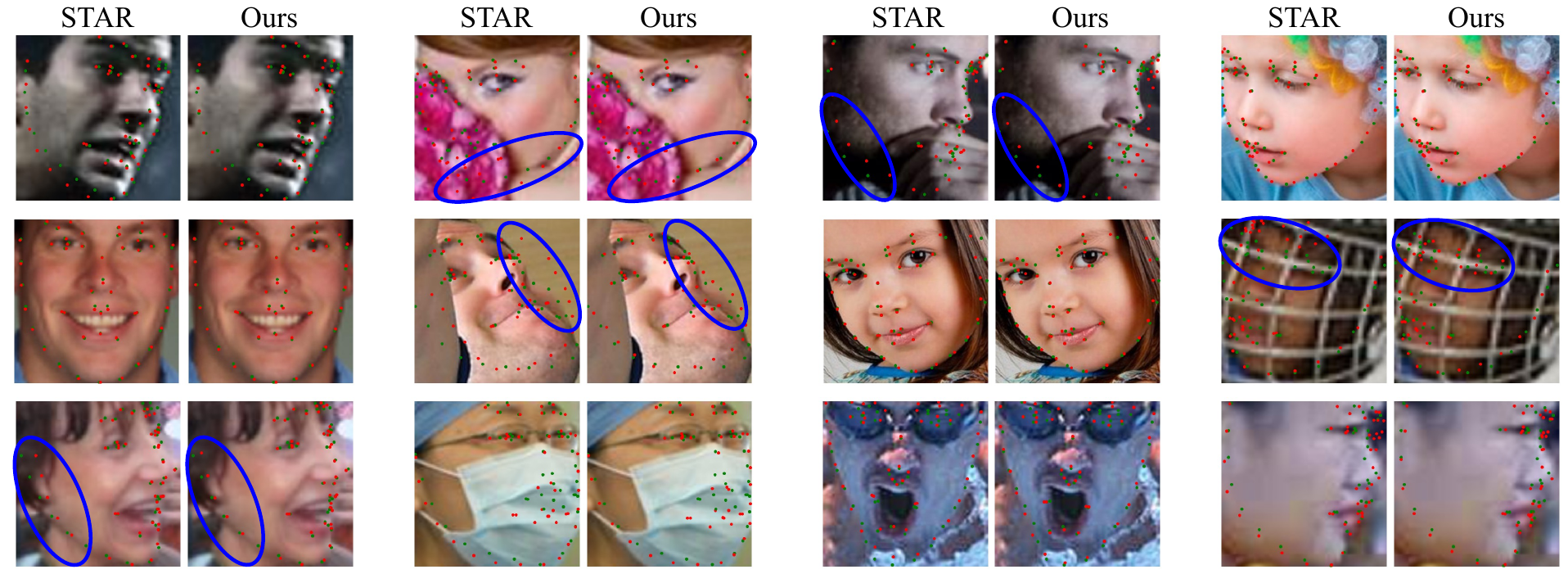}
  \vspace{-0.1cm}
  \caption{\textbf{Qualitative comparison on WFLW between STAR and ours.} The ground truth landmarks are marked in \darkgreen{green} while the predictions are marked in \textcolor{red}{red}. The regions highlighted in \textcolor{blue}{blue} circles emphasize where our method outperforms STAR~\cite{zhou2023star}.
  }
  \vspace{-0.05cm}
  \label{fig:qual-WFLW}
\end{figure*}

\begin{table*}[t!]
    \begin{minipage}[t]{0.45\linewidth}
            \centering
    \begin{tabular}[t]{c c c c}
        \specialrule{.15em}{.05em}{.05em}
        $\Delta$ & {NME$\downarrow$} & {FR$\downarrow$} & {AUC$\uparrow$}
        \\
        \cmidrule(lr){1-4}
        None & 4.01 & {2.16} & {0.605} \\ 
        $\ell_2$ & 4.00 & 2.00 & \best{0.606} \\
        $\ell_1$ & 4.00 & 2.04 & \best{0.606} \\ 
        \rowcolor{OursColor} smooth-$\ell_1$ & \best{3.99} & \best{1.84} & \best{0.606} \\ 
        \specialrule{.15em}{.05em}{.05em}
    \end{tabular}
     \captionof{table}{{\bf Ablation on the choice of margin $\Delta$.} Smooth-$\ell_1$ achieves the best performance.
    }
    \label{tab:ablation-task}

    \end{minipage}
    \hfill
    \begin{minipage}[t]{0.5\linewidth}
        \centering
\begin{tabular}[t]{c c c c c}
    \specialrule{.15em}{.05em}{.05em}
    Label Smoothing & Loss & {NME.$\downarrow$} & {FR$\downarrow$} & {AUC$\uparrow$}
    \\
    \cmidrule(lr){1-5}
    \ding{55} & STAR & 4.02 & 2.32 & 0.605 \\ 
    \ding{55} & Eq.~\ref{eq:loss-augmented}  & 4.02 & 2.23 & 0.604\\ 
    image-unaware & Eq.~\ref{eq:loss-augmented} & \best{3.99} & \second{2.12} & \best{0.606} \\ 
    image-aware & STAR & 4.00 & 2.12 & 0.602 \\ 
    \rowcolor{OursColor} 
    image-aware & Eq.~\ref{eq:loss-augmented} & \best{3.99} & \best{1.84} & \best{0.606} \\ 
    \specialrule{.15em}{.05em}{.05em}
\end{tabular}
    \captionof{table}{\textbf{Ablation on structure prediction and label smoothing.} 
}
\label{table:ablation-smooth}

    \end{minipage}
    \vspace{-0.2cm}
\end{table*}

{\bf\noindent Training convergence.}
In~\figref{fig:efficiency}, we visualize the learning efficiency of our method in comparison to STAR~\cite{zhou2023star}. We plot the training curve,~\ie, NME versus training epochs, on WFLW, COFW, and 300W.  Consistently across these experiments, we observe a much faster convergence speed with our method. Roughly speaking, our loss converges $2.2\times$ faster than STAR. For example, it takes STAR 44 epochs to achieve an NME of 4.50 on WFLW and ours to achieve the same performance in 20 epochs. 

In~\figref{fig:2d_vis}, we further demonstrate the convergence limitations of \texttt{Soft-Argmax}, we visualize the predictions $\hat \mH$ trained with STAR's \texttt{Soft-Argmax} versus ours on WFLW. We observe the same behavior as in \figref{fig:toy}. STAR's \texttt{Soft-Argmax} still suffers from bi-modal prediction at epoch $t=5$, while ours quickly finds the correct peak in 2 epochs. More comparisons are provided in the Appendix~\ref{supp:analysis-softargmax}.

{\bf\noindent Qualitative results.} In~\figref{fig:qual-WFLW}, we provide a qualitative comparison between STAR and Ours.
We observe comparable results in the predicted key points. In particular, we observe slight benefits along the facial contours,~\eg, jaw lines, and in more challenging cases involving occlusions.

\subsection{Ablation studies}
{\bf\noindent Choices for the margin term $\Delta$.} In~\tabref{tab:ablation-task}, we report the performance of other common distance functions in facial landmark detection as the margin term $\Delta$. We observe that the choices of $\Delta$ mildly impact the final performance. 
We note that by removing the margin term,~\equref{eq:loss-augmented} can be reduced to cross-entropy loss with a temperature term. 
Overall, our choice of $\Delta$ of smooth-$\ell_1$~\cite{girshick2015fast} has the best performance. %

{\bf\noindent Structure prediction and label smoothing.}
In~\tabref{table:ablation-smooth}, we report additional ablation studies on our loss. 
We consider different configurations for label smoothing,~\ie, no label smoothing, image-unaware smoothing, and image-aware smoothing. We also consider different configurations for loss design,~\ie,  STAR~\cite{zhou2023star} and ours based solely on structured prediction in~\equref{eq:loss-augmented}.
Specifically, the rows represent the following configurations: (a) no contribution; (b) only structured prediction; (c) only image-unaware label smoothing; (d) only image-aware label smoothing; (e) both structured prediction and image-aware label smoothing.

Without label smoothing, our loss achieves comparable performance to STAR~\cite{zhou2023star} on WFLW.
We can also observe improvement when applying label smoothing on STAR. 
We also observe a consistent gain in deploying label smoothing under all evaluation metrics and being image-aware further improves performance.
Finally, the best performance is achieved by incorporating all of our contributions.

Importantly, by incorporating image-aware label smoothing, we outperform without label smoothing by 0.02 on NME, 0.23 on FR, and 0.003 on AUC. Validating its effectiveness.

{\it Limitations:} While the proposed label smoothing works well, its design is hand-crafted and more of a heuristic based on our observation. In the future, we aim to explore data-driven smoothing approaches and study how to better quantify the uncertainty.

\section{Conclusion}
In this work, we re-examine and analyze the widely used {\tt Soft-argmax} in recent heatmap regression methods for facial landmark detection.
Instead, we propose to formulate the training objective based on deep structured prediction, for which we can more effectively train the model without {\tt Soft-argmax}. To further address semantic ambiguity, we propose an image-aware label-smoothing technique. With these two components, we arrived at a model with a principled training objective. Empirically, our method obtained state-of-the-art performance on three common facial landmark detection datasets with $2.2\times$ faster training convergence. 
We hope this work will inspire the revisiting of popular design choices in other computer vision tasks through the lens of structured prediction, which may lead to cleaner and more intuitive designs. 

\vspace{3pt}
{\noindent\bf Acknowledgments.} 
RAY is thankful for the ECE544 course at UIUC, taught by Alex Schwing, which provided the foundational concepts for this work.

\clearpage
{
    \small
    \bibliographystyle{ieeenat_fullname}
    \bibliography{main}
}

\newpage
\onecolumn

{\bf\noindent \Large Appendix}\\

\setcounter{section}{0}
\renewcommand{\theHsection}{A\arabic{section}}
\renewcommand{\thesection}{A\arabic{section}}
\renewcommand{\thetable}{A\arabic{table}}
\setcounter{table}{0}
\setcounter{figure}{0}
\renewcommand{\thetable}{A\arabic{table}}
\renewcommand\thefigure{A\arabic{figure}}
\renewcommand{\theHtable}{A.Tab.\arabic{table}}%
\renewcommand{\theHfigure}{A.Abb.\arabic{figure}}%
\renewcommand\theequation{A\arabic{equation}}
\renewcommand{\theHequation}{A.Abb.\arabic{equation}}%

\noindent The appendix is organized as follows:
\begin{itemize}[topsep=0pt, leftmargin=16pt]
\item In~\secref{supp:details}, we provide additional analysis and implementation details of our method.
\item In~\secref{supp:results}, we provide additional results on WFLW~\cite{wu2018lab}, COFW~\cite{burgos-artizzu_perona_dollar_2022}, and 300W~\cite{sagonas2013300}.
\end{itemize}

\section{Additional Details}
\label{supp:details}

\subsection{Additional discussion on Soft-argmax}
\label{supp:analysis-softargmax}

As mentioned in~\secref{sec:limit}, our work focuses on solving the problem of mismatch with Soft-argmax.
The mismatch occurs because the loss is not convex \wrt the heatmap's elements; as illustrated in the example of~\equref{eq:example} and~\equref{eq:example2}. Our proposed method is based on the structured learning framework of loss-augmented inference~\cite{taskar2003max, tsochantaridis2005large, iccv2015schwing}, where the loss is in the form of a log-sum-of-exponentials, which is convex \wrt the heatmap's elements. 
In other words, there will be no local minimum with respect to the heatmap, and hence, no mismatch. 
In Fig.~\ref{fig:more_2d_vis}, we visualize more examples to show the convergence efficiency comparison between STAR and ours.
\begin{table}[h]
    \centering
    \begin{tabular}{
        c@{\hskip 1pt}c @{\hskip 3pt}
        c@{\hskip 1pt}c
        c@{\hskip 1pt}c @{\hskip 3pt} 
        c@{\hskip 1pt}c
    }
        \multicolumn{2}{c}{STAR} & \multicolumn{2}{c}{Ours \quad} &
        \multicolumn{2}{c}{STAR} & \multicolumn{2}{c}{Ours \quad}
        \\
        {Epoch 1} & {Epoch 5} & {Epoch 1} & {Epoch 2} & {Epoch 1} & {Epoch 5} & {Epoch 1} & {Epoch 2} 
        \\
        \begin{minipage}{0.11\linewidth}
            \includegraphics[width=\linewidth]{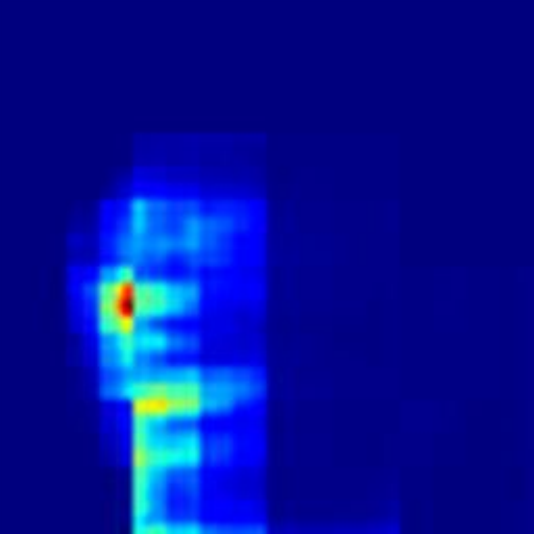}
        \end{minipage} &
        \begin{minipage}{0.11\linewidth}
            \includegraphics[width=\linewidth]{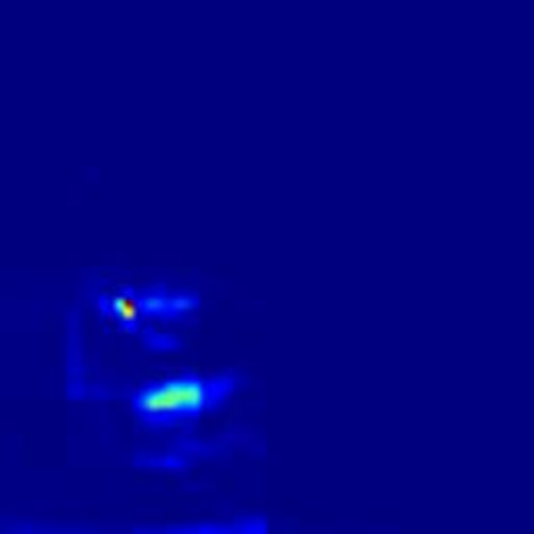}
        \end{minipage} &
        \begin{minipage}{0.11\linewidth}
            \includegraphics[width=\linewidth]{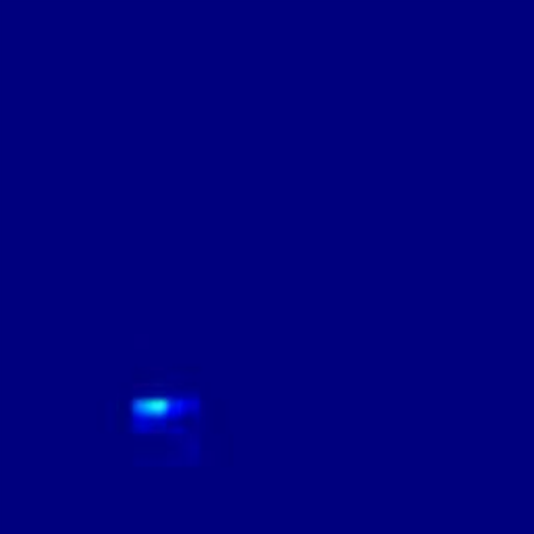}
        \end{minipage} &
        \begin{minipage}{0.11\linewidth}
            \includegraphics[width=\linewidth]{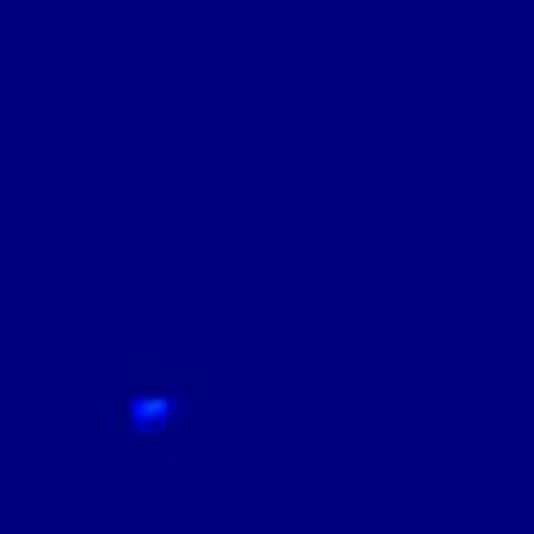}
        \end{minipage} 
        &
        \begin{minipage}{0.11\linewidth}
            \includegraphics[width=\linewidth]{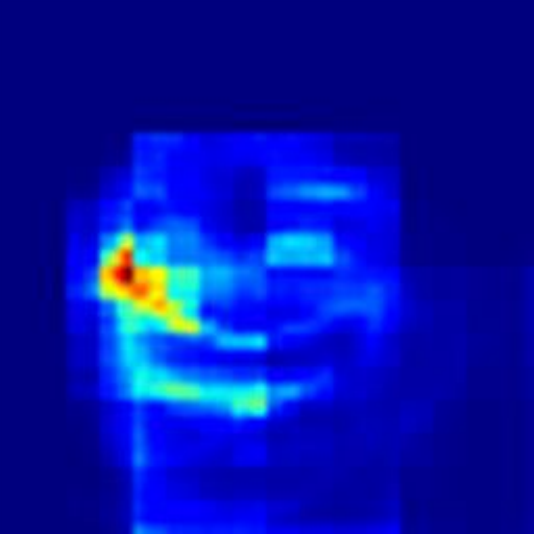}
        \end{minipage} &
        \begin{minipage}{0.11\linewidth}
            \includegraphics[width=\linewidth]{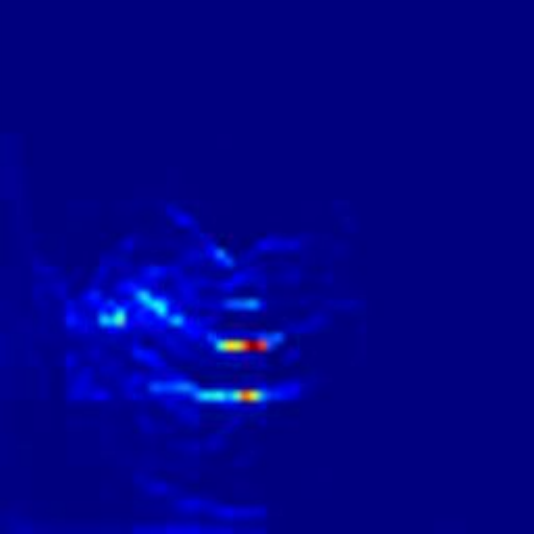}
        \end{minipage} &
        \begin{minipage}{0.11\linewidth}
            \includegraphics[width=\linewidth]{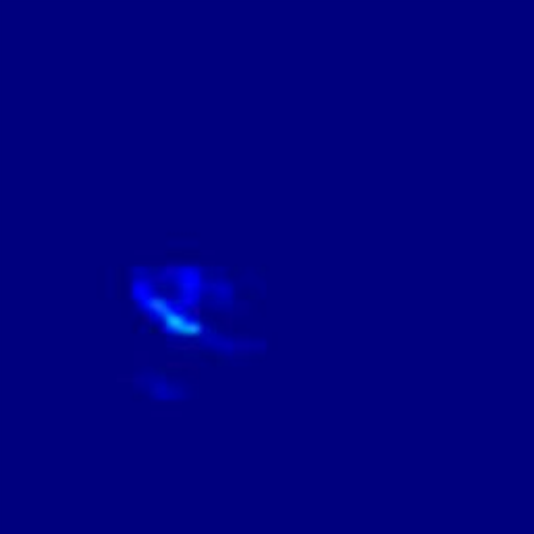}
        \end{minipage} &
        \begin{minipage}{0.11\linewidth}
            \includegraphics[width=\linewidth]{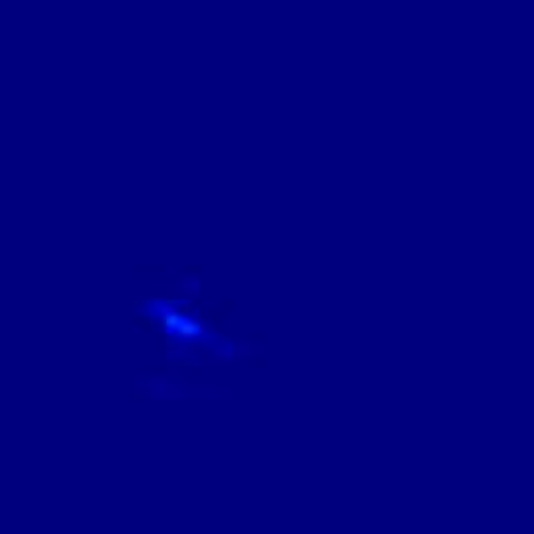}
        \end{minipage} 
        \vspace{8pt}
        \\
        \begin{minipage}{0.11\linewidth}
            \includegraphics[width=\linewidth]{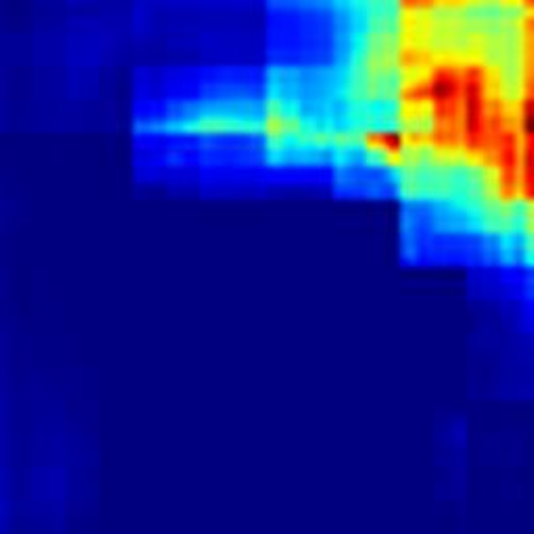}
        \end{minipage} &
        \begin{minipage}{0.11\linewidth}
            \includegraphics[width=\linewidth]{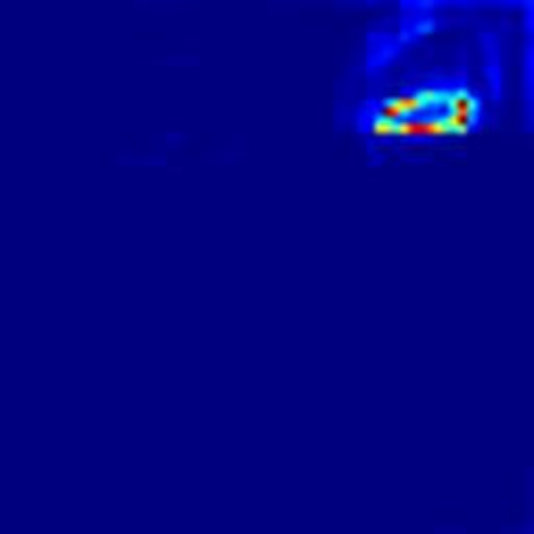}
        \end{minipage} &
        \begin{minipage}{0.11\linewidth}
            \includegraphics[width=\linewidth]{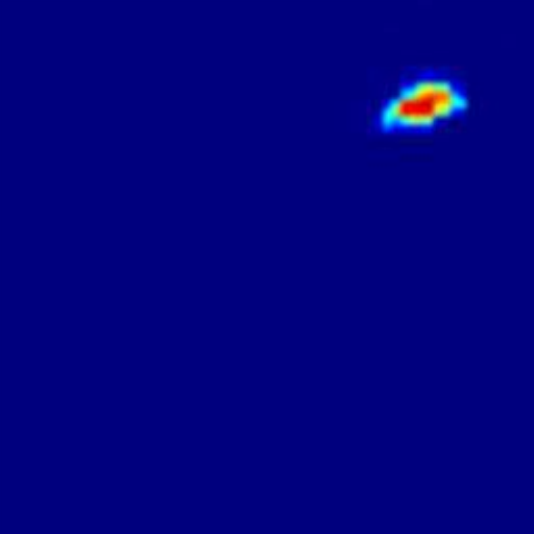}
        \end{minipage} &
        \begin{minipage}{0.11\linewidth}
            \includegraphics[width=\linewidth]{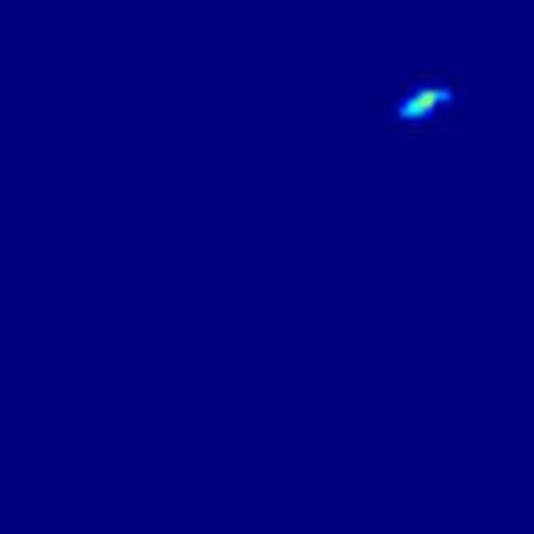}
        \end{minipage} 
        &
        \begin{minipage}{0.11\linewidth}
            \includegraphics[width=\linewidth]{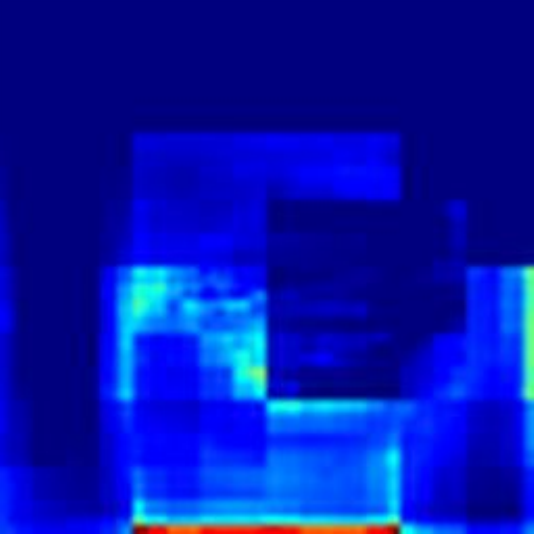}
        \end{minipage} &
        \begin{minipage}{0.11\linewidth}
            \includegraphics[width=\linewidth]{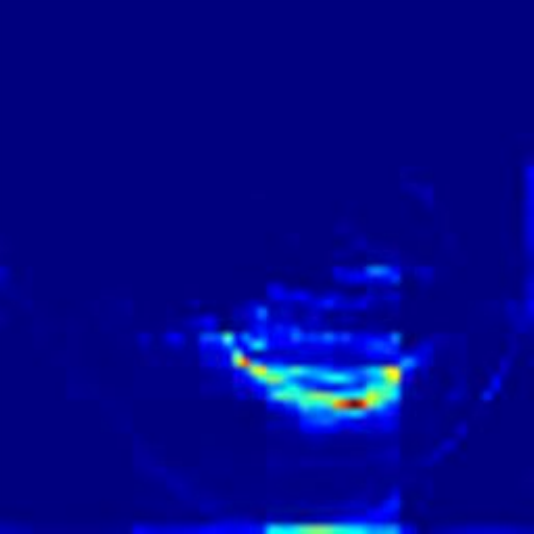}
        \end{minipage} &
        \begin{minipage}{0.11\linewidth}
            \includegraphics[width=\linewidth]{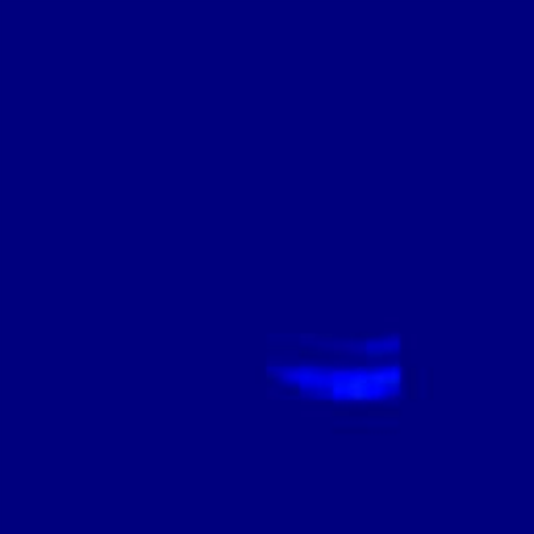}
        \end{minipage} &
        \begin{minipage}{0.11\linewidth}
            \includegraphics[width=\linewidth]{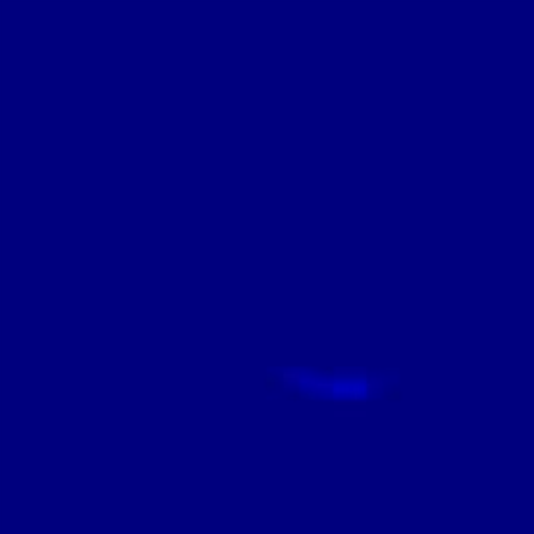}
        \end{minipage} 
        \vspace{8pt}
        \\
        \begin{minipage}{0.11\linewidth}
            \includegraphics[width=\linewidth]{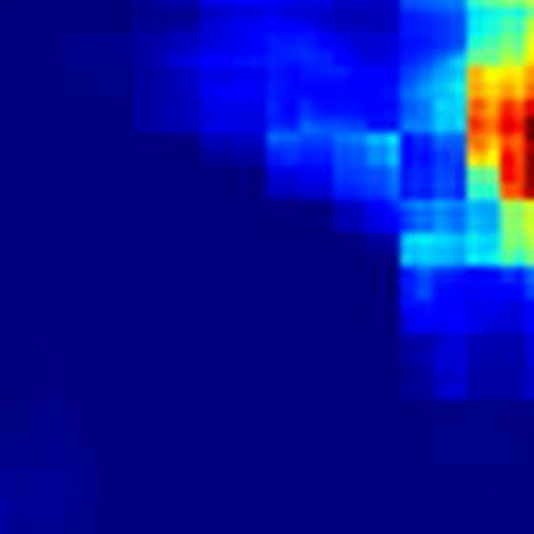}
        \end{minipage} &
        \begin{minipage}{0.11\linewidth}
            \includegraphics[width=\linewidth]{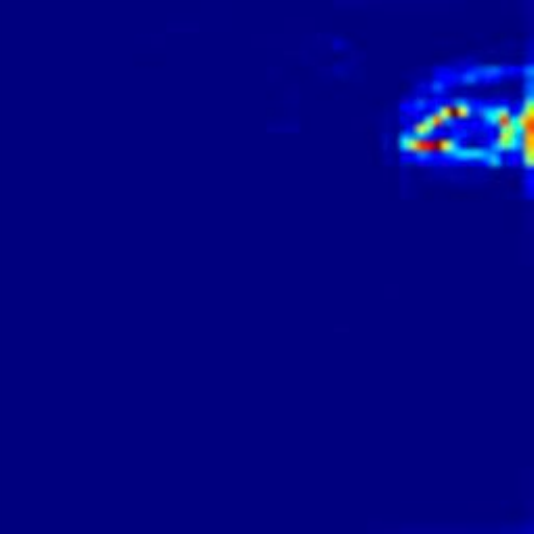}
        \end{minipage} &
        \begin{minipage}{0.11\linewidth}
            \includegraphics[width=\linewidth]{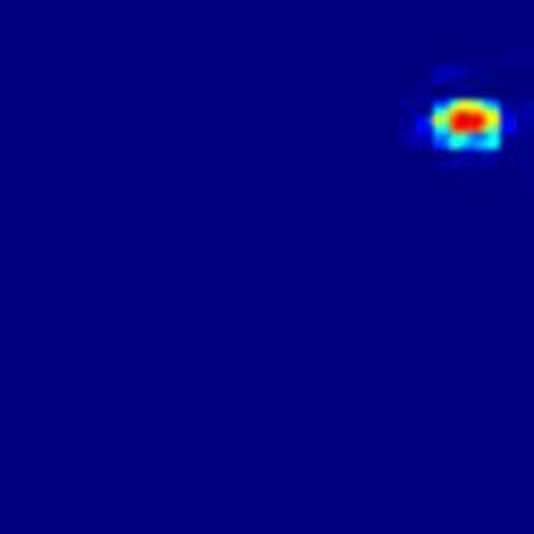}
        \end{minipage} &
        \begin{minipage}{0.11\linewidth}
            \includegraphics[width=\linewidth]{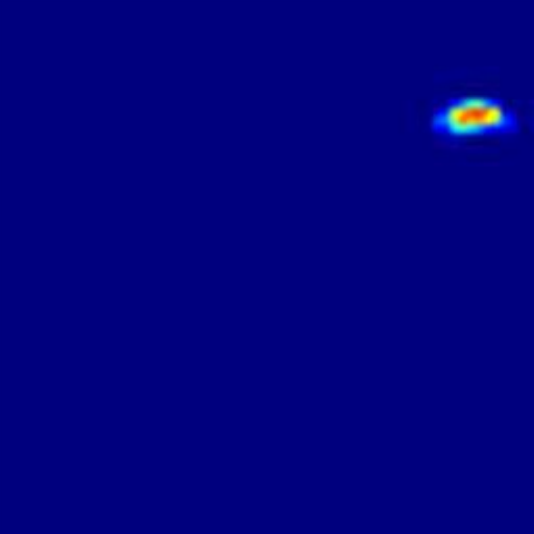}
        \end{minipage} 
        &
        \begin{minipage}{0.11\linewidth}
            \includegraphics[width=\linewidth]{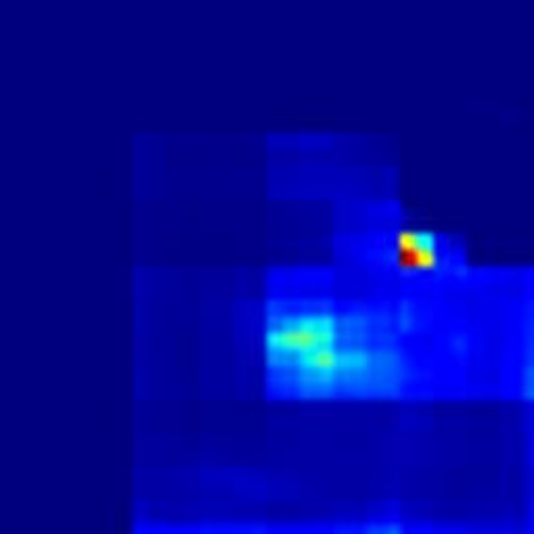}
        \end{minipage} &
        \begin{minipage}{0.11\linewidth}
            \includegraphics[width=\linewidth]{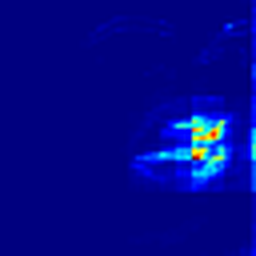}
        \end{minipage} &
        \begin{minipage}{0.11\linewidth}
            \includegraphics[width=\linewidth]{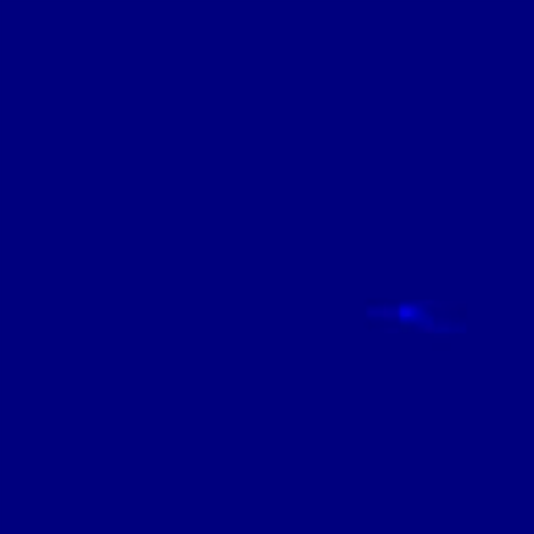}
        \end{minipage} &
        \begin{minipage}{0.11\linewidth}
            \includegraphics[width=\linewidth]{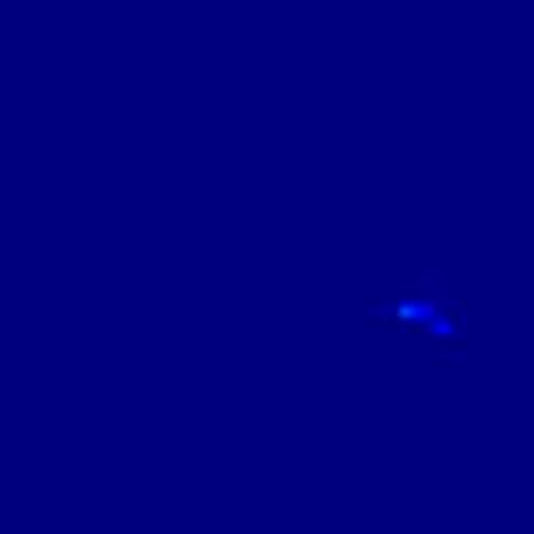}
        \end{minipage} 
    \end{tabular}
    \captionof{figure}{
        \textbf{More comparisons of convergence efficiency between STAR's {\tt Soft-argmax} and ours} following the settings in Fig.~\ref{fig:2d_vis}.
    }
    \label{fig:more_2d_vis}
\end{table}

\subsection{Additional justification for label smoothing}
In~\figref{fig:label_smoothing}, we demonstrate the pipeline of our image-aware label smoothing, which creates a ``soft ground-truth''. To verify that this softness resembles semantic ambiguity, in~\figref{fig:ls-human}, we ask 5 users to annotate facial images following a similar process of WFLW~\cite{wu2018lab}. As expected, the annotations consist of variations. Importantly, we observe a strong similarity between our label smoothing and the distribution of human annotations. This further confirms our assumption that the variation along the image edges is correct.
\begin{table}[h!]
    \centering
    \small
    \begin{tabular}{
        c@{\hskip 1pt}c @{\hskip 3pt}
        c@{\hskip 1pt}c
        c@{\hskip 1pt}c @{\hskip 3pt}
        c@{\hskip 1pt}c}
        \multicolumn{2}{c}{Human}
        &
        \multicolumn{2}{c}{Ours \quad}
        &
        \multicolumn{2}{c}{Human}
        &
        \multicolumn{2}{c}{Ours  \quad}
        \\
        \begin{minipage}{0.11\linewidth}
            \includegraphics[width=\linewidth, height=\textwidth]{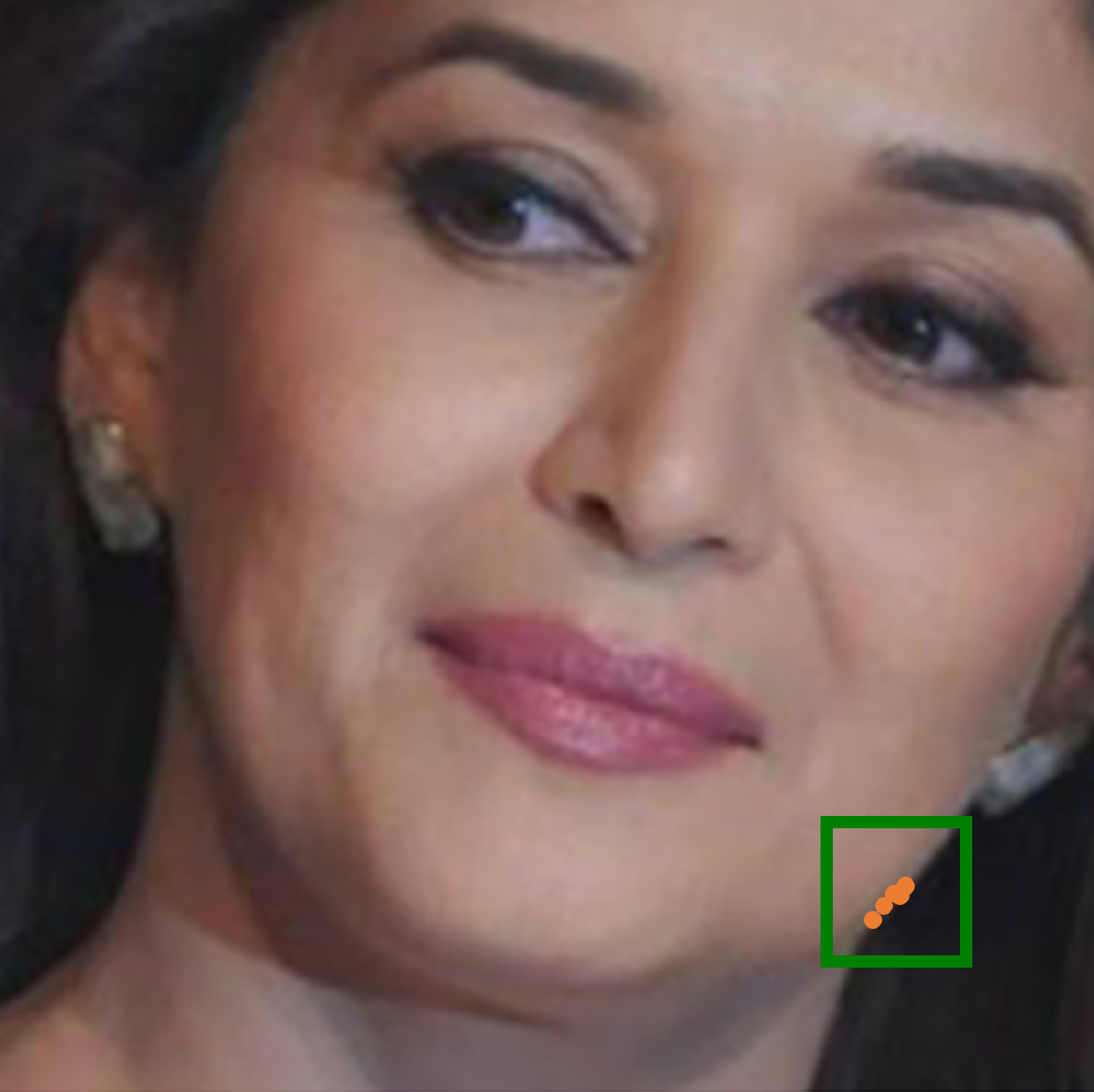}
        \end{minipage} &
        \begin{minipage}{0.11\linewidth}
            \includegraphics[width=\linewidth, height=\textwidth]{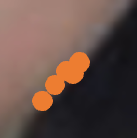}
        \end{minipage} &
        \begin{minipage}{0.11\linewidth}
            \includegraphics[width=\linewidth, height=\textwidth]{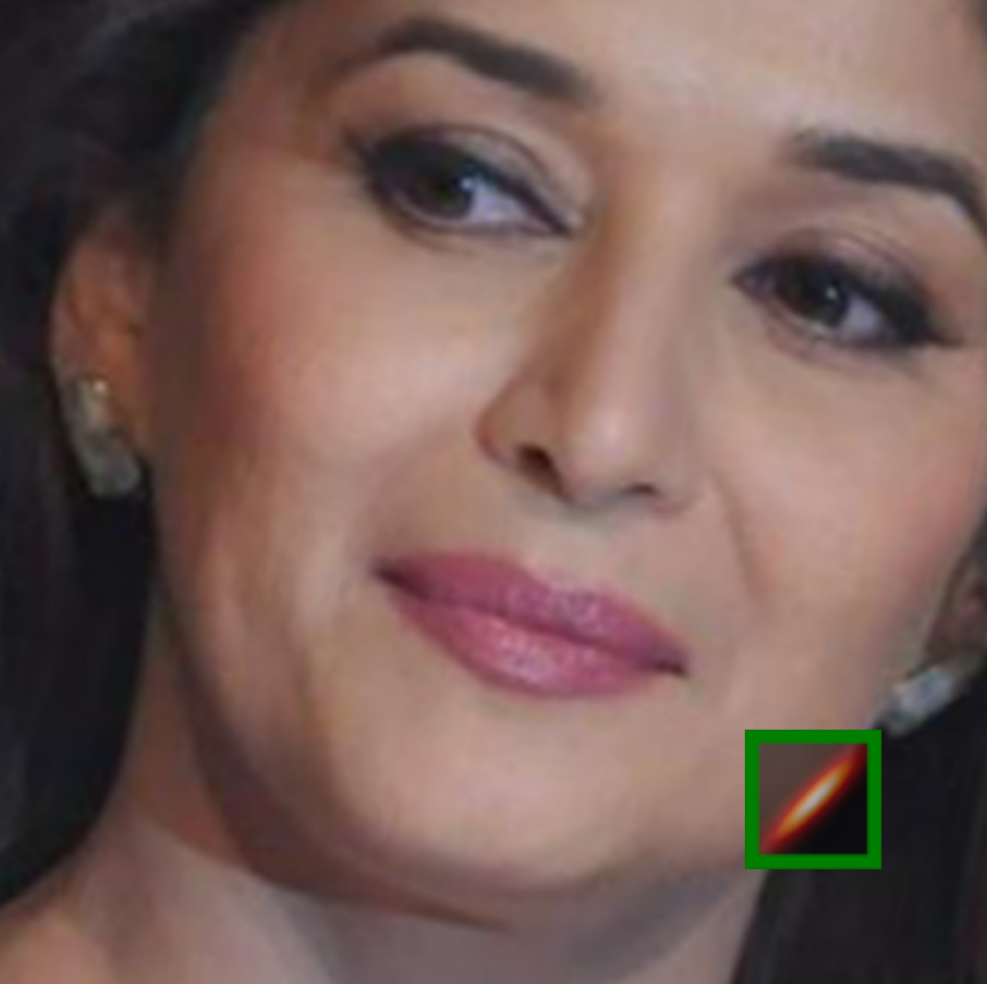}
        \end{minipage} &
        \begin{minipage}{0.11\linewidth}
            \includegraphics[width=\linewidth, height=\textwidth]{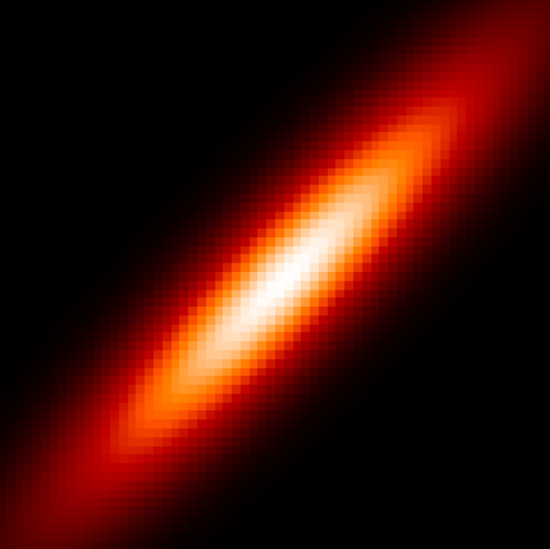}
        \end{minipage} 
        &
        \begin{minipage}{0.11\linewidth}
            \includegraphics[width=\linewidth, height=\textwidth]{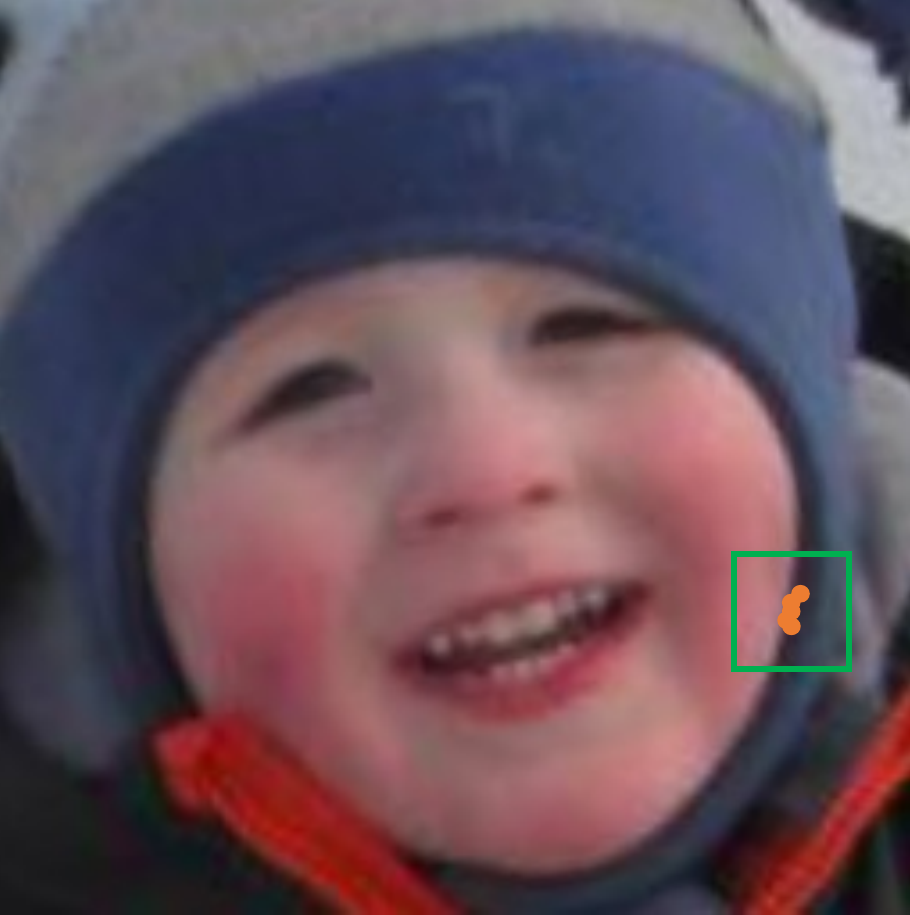}
        \end{minipage} &
        \begin{minipage}{0.11\linewidth}
            \includegraphics[width=\linewidth, height=\textwidth]{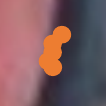}
        \end{minipage} &
        \begin{minipage}{0.11\linewidth}
            \includegraphics[width=\linewidth, height=\textwidth]{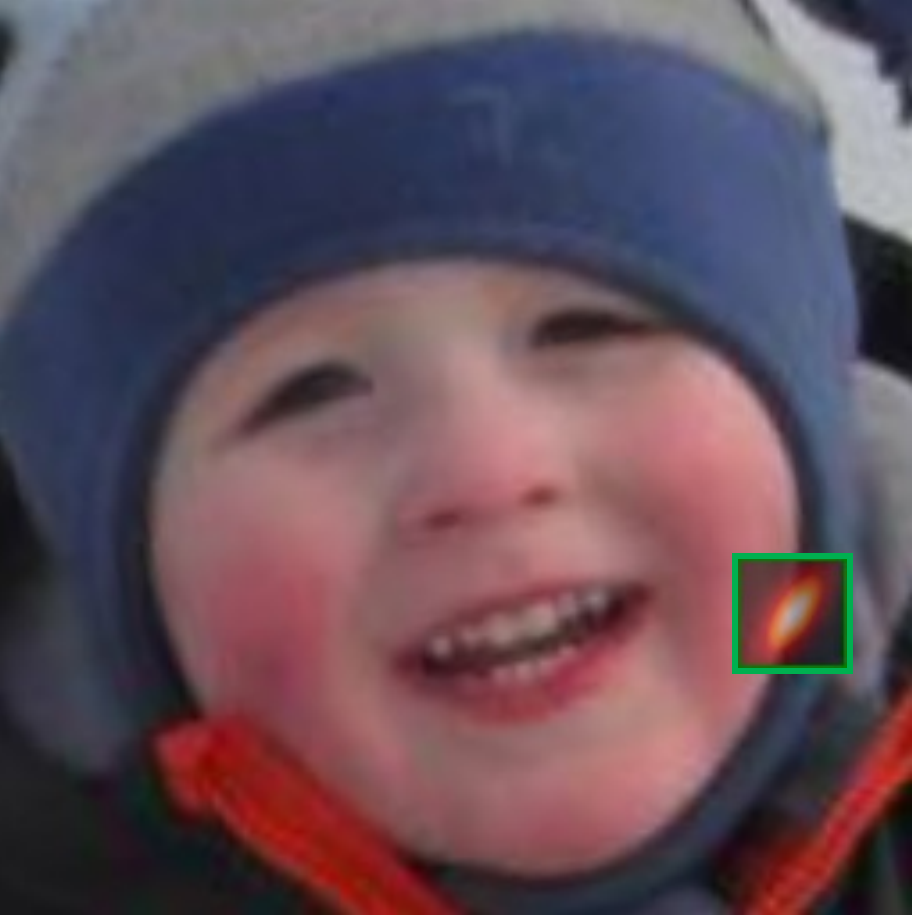}
        \end{minipage} &
        \begin{minipage}{0.11\linewidth}
            \includegraphics[width=\linewidth, height=\textwidth]{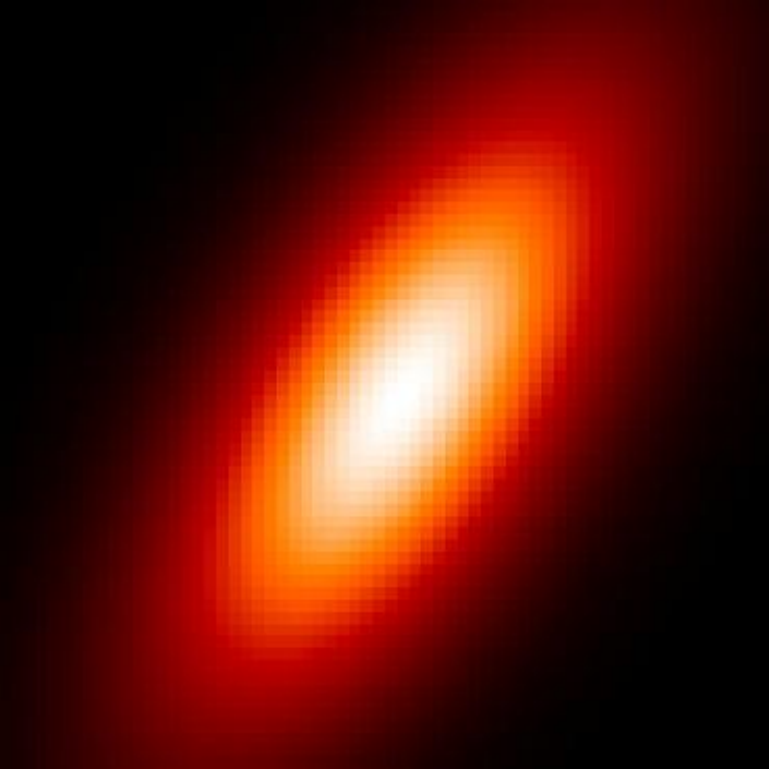}
        \end{minipage} 
        \vspace{8pt}
        \\
        \begin{minipage}{0.11\linewidth}
            \includegraphics[width=\linewidth, height=\textwidth]{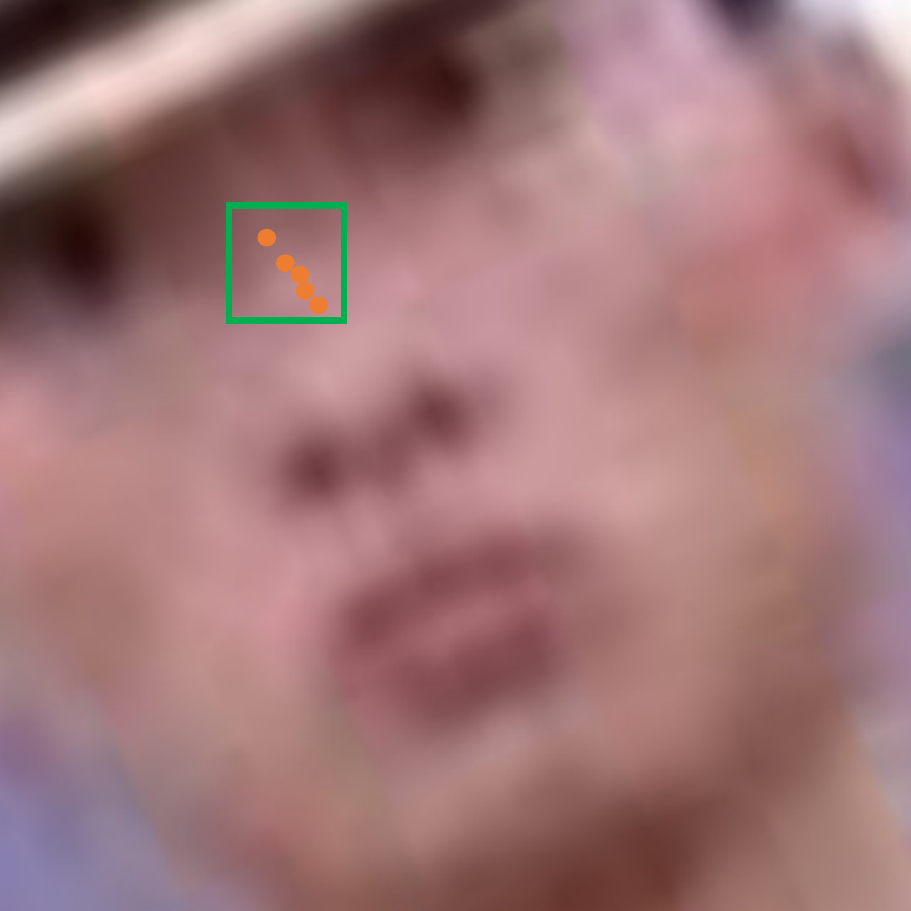}
        \end{minipage} &
        \begin{minipage}{0.11\linewidth}
            \includegraphics[width=\linewidth, height=\textwidth]{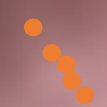}
        \end{minipage} &
        \begin{minipage}{0.11\linewidth}
            \includegraphics[width=\linewidth, height=\textwidth]{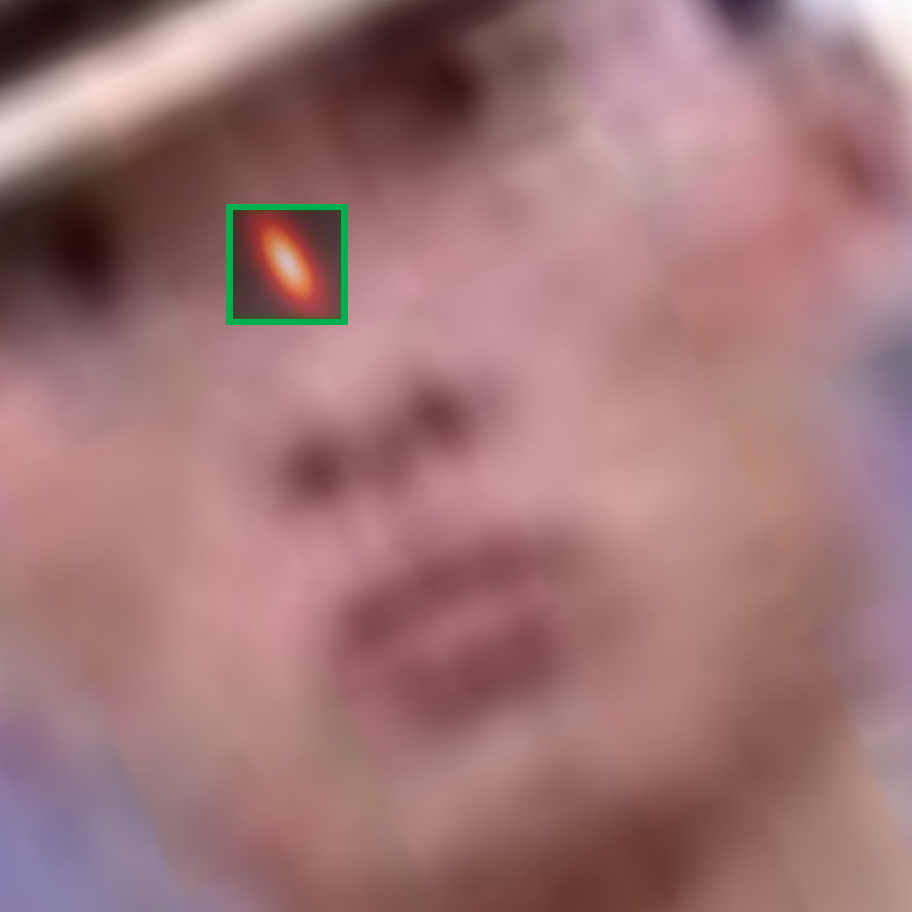}
        \end{minipage} &
        \begin{minipage}{0.11\linewidth}
            \includegraphics[width=\linewidth, height=\textwidth]{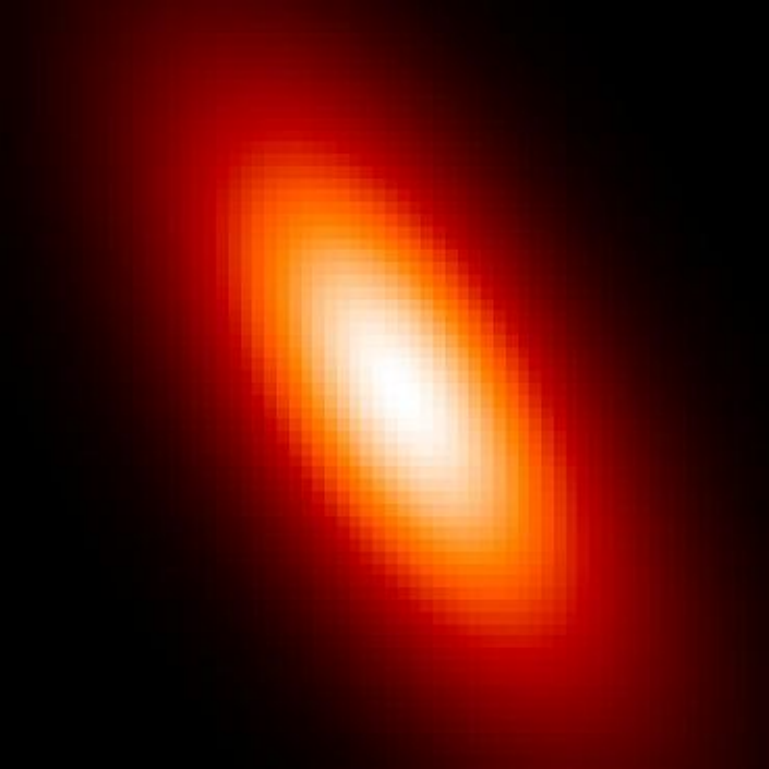}
        \end{minipage} 
        &
        \begin{minipage}{0.11\linewidth}
            \includegraphics[width=\linewidth, height=\textwidth]{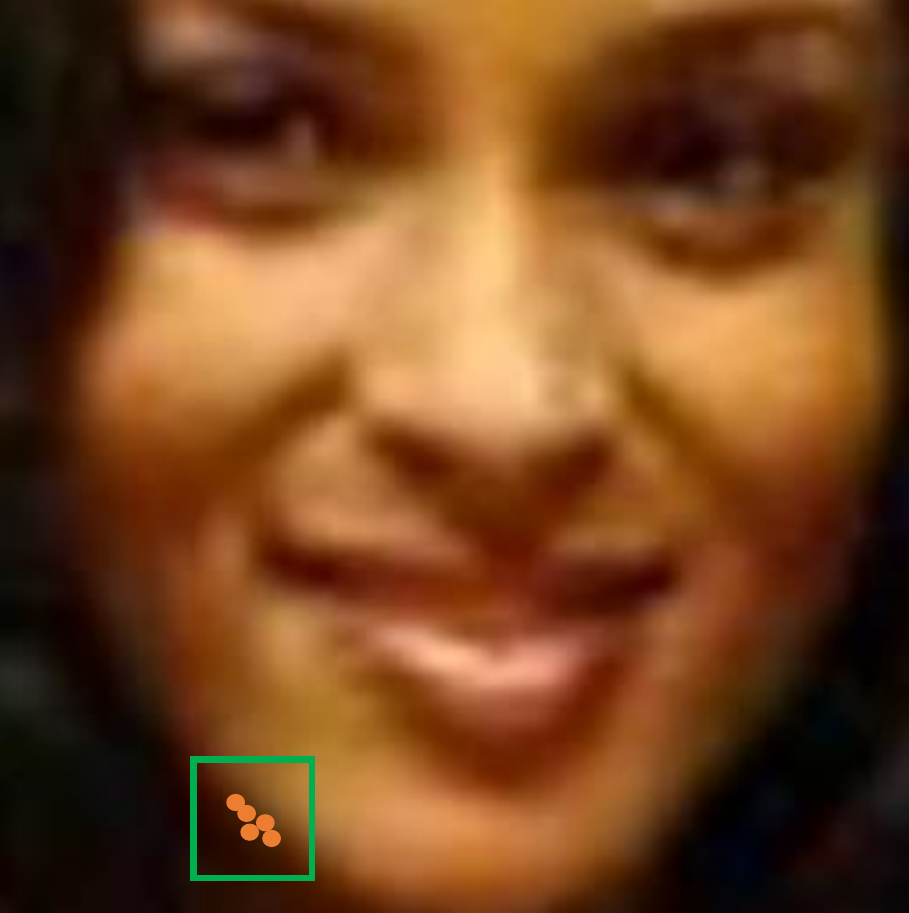}
        \end{minipage} &
        \begin{minipage}{0.11\linewidth}
            \includegraphics[width=\linewidth, height=\textwidth]{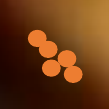}
        \end{minipage} &
        \begin{minipage}{0.11\linewidth}
            \includegraphics[width=\linewidth, height=\textwidth]{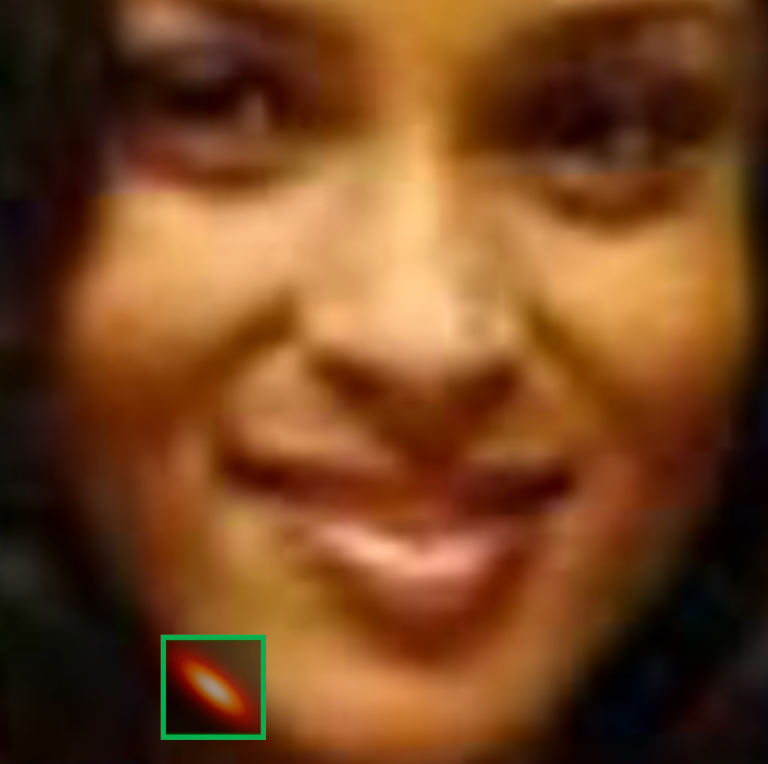}
        \end{minipage} &
        \begin{minipage}{0.11\linewidth}
            \includegraphics[width=\linewidth, height=\textwidth]{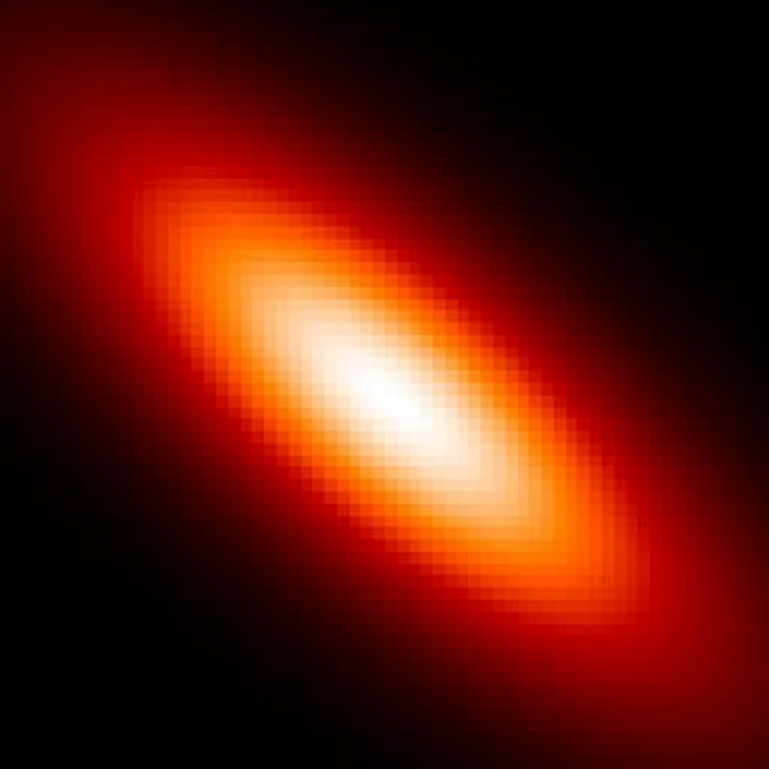}
        \end{minipage} 
        \vspace{8pt}
        \\
        \begin{minipage}{0.11\linewidth}
            \includegraphics[width=\linewidth, height=\textwidth]{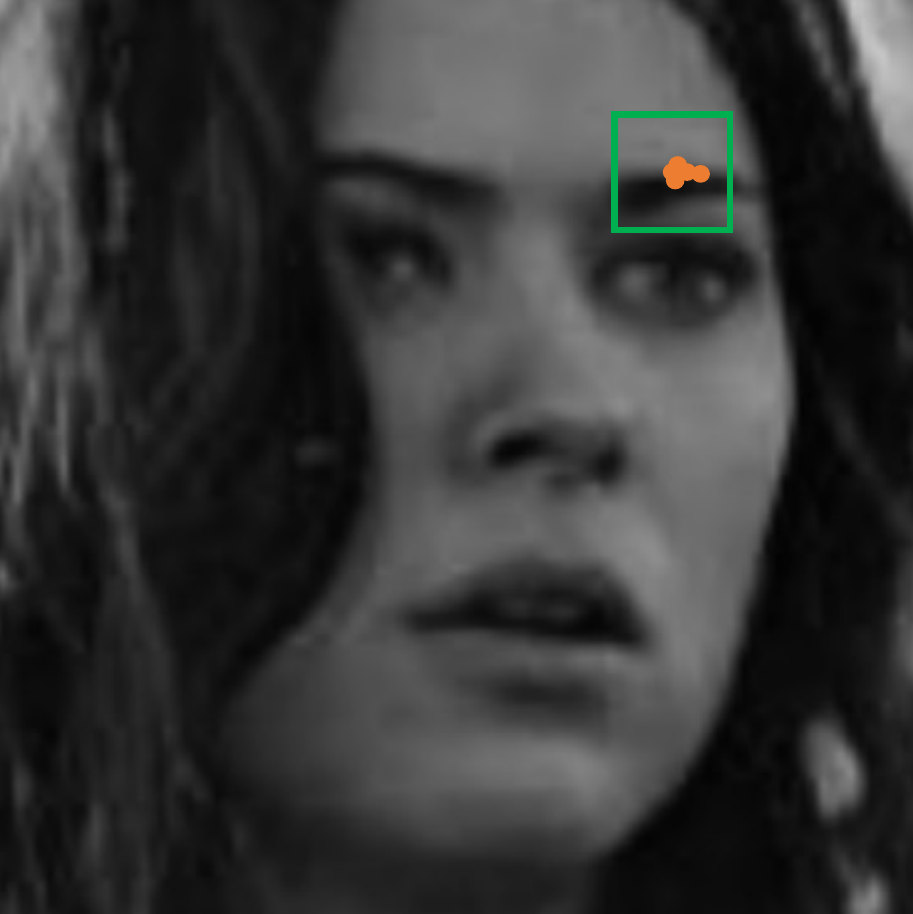}
        \end{minipage} &
        \begin{minipage}{0.11\linewidth}
            \includegraphics[width=\linewidth, height=\textwidth]{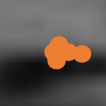}
        \end{minipage} &
        \begin{minipage}{0.11\linewidth}
            \includegraphics[width=\linewidth, height=\textwidth]{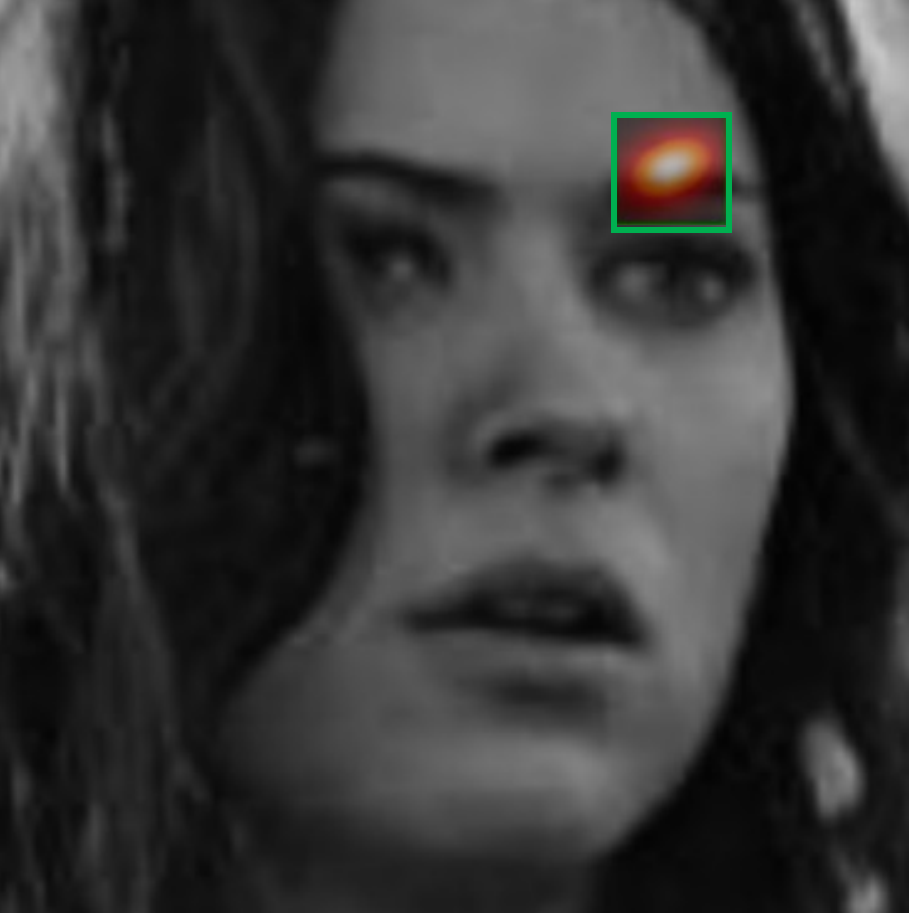}
        \end{minipage} &
        \begin{minipage}{0.11\linewidth}
            \includegraphics[width=\linewidth, height=\textwidth]{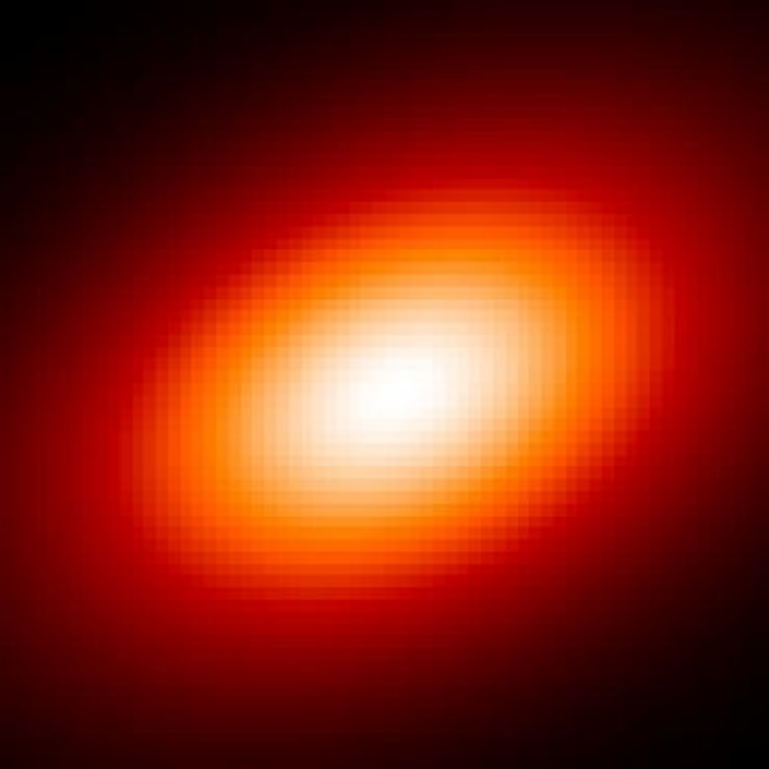}
        \end{minipage} 
        &
        \begin{minipage}{0.11\linewidth}
            \includegraphics[width=\linewidth, height=\textwidth]{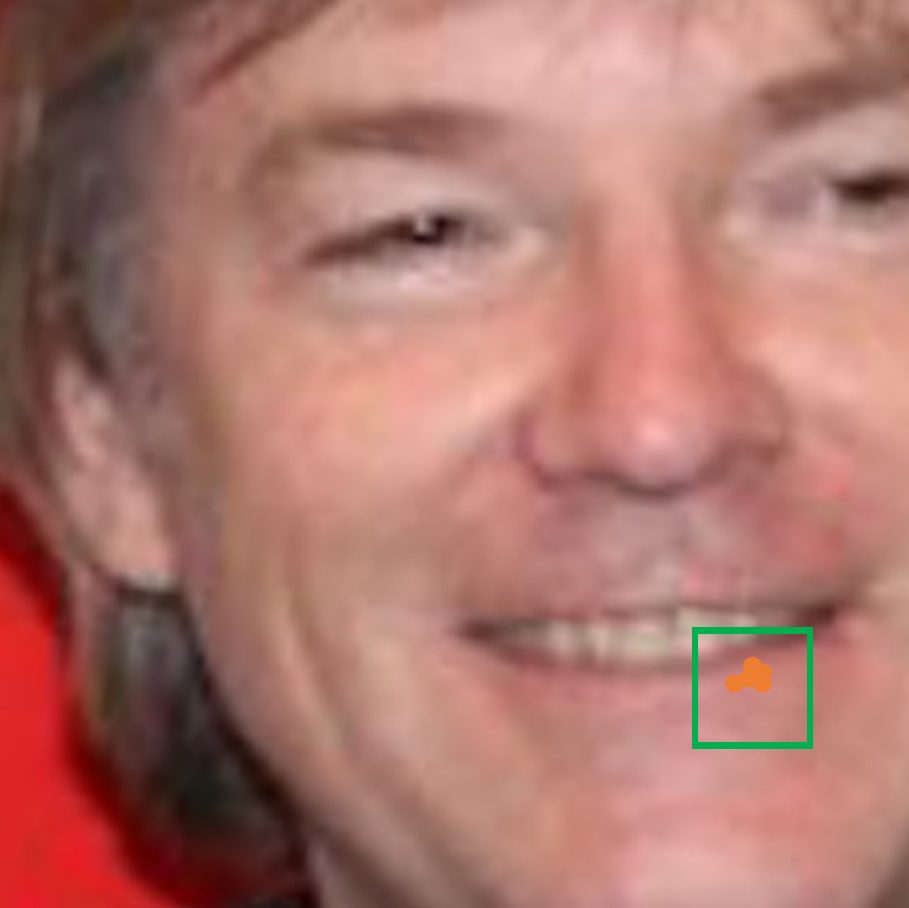}
        \end{minipage} &
        \begin{minipage}{0.11\linewidth}
            \includegraphics[width=\linewidth, height=\textwidth]{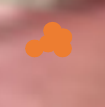}
        \end{minipage} &
        \begin{minipage}{0.11\linewidth}
            \includegraphics[width=\linewidth, height=\textwidth]{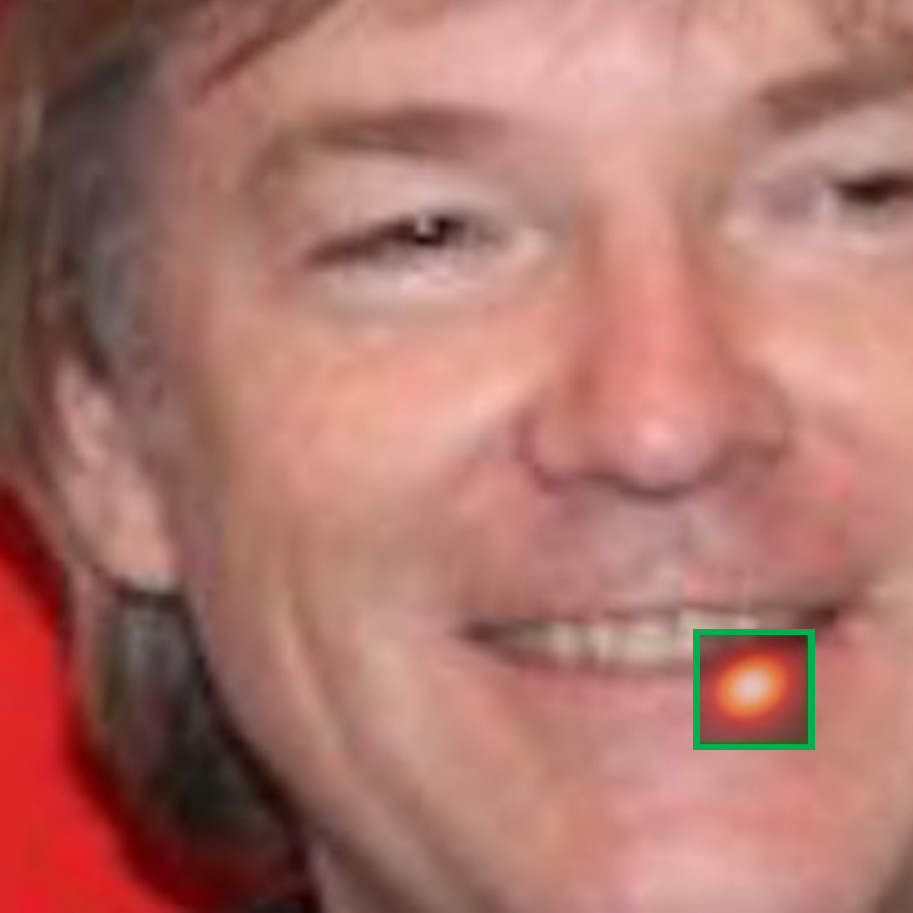}
        \end{minipage} &
        \begin{minipage}{0.11\linewidth}
            \includegraphics[width=\linewidth, height=\textwidth]{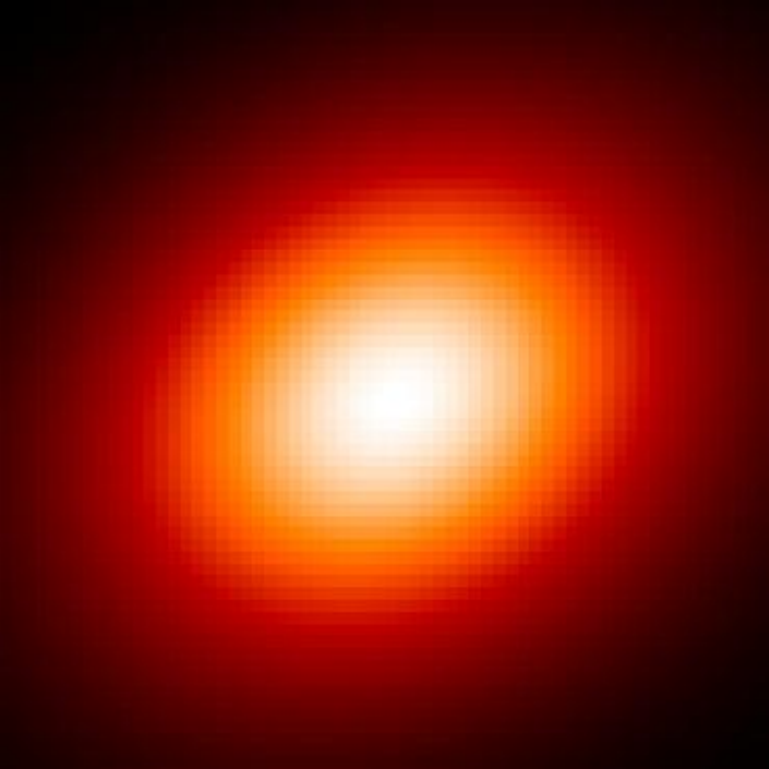}
        \end{minipage} 
        \vspace{8pt}
        \\
    \end{tabular}
    \captionof{figure}{
        \textbf{
            Comparison between real-life annotation ambiguity and our label smoothing.
        } The ``Human'' columns are annotated by 5 users. The ``Ours'' columns are generated from our label smoothing. 
    }
    \label{fig:ls-human}
\end{table}

\subsection{Implementation Details}
\myparagraph{Training details.} 
The training is conducted on 4 NVIDIA A6000 GPUs with 48GB memory. The training batch size is set to 128. 
We used the Adam optimizer~\cite{adam} with a learning rate of 0.001. 
The model is trained for 200 epochs and takes roughly 10 hours to finish.
For the loss function, we follow \citet{zhou2023star, huang2021adnet} and incorporate AAM~\cite{huang2021adnet}. 
AMM is trained on an 
auxiliary task of predicting edge heatmap ($\gL_{\tt Awing}$)
derived from the boundary lines.
The total loss function is as follows, 
\bea
\gL_{\tt total} = \gL_{\tt Awing} + \lambda \gL_{\tt ours},
\label{eq:final-loss}
\eea where $\gL_{\tt Awing}$ is the AwingLoss~\cite{wang2019awing} for training the AAM~\cite{huang2021adnet} and $\lambda$ controls the weight of our loss $\gL_{\tt ours}$ from~\equref{eq:loss-smooth}.

\myparagraph{Label smoothing.}
Here we provide the implementation details of our label smoothing. Please also refer to ~\figref{fig:label_smoothing}.
Given $N$ landmarks $\vy = [ \vy_1, \vy_2, \cdots, \vy_N ]$ of an input facial image of size 256px, we the follow hand-designed procedure~\cite{wu2018lab, wang2019awing} and construct an edge heatmap $\mE$ of size 64px. We then apply the {\tt Pytorch} Gaussian blur with a kernel size of 9 and the sharpness filter with a factor of 5 onto $\mE$. The refined edge heatmap is denoted as $\mE'$. For each landmark $\vy_n$, we consider the patch of size $2k+1$ around $\vy_n$ on $\mE'$, denoted as $\mE'_n$. On the other hand, we construct a Gaussian heatmap $\mN_n$ centered around $\vy_n$ with a kernel size of 5. After both $\mE'_n$ and $\mN_n$ are normalized so that their maximum value is 1, we construct a joint heatmap $\mM_n = 0.01 \mE_n' + \mN_n$.  Finally, we build the directional Gaussian heatmap $\mG$ for image-aware label smoothing based on the covariance of $\mM_n$, i.e. $\mG_n = \gN(\vy_n, \gamma \Sigma_{\mM_n})$. The $\gamma$ is set to 0.001 for COFW but 0.01 for WFLW and 300W.

For each image, we sample 10 $\vy'_n$ following the distribution $\mG$ from our label smoothing. The sampling strategy is lightweight and does not have an observable impact on the overall inference speed. 

\begin{figure*}[h!]
    \begin{subfigure}[h!]{0.32\textwidth}
        \includegraphics[width=\textwidth]{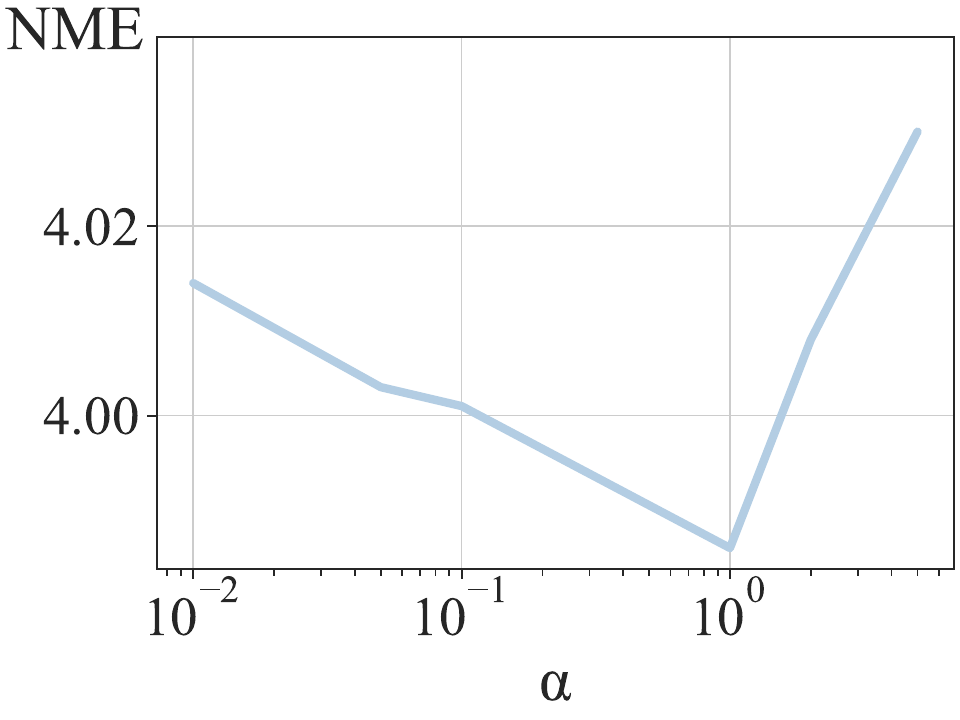}
        \caption{Different choices of $\alpha$.}
    \end{subfigure}
    \hfill
    \begin{subfigure}[h!]{0.32\textwidth}
        \includegraphics[width=\textwidth]{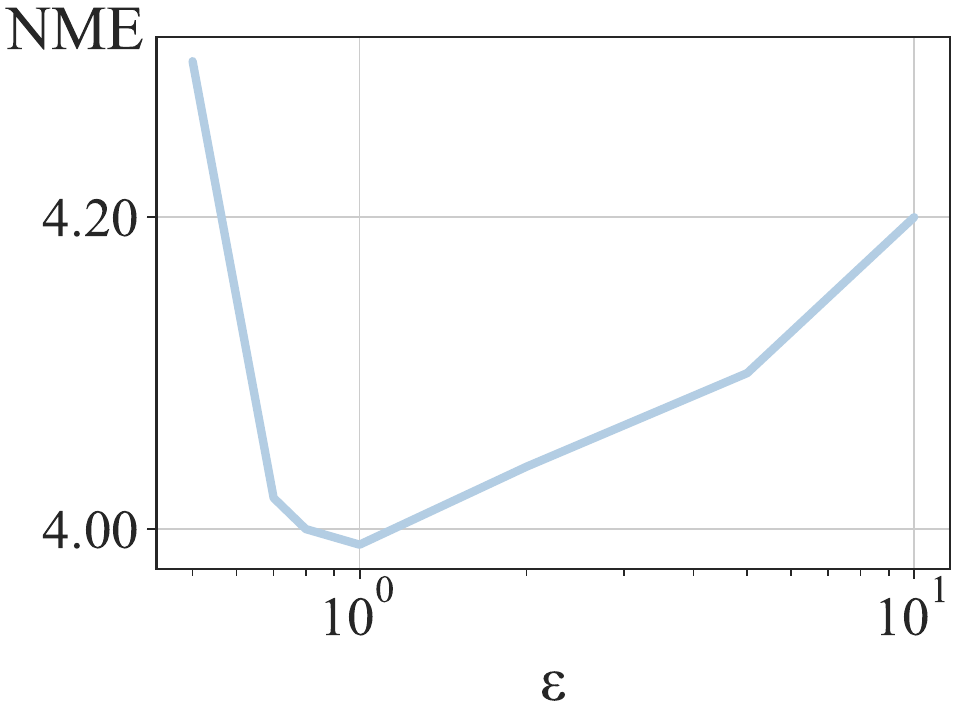}
        \caption{Different choices of $\epsilon$.}
    \end{subfigure}
    \hfill
    \begin{subfigure}[h!]{0.32\textwidth}
        \includegraphics[width=\textwidth]{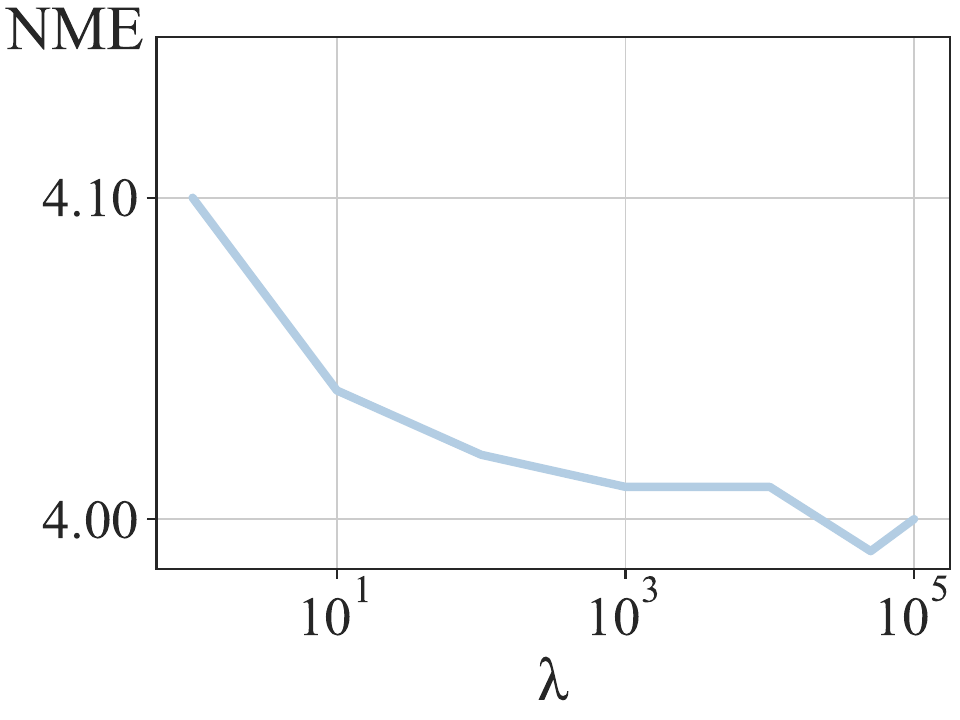}
        \caption{Different choices of $\lambda$.}
    \end{subfigure}

  \caption{ \textbf{The NME($\downarrow$) of different choices of hyper-parameters on WFLW.}
  }
  \label{fig:ablation-coeffecients}
\end{figure*}

\myparagraph{Hyperparameters.} We identify the following hyper-parameters which affect the performance of our approach: $\alpha$~(Eq.~\ref{eq:smooth_l1}), $\eps$ ~(Eq.~\ref{eq:loss-augmented}), and $\lambda$~(Eq.~\ref{eq:final-loss}).
We experiment with different choices, \ie, a grid search, on these hyperparameters to observe their impact on the performance. The results are reported in~\figref{fig:ablation-coeffecients}. It is shown that our method is sensitive to the choices of $\eps$ and $\lambda$. Overall, we choose $\alpha = 1, \eps = 1$ and $\lambda = 5000$.

\myparagraph{Architecture.} Our architecture strictly follows the stacked hourglass network used in STAR and ADNet. Specifically, it has a parameter size of 13.48M and the FLOPs are 17.54G. The throughput is 50 images of 256 pixels by 256 pixels per second.

\section{Additional Results}
\label{supp:results}
In the main paper, we do not use the same evaluation settings across the three datasets presented in \tabref{tab:main-WFLW} to \tabref{tab:main-300W}. This is because not all works report their results on all datasets (WFLW, COFW, and 300W) or all evaluation metrics (NME, FR, and AUC) and settings (inter-ocular or inter-pupil), especially the older methods pre-2020. Additionally, some works do not have open-sourced code or are not reproducible. Note, that our experiment setup follows the state-of-the-art STAR~\cite{zhou2023star}, i.e., we compared our method to the same baseline they do. For completeness, we report the comparison of additional evaluation metrics in the following paragraphs.

\myparagraph{Additional ablation studies.}
In \tabref{tab:ablation-smooth-WFLW}, we evaluate the robustness of our label smoothing under challenging conditions. We observe a consistent gain on most subsets. 

\begin{table*}[h!]
    \centering
    \begin{minipage}[t]{0.8\linewidth}
    \small

\centering
\resizebox{\columnwidth}{!}{

\begin{tabular}{ccccccc} 
\specialrule{.15em}{.05em}{.05em}
Label smoothing  & 
WFLW-L & WFLW-E & WFLW-I
& WFLW-M
& WFLW-O
& WFLW-B
\\ 
\cmidrule{1-7}
 \ding{55} & {6.68} & \best{4.26} & {3.94} & {3.93} & {4.77} & \best{4.56} \\
\rowcolor{OursColor} \checkmark & \best{6.58} & \best{4.26} & \best{3.90} & \best{3.89} & \best{4.74} & \second{4.57} \\
\specialrule{.15em}{.05em}{.05em}
\end{tabular}
}
\caption{
    \textbf{Ablation of label smoothing on the subsets of WFLW.} We ablate our label smoothing and report the inter-ocular NME$\downarrow$ on the six subsets of WFLW~\cite{burgos2013robust}.
    \label{tab:ablation-smooth-WFLW}
}
\vspace{-0.3cm}

    \end{minipage}
\end{table*}

\begin{figure*}[h!]
    \includegraphics[width=\textwidth]{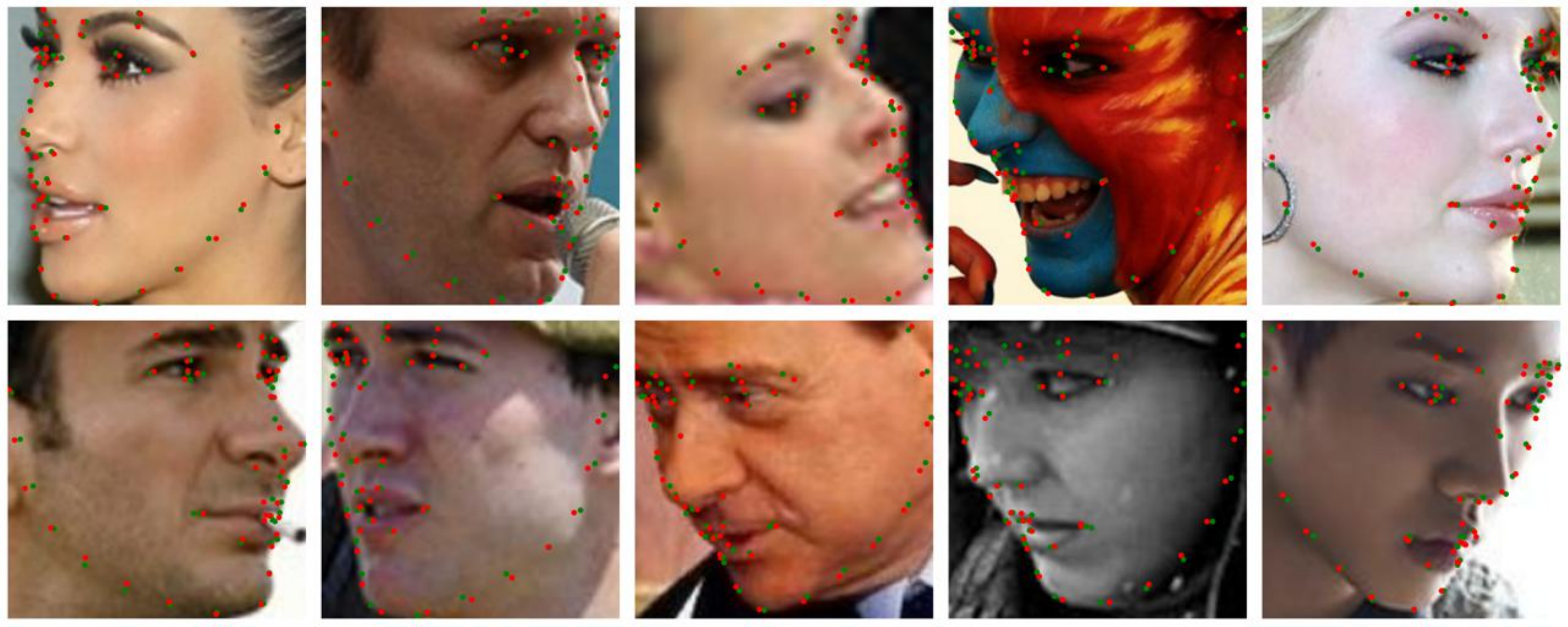}
  \caption{\textbf{Qualitative results on the ``largepose'' subset of WFLW.} The ground truth landmarks are marked in \darkgreen{green} while our predictions are marked in \textcolor{red}{red}.
  }
  \label{fig:qual-WFLW-L}
\end{figure*}

\begin{figure*}[h!]
    \includegraphics[width=\textwidth]{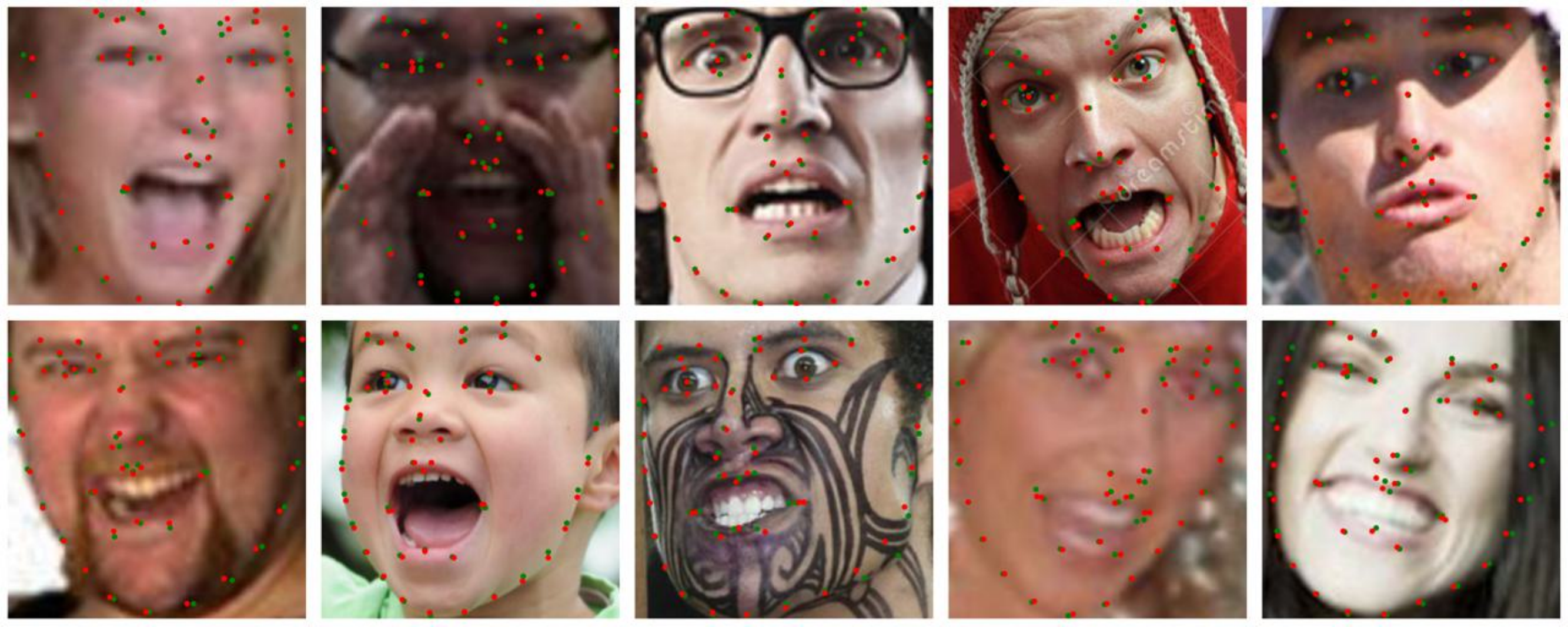}
  \caption{\textbf{Qualitative results on the ``expression'' subset of WFLW.} The ground truth landmarks are marked in \darkgreen{green} while our predictions are marked in \textcolor{red}{red}.
  }
  \label{fig:qual-WFLW-E}
\end{figure*}

\begin{figure*}[h!]
    \includegraphics[width=\textwidth]{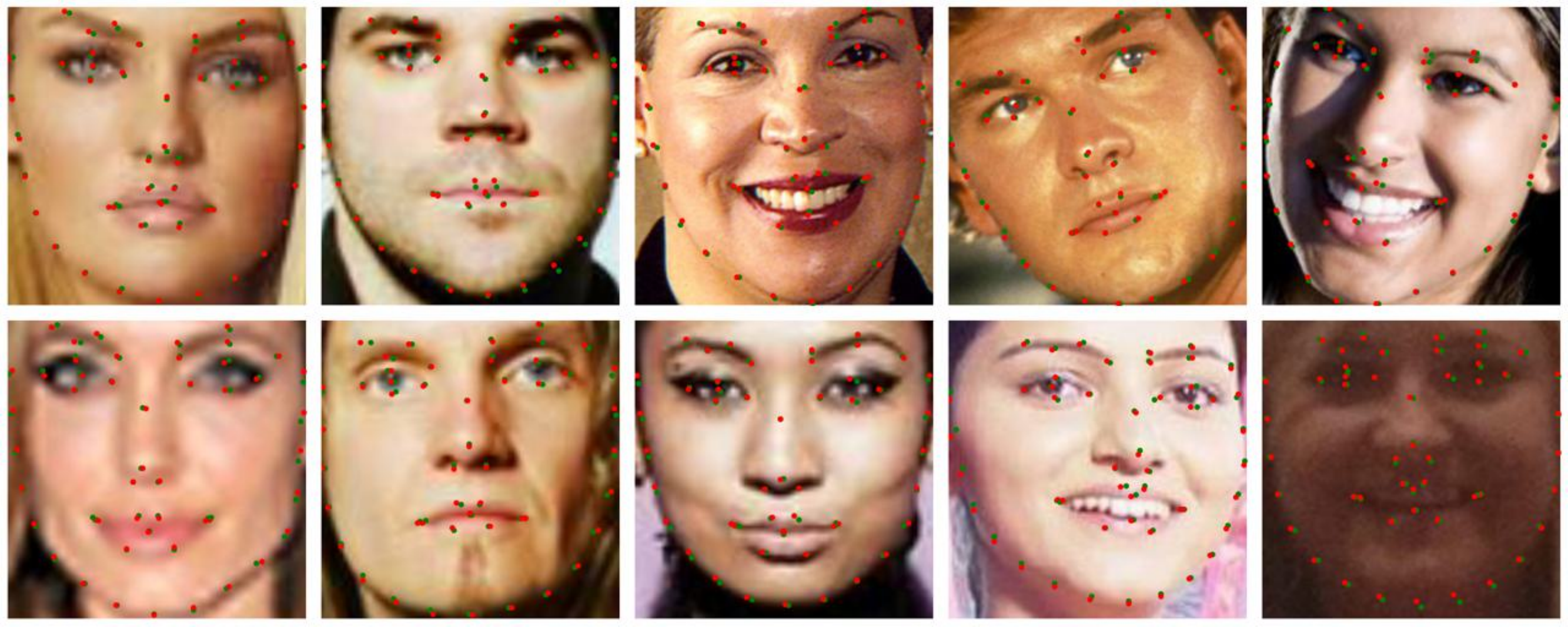}
  \caption{\textbf{Qualitative results on the ``illumination'' subset of WFLW.} The ground truth landmarks are marked in \darkgreen{green} while our predictions are marked in \textcolor{red}{red}.
  }
  \label{fig:qual-WFLW-I}
\end{figure*}

\begin{figure*}[h!]
    \includegraphics[width=\textwidth]{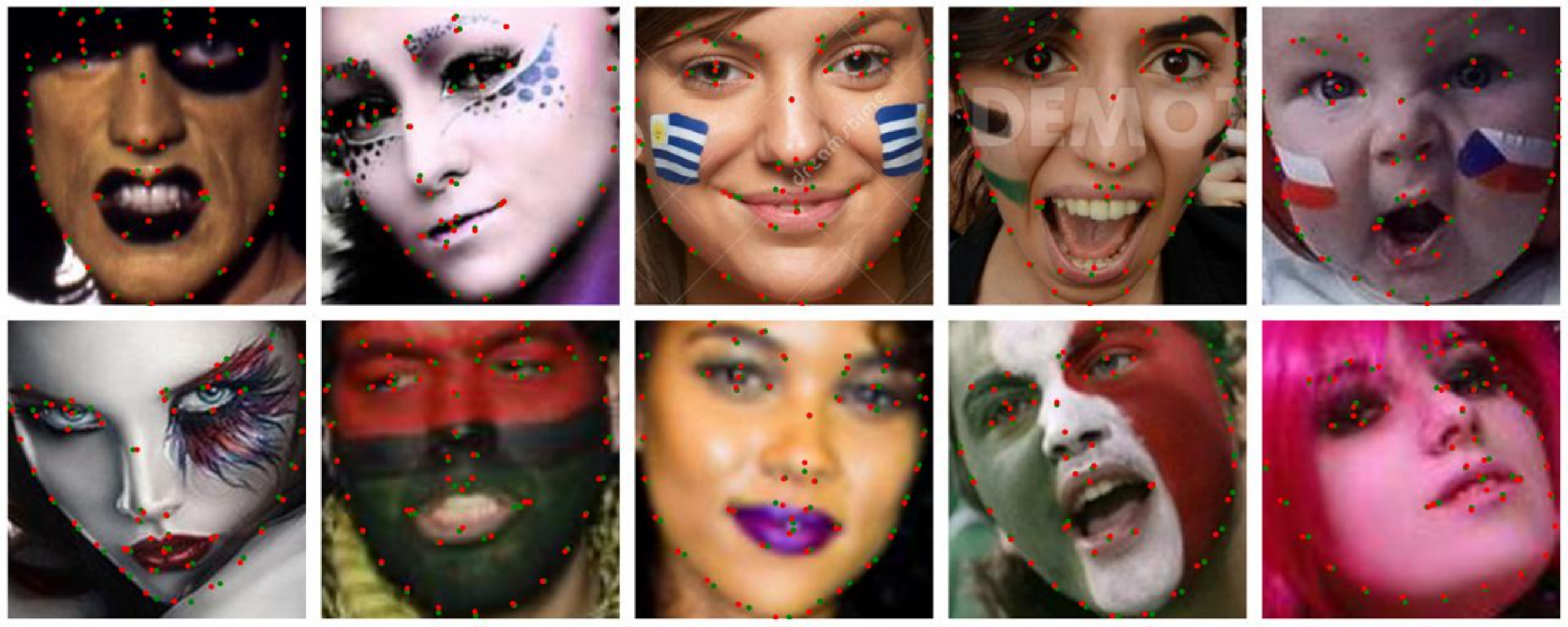}
  \caption{\textbf{Qualitative results on the ``makeup'' subset of WFLW.} The ground truth landmarks are marked in \darkgreen{green} while our predictions are marked in \textcolor{red}{red}.
  }
  \label{fig:qual-WFLW-M}
\end{figure*}

\begin{figure*}[h!]
    \includegraphics[width=\textwidth]{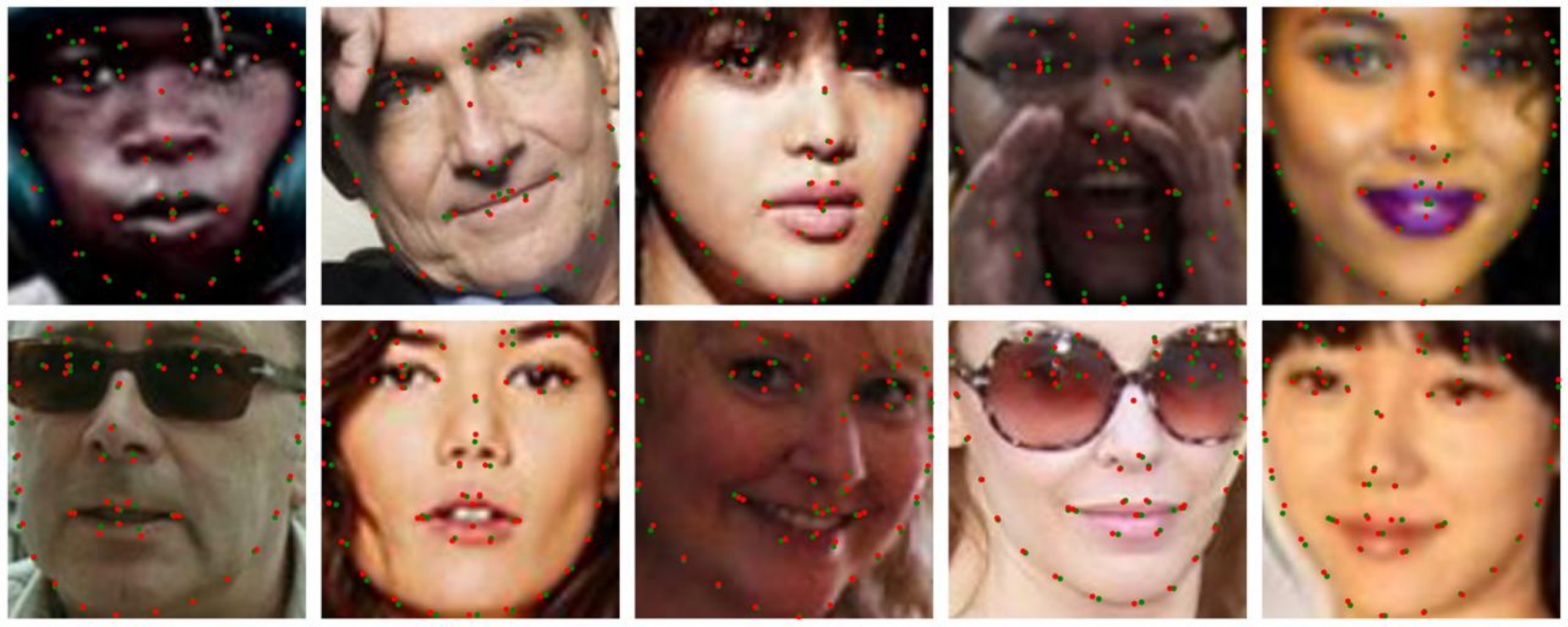}
  \caption{\textbf{Qualitative results on the ``occlusion'' subset of WFLW.} The ground truth landmarks are marked in \darkgreen{green} while our predictions are marked in \textcolor{red}{red}.
  }
  \label{fig:qual-WFLW-O}
\end{figure*}

\begin{figure*}[h!]
    \includegraphics[width=\textwidth]{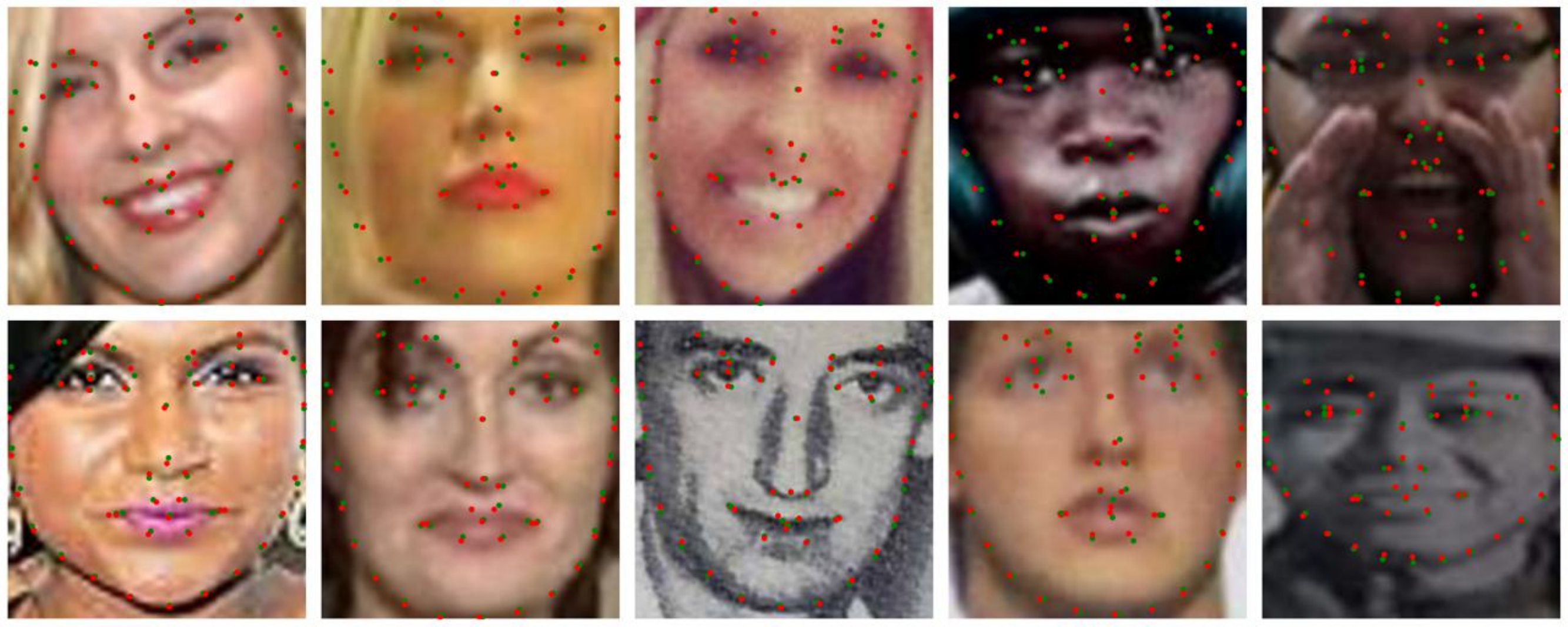}
  \caption{\textbf{Qualitative results on the ``blur'' subset of WFLW.} The ground truth landmarks are marked in \darkgreen{green} while our predictions are marked in \textcolor{red}{red}.
  }
  \label{fig:qual-WFLW-B}
\end{figure*}

\myparagraph{Additional results on WFLW.}
In \tabref{tab:WFLW-subsets-FR} and \tabref{tab:WFLW-subsets-AUC}, we report the comparison of FR($\downarrow$) and AUC($\uparrow$) to SOTA facial landmark detection methods on the six subsets of WFLW. We achieve competitive results on all subsets. Notably, our method excels in the ``largepose'' subset. We provide qualitative results of the six subsets in~\figref{fig:qual-WFLW-L},~\figref{fig:qual-WFLW-E},~\figref{fig:qual-WFLW-I},~\figref{fig:qual-WFLW-M},~\figref{fig:qual-WFLW-O}, and~\figref{fig:qual-WFLW-B}.
\begin{table*}[h!]
\small
\centering
\resizebox{0.8\linewidth}{!}{

\begin{tabular}{lccccccc} 
\specialrule{.15em}{.05em}{.05em}
Method & 
Type  & 
WFLW-L & WFLW-E & WFLW-I
& WFLW-M
& WFLW-O
& WFLW-B
\\ 
\cmidrule{1-8}
Wing~\cite{feng2018wing} & C & 22.70 & 4.78 & 4.30 & 7.77 & 12.50 & 7.76\\
\cmidrule{1-8}
LAB~\cite{wu2018lab} & H & 28.83 & 6.37 & 6.73 & 7.77 & 13.72 & 10.74 \\
DeCaFA~\cite{dapogny2019decafa} & H & 21.40 & 3.73 & 3.22 & 6.15 & 9.26 & 6.61\\
AWing~\cite{wang2019awing} & H & 13.50 & 2.23 & 2.58 & 2.91 & 5.98 & 3.75 \\
ADNet~\cite{huang2021adnet} & H & 12.72 & 2.15 & 2.44 & 1.94 & 5.79 & 3.54 \\
HIH~\cite{lan2021hih} & H & 12.88 & \best{1.27} & 2.43 & \second{1.45} & 5.16 & 3.10\\
STAR~\cite{zhou2023star} & H & \second{10.77} & 2.24 & \second{1.58} & \best{0.98} & \second{4.76} & \second{2.98} \\
\rowcolor{OursColor} Ours & H & \best{9.23} & \second{1.92} & \best{0.86} & 1.46 & \best{3.95} & \best{2.33} \\
\specialrule{.15em}{.05em}{.05em}
\end{tabular}
}
\caption{
    \textbf{Comparison to SOTA facial landmark detection methods on the subsets of WFLW.} We report the FR$\downarrow$ on the six subsets of WFLW~\cite{burgos2013robust},~\ie, large pose (WFLW-L), expression (WFLW-E), illumination (WFLWI), make-up (WFLW-M), occlusion (WFLW-O), and blur (WFLW-B). Type ``C'' and ``H'' stand for coordinate-regression and heatmap-regression methods. 
    \label{tab:WFLW-subsets-FR}
}
\vspace{-0.3cm}
\end{table*}

\begin{table*}[h!]
\small
\centering
\resizebox{0.8\linewidth}{!}{

\begin{tabular}{lccccccc} 
\specialrule{.15em}{.05em}{.05em}
Method & 
Type  & 
WFLW-L & WFLW-E & WFLW-I
& WFLW-M
& WFLW-O
& WFLW-B
\\ 
\cmidrule{1-8}
Wing~\cite{feng2018wing} & C & 0.310 & 0.496 & 0.541 & 0.558 & 0.489 & 0.491\\
\cmidrule{1-8}
LAB~\cite{wu2018lab} & H & 0.235 & 0.495 & 0.543 & 0.539 & 0.449 & 0.463 \\
DeCaFA~\cite{dapogny2019decafa} & H & 0.292 & 0.546 & 0.579 & 0.575 & 0.485 & 0.494 \\
AWing~\cite{wang2019awing} & H & 0.312 & 0.515 & 0.578 & 0.572 & 0.502 & 0.512 \\
ADNet~\cite{huang2021adnet} & H & 0.344 & 0.523 & 0.581 & 0.601 & 0.530 & 0.548 \\
HIH~\cite{lan2021hih} & H & 0.358 & \best{0.601} & \best{0.613} & 0.618 & \best{0.539} & \best{0.561} \\
STAR~\cite{zhou2023star} & H & \second{0.362} & 0.584 & 0.609 & \best{0.622} & \second{0.538} &0.551 \\
\rowcolor{OursColor} Ours & H & \best{0.370} & \second{0.587} & \second{0.612} & \second{0.619} & \best{0.539} & \second{0.552} \\
\specialrule{.15em}{.05em}{.05em}
\end{tabular}
}
\caption{
    \textbf{Comparison to SOTA facial landmark detection methods on the subsets of WFLW.} We report the AUC$\uparrow$ on the six subsets of WFLW~\cite{burgos2013robust},~\ie, large pose (WFLW-L), expression (WFLW-E), illumination (WFLWI), make-up (WFLW-M), occlusion (WFLW-O), and blur (WFLW-B). Type ``C'' and ``H'' stand for coordinate-regression and heatmap-regression methods. 
    \label{tab:WFLW-subsets-AUC}
}
\vspace{-0.3cm}
\end{table*}

\begin{table}[h!]
    \centering
    \begin{minipage}[t]{0.4\textwidth}
        \vspace{0pt}
        \setlength{\tabcolsep}{5pt}
\centering
\begin{tabular}{lccccccccc} 
\specialrule{.15em}{.05em}{.05em}
\multirow{2}{*}{Method} & \multicolumn{2}{c}{Inter-Pupil} & \multicolumn{2}{c}{Inter-Ocular}
\\
& \multicolumn{1}{c}{NME $\downarrow$} & $\text{FR} \downarrow$ & \multicolumn{1}{c}{NME $\downarrow$} & $\text{FR} \downarrow$ \\ 
\midrule
TCDCN~\cite{zhang2014facial} & 8.05 & - & - & - \\
\citet{wu2015robust} & 5.93 & - & - & - \\
DAC-CSR~\cite{feng2017dynamic} & - & - & 6.03 & 4.73  \\
Wing~\cite{feng2018wing} & 5.44 &  3.75 & - & - \\
DCFE~\cite{valle2018deeply} & 5.27 & 7.29 & - & -  \\
LAB~\cite{wu2018lab} & - & - & 3.92 & 0.39 \\
SDFL~\cite{lin2021structure} & - & - & 3.63 & \best{0.00} \\
SLPT~\cite{xia2022splt} & 4.79 & 1.18 & 3.32 & \best{0.00} \\
Awing~\cite{wang2019awing} & 4.94 & 0.99 & - & - \\
ADNet~\cite{huang2021adnet} &  4.68 & \second{0.59} & - & - \\
HIH~\cite{lan2021hih} & {4.63} & \best{0.39} & - & - \\
STAR~\cite{zhou2023star}     &  \second{4.62} & {0.79} & \second{3.21} & \best{0.00} \\
\rowcolor{OursColor} Ours & \best{4.58} & {0.79} & \best{3.15} & \best{0.00}
\\
\specialrule{.15em}{.05em}{.05em}
\end{tabular}
\captionof{table}{
    \textbf{Additional comparison to SOTA on COFW.} 
    \label{tab:supp-COFW}
}

    \end{minipage}
    \hfill
    \begin{minipage}[t]{0.4\textwidth}
        \vspace{0pt}
        \setlength{\tabcolsep}{5pt}
\centering
\begin{tabular}{lccccccccc} 
\specialrule{.15em}{.05em}{.05em}
{\multirow{1}{*}{Method}} & 
Full & Comm. & Chal. \\ 
\midrule
SDM~\cite{xiong2013supervised} & 7.50 & 5.57 & 15.40  \\
CFSS~\cite{zhu2015face} & 5.76 & 4.73 & 9.98 \\
MDM~\cite{trigeorgis2016mnemonic} & 5.88 & 4.83 & 10.14 \\
RAR~\cite{xiao2016robust} & 4.94 & 4.12 & 8.36 \\
DVLN~\cite{wu2017leveraging} & 4.66 & 3.94 & 7.62 \\
HG-HSLE~\cite{xu2019hsle} & 4.59 & 3.94 & 7.24 \\
DCFE~\cite{valle2018deeply} & 4.55 & 3.83 & 7.54 \\ 
LAB~\cite{wu2018lab} & 4.12 & 3.42 & 6.98 \\
Wing~\cite{feng2018wing} & 4.04 & \best{3.27} & 7.18 \\
Awing~\cite{wang2019awing} & 4.31 & 3.77 & 6.52 \\
ADNet~\cite{huang2021adnet} & 4.08 & 3.51 & 6.47 \\
STAR~\cite{zhou2023star} & \best{4.03} & 3.55 & 6.22 \\
\rowcolor{OursColor} Ours & \best{4.03} & {3.51} & \best{6.09} 
\\
\specialrule{.15em}{.05em}{.05em}
\end{tabular}
\captionof{table}{
    \textbf{Inter-pupil NME$\downarrow$ comparisons on 300W and common/challenging subsets.} 
    \label{tab:supp-300W}
}

\end{minipage}
\end{table}

\begin{figure*}[h!]
    \includegraphics[width=\textwidth]{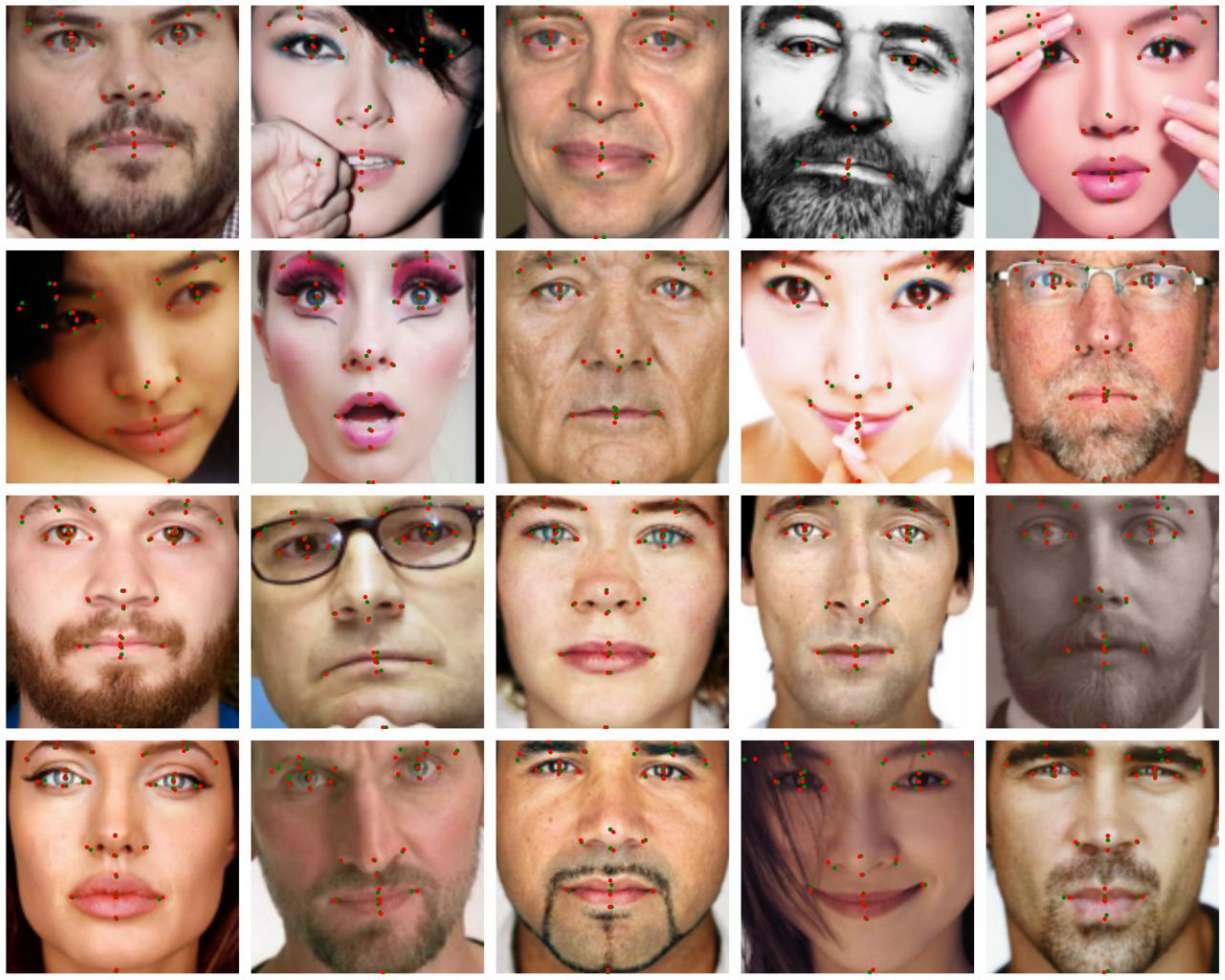}
  \caption{\textbf{Qualitative results on COFW.} The ground truth landmarks are marked in \darkgreen{green} while our predictions are marked in \textcolor{red}{red}.
  }
  \label{fig:qual-COFW}
\end{figure*}

\myparagraph{Additional results on COFW.}
In ~\tabref{tab:supp-COFW}, we report more quantitative comparisons on COFW.
We also provide qualitative results of COFW in~\figref{fig:qual-COFW}.

\begin{figure*}[h!]
    \includegraphics[width=\textwidth]{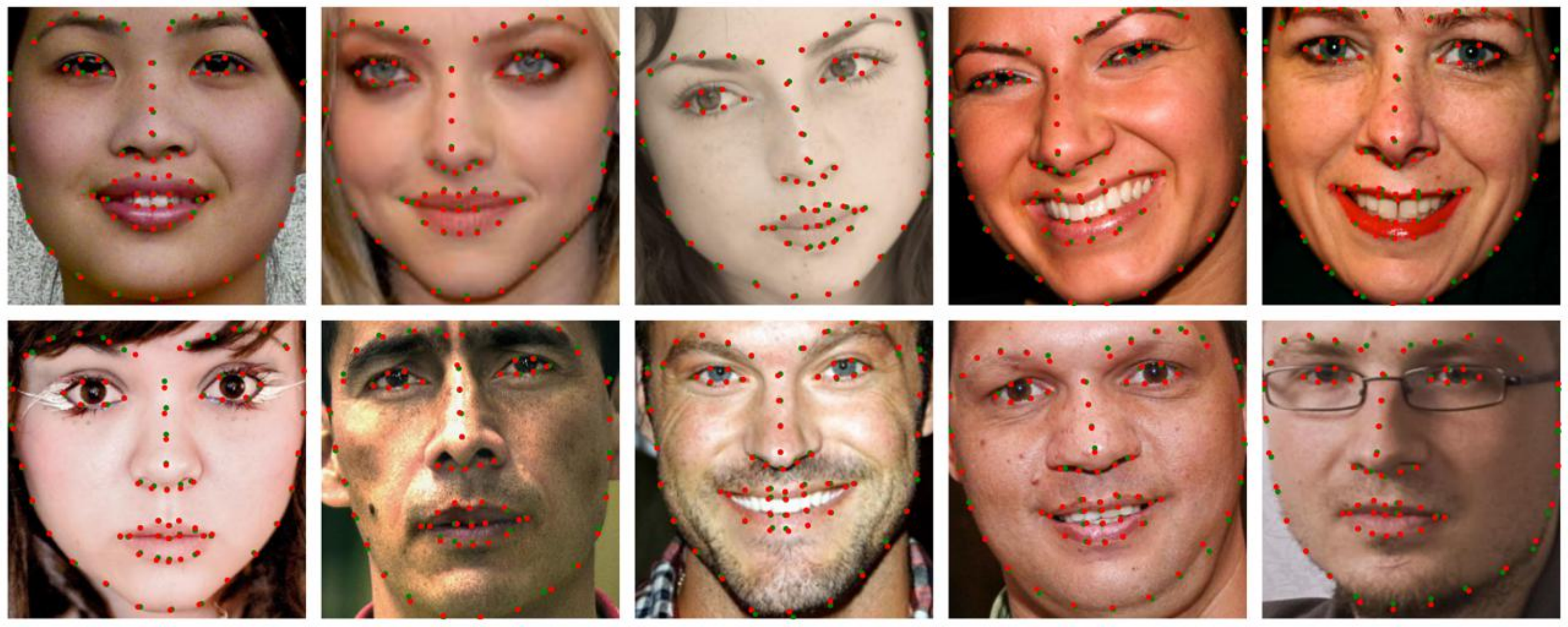}
  \caption{\textbf{Qualitative results on the ``common'' subset of  COFW.} The ground truth landmarks are marked in \darkgreen{green} while our predictions are marked in \textcolor{red}{red}.
  }
  \label{fig:qual-300W-comm}
\end{figure*}

\begin{figure*}[h!]
    \includegraphics[width=\textwidth]{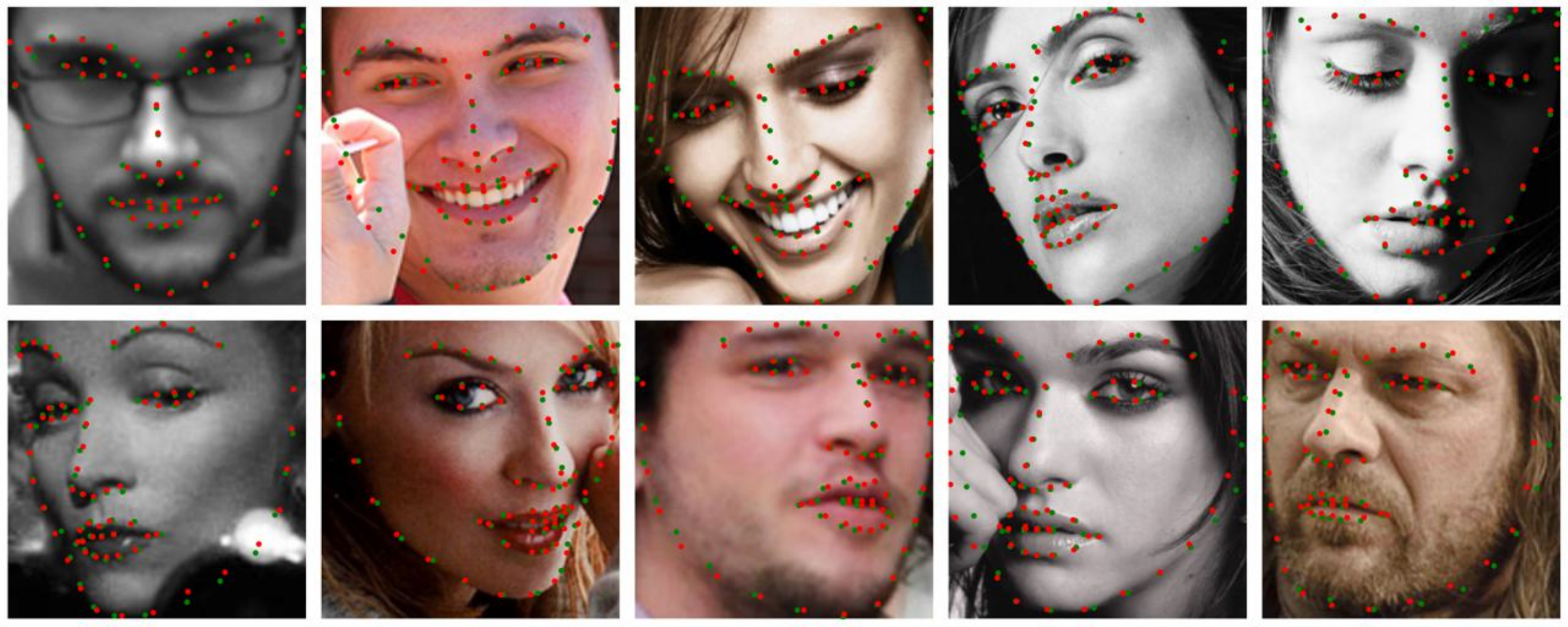}
  \caption{\textbf{Qualitative results on the ``challenge'' subset of 300W.} The ground truth landmarks are marked in \darkgreen{green} while our predictions are marked in \textcolor{red}{red}.
  }
  \label{fig:qual-300W-chal}
\end{figure*}

\myparagraph{Additional results on 300W.}
In ~\tabref{tab:supp-300W}, we report more quantitative comparisons on COFW.
We provide qualitative results of the two subsets of 300W in~\figref{fig:qual-300W-comm} and~\figref{fig:qual-300W-chal}.

\end{document}